%% file: arxiv.tex
\documentclass[11pt]{article}

\usepackage[a4paper,margin=1in]{geometry}
\usepackage{amsmath,amsfonts,amssymb}
\usepackage{graphicx}
\usepackage{xcolor}
\usepackage[caption=false]{subfig}
\usepackage{booktabs}
\usepackage{longtable}
\usepackage{pdflscape}
\usepackage[section]{placeins}
\usepackage{cite}
\usepackage[hidelinks]{hyperref}

\title{Overlapping Visual Grouping Without Semantic Priors}
\author{%
Teemu Saukkio$^{1,*}$ \quad Hashem Haghbayan$^{1}$ \quad Juha Plosila$^{1}$\\[0.6em]
\small $^{1}$University of Turku, Faculty of Technology, Department of Computing, Turku, Finland\\
\small $^{*}$Corresponding author: \href{mailto:tjsauk@utu.fi}{tjsauk@utu.fi}
}
\date{}

\begin{document}
\maketitle

\begin{abstract}
Most computer-vision systems organize visual input toward a predefined
interpretation, such as semantic categories, prompted regions, learned
object-like representations, or a single spatial partition. This work
considers an earlier stage of visual organization: the formation of candidate
perceptual units directly from sensor measurements before their identity,
meaning, or task relevance is known. We introduce Domain Parent Grouping
(DPG), a sensor-grounded grouping method in which complementary measurement
relationships are represented in separate processing domains. Spatially
connected groups formed within these domains are related through cross-domain
overlap, yielding a non-exclusive grouping representation rather than a single
mutually exclusive segmentation. This representation retains broader and more
localized groups, as well as alternative grouping boundaries over the same
image locations, simultaneously available. DPG also includes a native mechanism
for reprocessing selected group content, in which input-relative measurement
ranges allow the observational resolution to change while preserving previously
formed groups. DPG is implemented using three domains representing
locally contextualized luminance, direct chromatic relationships, and
contextual chromatic relationships. Experiments on the BSDS500 dataset
demonstrate the benefit of combining the three domains. The results further
show that DPG forms measurement-supported groups corresponding to low-level
image structure, and that these groups exhibit measurable correspondence with
human-annotated regions and boundaries. This demonstrates that structured visual
organization can emerge directly from relationships among sensor measurements.
\end{abstract}


\section{Introduction}

Contemporary computer vision addresses visual organization through several
task-specific formulations. Object detection seeks to identify and localize
instances belonging to target categories, while semantic segmentation assigns
category labels to image locations. Instance segmentation separates
individual object instances, and panoptic segmentation combines object
instances and background into a coherent interpretation of the
complete image \cite{Kirillov2019,Cheng2022}. Recent foundation-model
approaches extend these formulations through open-vocabulary, referring, and
promptable segmentation, in which the target region can be specified through
text, points, bounding boxes, masks, or other external prompts
\cite{Kirillov2023,Zhou2024}. Self-supervised models can learn general-purpose
visual representations without manually assigned training labels, after
which the representations can be applied to downstream tasks such as
recognition, localization, segmentation, and depth estimation
\cite{Oquab2024}.\\

These approaches differ substantially in their supervision, architecture,
and intended application. In supervised recognition and segmentation, the
target structures are operationalized through category inventories and
annotation protocols, as exemplified by datasets containing labelled objects,
scene regions, materials, and object parts \cite{Lin2014,Zhou2017}. Promptable
methods remove the requirement for a fixed category set, but an external
prompt still specifies the spatial entity to be returned
\cite{Kirillov2023}. Self-supervised and object-centric methods reduce the
dependence on explicit category annotations, but commonly learn reusable
representations or structured object-like components for subsequent tasks
\cite{Locatello2020,Oquab2024}. Classical category-independent methods instead
partition an image according to predefined similarity, connectivity, or
boundary criteria \cite{Felzenszwalb2004,Vedaldi2008}. Thus, the intended
output is generally determined by some combination of a semantic ontology,
an annotation convention, a user prompt, a learned representation, a
predefined grouping criterion, or the requirement to produce one spatial
partition.\\

These formulations are effective when the visual entities required by an
application can be specified in advance. A different requirement arises for
agents operating in open-ended environments, where previously unspecified
visual structure may later become relevant through learning, interaction, or
changes in the task. If the initial representation retains only structures
defined by an existing category set, an external prompt, or a single
partitioning criterion, other organization present in the sensor input may
not remain explicitly represented for subsequent processing.\\

The present work therefore considers an earlier stage of visual organization:
the formation of candidate perceptual units before object identity, semantic
category, or task-specific relevance is assigned. The goal is not to replace
task-specific recognition or segmentation, but to provide an intermediate
sensor-grounded representation that preserves currently supported visual
organization for later learning and task-dependent interpretation. For
example, an illumination-related structure extending across an object can
remain available as a separate group without requiring it either to be
interpreted semantically or absorbed into the representation of the
underlying object.\\ 

For this purpose, this work introduces \emph{Domain Parent Grouping} (DPG), a
sensor-grounded grouping method that operates on visual measurements derived
from the input image. A \emph{domain} is a processing representation that
makes a particular relationship among those measurements available for
spatial grouping, while \emph{parent grouping} retains broader groups together
with direct spatial relations to smaller groups produced in other domains.
DPG forms spatially connected candidate perceptual units without requiring
their identity or later relevance to be specified during grouping. The
intended system-level property is adaptability to visual structure that was
not anticipated when the perception system was designed: multiple
sensor-supported groups can remain available for later comparison, learning,
interaction, or task-dependent selection.\\

DPG implements three complementary sensor-grounded processing domains
representing locally contextualized luminance, direct chromatic relationships,
and contextual chromatic relationships. Each domain derives a separate
grouping representation from the same input measurements. Selected group
content can subsequently be processed again without discarding the broader
group already formed, allowing internal measurement variation to become
available as additional groups when evaluated in its own context. This
provides a mechanism for changing the observational resolution when particular
content requires more detailed processing. Here, observational resolution
refers to the level of measurement variation made available for grouping
rather than to the spatial pixel resolution of the image.\\

The paper first formulates the grouping problem and then presents the DPG
implementation and its three processing domains. The experiments qualitatively
examine overlapping groups and reprocessing, quantitatively evaluate the
contributions of the individual domains, and compare the complete DPG
configuration with conventional category-independent grouping methods.

\section{Materials and Methods}

\subsection{Problem Formulation}

Let an image be represented as a spatially ordered field of pixel
measurements

\[
I=\{I(x,y)\mid(x,y)\in\Omega\},
\]

where

\[
\Omega=\{1,\ldots,W\}\times\{1,\ldots,H\}
\]

is the set of pixel positions in the single processed image of width $W$ and
height $H$. In this work,

\[
I(x,y)\in\mathcal{X},
\qquad
\mathcal{X}=[0,1]^3,
\]

where $I(x,y)$ is the normalized RGB measurement available at pixel position
$(x,y)$. The image therefore specifies a measurement at each spatial position,
but it does not specify which positions belong to the same structure or where
one structure ends and another begins. In the segmentation task, spatial
groups must instead be identified from relationships and spatial organization
among these pixel measurements.\\

The required output is a family of spatially connected groups

\[
\mathcal{G}(I)=\{G_1,G_2,\ldots,G_n\},
\qquad
G_i\subseteq\Omega.
\]

Each $G_i$ is a set of pixel positions whose measurements and spatial
organization support their inclusion in a common group. The output is therefore
defined in terms of measurement-supported spatial structure rather than
predefined interpretations of that structure. No object identity, surface
class, shape category, or target boundary needs to be assigned to any $G_i$ in
advance.\\

Determining these groups is not equivalent to comparing individual pixel
measurements. Local measurement similarity or boundary evidence alone may be
insufficient to determine spatial organization
\cite{Martin2004,Elder2018Global}. Measurements belonging to one structure may
vary substantially because of illumination, shading, texture, and specular
reflection \cite{Vazquez2011}, while measurements belonging to different
structures may be only weakly separated. Thus, similar measurements need not
belong to the same group, and different measurements need not belong to
separate groups.\\

The segmentation problem is consequently to determine which measurement
variation can be supported as internal to a spatial group and when the
relations among the measurements instead support the formation of a separate
group. This decision concerns the structure supported by the measurements, different
measurement relationships may therefore support different groups containing
some of the same pixel positions. Thereby the output is not required to form one
mutually exclusive partition of the image, and distinct groups may overlap:

\[
G_i\cap G_j\neq\varnothing,
\qquad i\neq j.
\]

The grouping supported by the measurements is also relative to the spatial
extent and measurement range within which those relationships are evaluated.
Let a retained group $G_k\in\mathcal{G}(I)$ define a selected set of valid
pixel positions

\[
S=G_k,
\qquad S\subseteq\Omega,
\]

and let $I|_S$ denote the original image measurements restricted to those
positions. When $I|_S$ is processed, only measurements belonging to $S$
contribute to the evaluated measurement range and spatial context. The
restricted processing may therefore produce a new group

\[
G_{\mathrm{new}}\in\mathcal{G}(I|_S),
\qquad
G_{\mathrm{new}}\notin\mathcal{G}(I).
\]

DPG finds these structures through a set of measurement-processing domains

\[
\mathcal{D}=\{L,VC,CC\},
\]

representing the luminance, vector-colour, and colour-consistency domains. Each
domain $d\in\mathcal{D}$ evaluates a different relationship among the
measurements and produces a domain-specific family of spatial groups

\[
\mathcal{G}_{d}(I)
=
\{G_{d,1},G_{d,2},\ldots\}.
\]

The domain-specific families provide alternative groupings of the same
spatially ordered input measurements. DPG compares the groups across domains
and retains groups that receive sufficient spatial support from groups
produced in another domain. The final DPG output therefore satisfies

\[
\mathcal{G}_{\mathrm{DPG}}(I)
\subseteq
\bigcup_{d\in\mathcal{D}}\mathcal{G}_{d}(I).
\]

The precise measurement transformations, spatial grouping procedure, and
cross-domain support criterion are defined in the following sections.\\

Each retained group is represented by a mask of spatially connected pixel
positions. Its boundary is determined by the spatial extent of that mask and
is therefore a consequence of the resulting group rather than the primary
entity being detected. A retained group $G_k$ may subsequently be selected as
the new processing extent $S$, allowing additional groups supported within
that restricted context to be formed and retained alongside the original
groups. DPG can consequently represent overlapping spatial organization at
different grouping resolutions without requiring a predefined semantic
interpretation.

\subsection{Method}

During method development, these processing principles were formulated from conclusions drawn from controlled visual experiments performed with purpose-built tools that allow systematic evaluation of how distinguishable structure changes as a function of relationships among the available measurements \cite{Saukkio2026ColorIllusions}. These
established configurations provided simplified cases for analysing how a
specified change in image measurements changes the spatial structure available
for grouping. The analysis was conducted at the computational level of image
measurements, their relationships, and the resulting grouping structure. The
identified dependencies were used to define the processing domains and to
select the transformations performed within them. \\

The numerical parameters used by DPG were selected empirically through
iterative qualitative inspection of a heterogeneous collection of development
images. Each domain was examined separately, and its parameters were adjusted
so that visually distinguishable luminance or chromatic structures were
retained consistently across the collection rather than optimized for any
individual image. None of the images used in the statistical or qualitative
evaluations were included in this parameter-selection material.\\

The formulation below represents the adjustable quantities symbolically.
These include parameters governing local contextual processing, measurement
classification and representational resolution, minimum group size, and
cross-domain relations. The fixed values used in the experiments are reported
in Section \ref{subsec:implementation_and_execution_enviroment}.\\

At the image boundaries, the neighbourhood specified below is extended by reflection. The same boundary treatment is used for all local neighbourhood operations used in this work.

\subsubsection{Luminance-Domain Processing}
\label{subsec: luminance}
The luminance domain is intended to reduce minor variation within locally
coherent regions while preserving or strengthening measurements that are
distinct relative to their immediate surroundings. The algorithm receives an
RGB image whose channel values are normalized to the range $[0,1]$. No
separate radiometric linearization is applied.

A weighted luminance field is defined as

\begin{equation*}
Y(x,y)
=
\alpha_R R(x,y)
+
\alpha_G G(x,y)
+
\alpha_B B(x,y),
\end{equation*}

where $R(x,y)$, $G(x,y)$, and $B(x,y)$ denote the normalized red, green,
and blue channel measurements at image location $(x,y)$, and
$\alpha_R$, $\alpha_G$, and $\alpha_B$ are the luminance-channel weights. \\

For each pixel, the local mean $\mu_Y(x,y)$ and local standard deviation
$\sigma_Y(x,y)$ are calculated over a square $n_Y\times n_Y$
neighbourhood. Let $\mathcal{N}_Y(x,y)$ denote this neighbourhood centred
at $(x,y)$. The local mean is

\begin{equation*}
\mu_Y(x,y)
=
\frac{1}{|\mathcal{N}_Y(x,y)|}
\sum_{(p,q)\in\mathcal{N}_Y(x,y)}
Y(p,q).
\end{equation*}

The local variance is obtained as

\begin{equation*}
\sigma_Y^2(x,y)
=
\operatorname{mean}_{\mathcal{N}_Y(x,y)}
\left(Y^2\right)
-
\mu_Y^2(x,y).
\end{equation*}

The standard deviation is then

\begin{equation*}
\sigma_Y(x,y)
=
\sqrt{
\max\left(
\sigma_Y^2(x,y),
0
\right)
}.
\end{equation*}

A small value of $\sigma_Y$ indicates a locally uniform region, whereas a
large value indicates substantial luminance variation within the
neighbourhood. \\

The signed deviation of a pixel from its local mean is defined as

\begin{equation*}
d_Y(x,y)
=
Y(x,y)-\mu_Y(x,y).
\end{equation*}

A positive value means that the pixel is brighter than the local mean, while a negative value means that it is darker. The magnitude $\lvert d_Y(x,y)\rvert$ describes how strongly the pixel differs from its surroundings.\\

The deviation is evaluated using an adaptive local threshold

\begin{equation*}
\tau_Y(x,y)
=
\alpha_{\tau}
+
\beta_{\tau}\sigma_Y(x,y),
\end{equation*}

where $\alpha_{\tau}$ defines the baseline deviation threshold and
$\beta_{\tau}$ determines how strongly the threshold increases with local
luminance variation.

The threshold is therefore smaller in locally uniform regions and larger in regions that  already contain substantial variation. Consequently, a relatively small deviation may be significant in a uniform region, whereas a larger deviation is required in a textured or otherwise heterogeneous region.\\

A separate neighbourhood-influence term determines how strongly the local
context is allowed to modify the pixel:

\begin{equation*}
w_Y(x,y)
=
\exp\left(
-\frac{\sigma_Y(x,y)}{\lambda_Y}
\right),
\end{equation*}

where $\lambda_Y$ controls the rate at which the neighbourhood influence
decreases as local luminance variation increases.

The value of $w_Y$ approaches one in uniform regions and decreases as local variation increases. The adaptive threshold and the neighbourhood-influence term therefore serve different purposes. The threshold determines whether a pixel is treated as locally similar or locally distinct, while the influence term determines the strength of the resulting modification.\\

When the deviation is smaller than the adaptive threshold,

\begin{equation*}
\lvert d_Y(x,y)\rvert
<
\tau_Y(x,y),
\end{equation*}

the pixel is treated as belonging to the prevailing local luminance region.
Its value is moved towards the local mean according to

\begin{equation*}
Y'(x,y)
=
Y(x,y)
+
\gamma_{\mathrm{assim}}
w_Y(x,y)
\left(
\mu_Y(x,y)-Y(x,y)
\right),
\end{equation*}

where $\gamma_{\mathrm{assim}}$ determines the strength of the movement
towards the local mean.\\

The modification is proportional to the difference between the pixel value
and the local mean. When the neighbourhood influence is maximal,
$\gamma_{\mathrm{assim}}$ determines the fraction of this difference applied
to the pixel. As local variation increases, $w_Y(x,y)$ decreases and the
modification becomes correspondingly weaker. The operation therefore reduces
minor within-region luminance variation without directly replacing the pixel
with the neighbourhood mean.\\

When the deviation reaches or exceeds the adaptive threshold,

\begin{equation*}
\lvert d_Y(x,y)\rvert
\geq
\tau_Y(x,y),
\end{equation*}

the pixel is treated as locally distinct. Its deviation from the local mean
is increased according to

\begin{equation*}
Y'(x,y)
=
Y(x,y)
+
\gamma_{\mathrm{contrast}}
w_Y(x,y)d_Y(x,y),
\end{equation*}

where $\gamma_{\mathrm{contrast}}$ determines the strength with which a
locally distinct measurement is moved further from the local mean.\\

A pixel brighter than its neighbourhood is therefore moved towards a larger luminance value, whereas a darker pixel is moved towards a smaller luminance value. The neighbourhood-influence term prevents strong amplification in regions that already contain substantial variation.\\

After either modification, the resulting value is restricted to the
normalized range:

\begin{equation*}
Y'(x,y)
\leftarrow
\min\left(
1,
\max\left(0,Y'(x,y)\right)
\right).
\end{equation*}

The processed luminance field is subsequently discretized relative to the
luminance range present in the current input. Let

\begin{equation*}
Y'_{\min}
=
\min_{x,y} Y'(x,y)
\end{equation*}

and

\begin{equation*}
Y'_{\max}
=
\max_{x,y} Y'(x,y)
\end{equation*}

denote the minimum and maximum values of the processed luminance field. Each
processed value is mapped to an input-relative luminance coordinate

\begin{equation*}
\widehat{Y}(x,y)
=
\frac{
Y'(x,y)-Y'_{\min}
}{
\max\left(
Y'_{\max}-Y'_{\min},
\varepsilon
\right)
},
\end{equation*}

where $\varepsilon=10^{-8}$ prevents division by zero. The resulting coordinate $\widehat{Y}(x,y)$ lies in the normalized range $[0,1]$ and expresses the pixel luminance relative to the processed luminance range of the current input rather than relative to the complete normalized range of the RGB-derived luminance field.\\

The relative luminance coordinate is divided into $K_L$ equally wide
intervals. Equivalently, the corresponding boundaries in the processed
luminance field are

\begin{equation*}
b_k
=
Y'_{\min}
+
\frac{k}{K_L}
\left(
Y'_{\max}-Y'_{\min}
\right),
\qquad
k=0,\ldots,K_L.
\end{equation*}

The number and relative widths of the intervals remain fixed within one
parameterization, while their corresponding luminance values are determined
separately for each processed input.\\

The discretization is therefore input-relative but spatially global. A single set of $K_L+1$ boundaries is used throughout one processing input, while the set is recalculated whenever a new image or selected group is processed. This allows the available luminance resolution to adapt to dark, bright, or restricted-range inputs. For example, when a selected parent group occupies only a narrow interval of the original image luminance range, reprocessing expands that interval across the available $K_L$ luminance classes and can therefore expose internal luminance structure that was previously assigned to the same class.\\

This adaptation does not modify the processed luminance field used by the preceding contextual operations. It changes only the coordinate used for the final input-relative class assignment. Consequently, the local mean, standard deviation, deviation threshold, and neighbourhood influence remain defined from the input luminance field, while the final group classes are formed relative to the luminance variation available at the current
processing resolution.

\subsubsection{Colour-Domain Processing}

The colour processing produces two complementary class maps using a shared
pixel-classification procedure. The vector-colour domain applies this
classification directly to the normalized RGB input. The colour-consistency
domain first constructs a globally compensated chromatic representation and
then applies the same classification procedure to the modified RGB values.
In both domains, the processed luminance field $Y'$ is used to classify
achromatic measurements according to their luminance.

\paragraph{Vector-colour-domain processing}

For each pixel, the relative difference between its RGB channels is calculated as

\begin{equation*}
S_{\mathrm{RGB}}(x,y)
=
\frac{
\max\left(
R(x,y),G(x,y),B(x,y)
\right)
-
\min\left(
R(x,y),G(x,y),B(x,y)
\right)
}{
\max\left(
R(x,y)+G(x,y)+B(x,y),
\varepsilon
\right)
},
\end{equation*}

where $\varepsilon$ is the numerical safeguard defined in \ref{subsec: luminance}. The numerator gives the difference between the strongest and weakest
channel, while the denominator expresses this difference relative to the
combined RGB magnitude. Except near zero combined intensity or when channel clipping occurs, the measure is invariant to a common positive scaling of all three channels.\\

A pixel is treated as achromatic when

\begin{equation*}
S_{\mathrm{RGB}}(x,y)
<
\tau_{\mathrm{ach}},
\end{equation*}

where $\tau_{\mathrm{ach}}$ is the achromaticity threshold.

The processed luminance field is then used to divide the achromatic pixels
into three classes:

\begin{equation*}
c_{\mathrm{ach}}(x,y)
=
\begin{cases}
\text{black},
&
Y'(x,y)<\tau_{\mathrm{dark}},
\\
\text{grey},
&
\tau_{\mathrm{dark}}
\leq
Y'(x,y)
\leq
\tau_{\mathrm{light}},
\\
\text{white},
&
Y'(x,y)>\tau_{\mathrm{light}}.
\end{cases}
\end{equation*}

Here, $\tau_{\mathrm{dark}}$ and $\tau_{\mathrm{light}}$ define the
luminance boundaries between the three achromatic classes.

Thus, luminance processing affects the transition between black, grey, and white, but it does not determine the class of a chromatic pixel.\\

Pixels satisfying

\begin{equation*}
S_{\mathrm{RGB}}(x,y)
\geq
\tau_{\mathrm{ach}}
\end{equation*}

are treated as chromatic. Their colour direction is derived directly from
two RGB channel relations:

\begin{equation*}
u(x,y)
=
R(x,y)-G(x,y),
\end{equation*}

and

\begin{equation*}
v(x,y)
=
\frac{R(x,y)+G(x,y)}{2}
-
B(x,y).
\end{equation*}

The first relation describes the red-green balance. The second compares the combined red and green contribution with the blue contribution. Together, they specify a direction in a two-dimensional colour plane:

\begin{equation*}
\theta(x,y)
=
\operatorname{atan2}
\left(
v(x,y),
u(x,y)
\right).
\end{equation*}

The angle is mapped to the interval

\begin{equation*}
\theta(x,y)\in[0,2\pi).
\end{equation*}

The angular range is divided into $K_C$ globally fixed sectors of equal
width:

\begin{equation*}
\Delta\theta
=
\frac{2\pi}{K_C}.
\end{equation*}

A chromatic pixel is assigned to class
$j\in\{0,\ldots,K_C-1\}$ when

\begin{equation*}
\frac{2\pi j}{K_C}
\leq
\theta(x,y)
<
\frac{2\pi(j+1)}{K_C}.
\end{equation*}

The vector-colour domain therefore produces three achromatic
classes and $K_C$ chromatic direction classes. The same thresholds and
sector boundaries are used throughout one processing input.\\

The direct channel relations above define the angular coordinate used for
the final chromatic class assignment. The colour-consistency compensation
described below uses an orthonormal chromatic basis to construct an RGB
displacement vector. After compensation, the modified RGB values are
classified using the same direct channel relations as in the vector-colour
domain.

\paragraph{Colour-consistency-domain processing}

The colour-consistency domain was motivated by contextual colour
demonstrations such as Kitaoka's apparent-reddish-strawberries stimulus
\cite{Kitaoka2015}. In an analysis of a closely related stimulus, Shapiro
et al.\ showed that transformations of spatial and statistical information
already present in the image can make the apparent object colour explicit
\cite{Shapiro2018}. DPG uses this as a computational design case for
constructing an additional chromatic representation from relationships
already present in the input measurements.\\

The colour-consistency domain estimates the strongest direction-specific
mean chromatic displacement in the current input and applies a global
compensation in the opposite direction. This reduces the dominant shared
chromatic tendency and makes residual chromatic relationships that are
weakly represented in the uncompensated vector-colour representation
available for classification.\\

For computational efficiency, the chromatic displacement estimate is
evaluated using a spatial sampling stride $s_K$ in both image directions. For each sampled pixel, two orthogonal chromatic components are
calculated directly from its RGB values:

\begin{equation*}
u_K(x,y)
=
\frac{
R(x,y)-G(x,y)
}{
\sqrt{2}
},
\end{equation*}

and

\begin{equation*}
v_K(x,y)
=
\frac{
R(x,y)+G(x,y)-2B(x,y)
}{
\sqrt{6}
}.
\end{equation*}

These components contain no common achromatic RGB contribution. A pixel with
equal red, green, and blue values therefore gives

\begin{equation*}
u_K(x,y)=v_K(x,y)=0.
\end{equation*}

For each sampled pixel, the chromatic direction is first obtained using $\operatorname{atan2}$ and then mapped circularly to $[0,2\pi)$:

\begin{equation*}
\theta_K(x,y)
=
\operatorname{atan2}
\left(
v_K(x,y),
u_K(x,y)
\right)
\bmod 2\pi.
\end{equation*}

and the strength of the chromatic displacement is

\begin{equation*}
r_K(x,y)
=
\sqrt{
u_K^2(x,y)+v_K^2(x,y)
}.
\end{equation*}

The complete angular range is divided into $K_D$ equal direction bins. A
sampled pixel is assigned to bin $k\in\{0,\ldots,K_D-1\}$ when

\begin{equation*}
\frac{2\pi k}{K_D}
\leq
\theta_K(x,y)
<
\frac{2\pi(k+1)}{K_D}.
\end{equation*}

For each bin $k$, the mean chromatic magnitude of the sampled pixels assigned
to that bin is calculated:

\begin{equation*}
L(k)
=
\frac{
\displaystyle
\sum_{(x,y)\in\mathcal{B}_k}
r_K(x,y)
}{
\max\left(
\lvert\mathcal{B}_k\rvert,
1
\right)
},
\end{equation*}

where $\mathcal{B}_k$ is the set of sampled pixels whose chromatic direction
belongs to bin $k$. Thus, $L(k)$ describes the mean strength of the chromatic displacement among the sampled pixels assigned to direction bin $k$. It does not directly represent how frequently that direction occurs in the input.\\

The direction-specific mean magnitudes are smoothed circularly using a
symmetric five-bin kernel
$\mathbf{h}=(h_{-2},h_{-1},h_0,h_1,h_2)$:

\begin{equation*}
L_s(k)
=
\frac{
\displaystyle
\sum_{i=-2}^{2}
h_i L(k+i)
}{
\displaystyle
\sum_{i=-2}^{2} h_i
}.
\end{equation*}

The bin indices wrap circularly between the first and last direction bins. The direction with the largest smoothed mean chromatic magnitude is selected as

\begin{equation*}
k_{\mathrm{dom}}
=
\underset{k}{\operatorname{arg\,max}}
\;L_s(k).
\end{equation*}

For an even number of direction bins, the opposite direction is

\begin{equation*}
k_{\mathrm{opp}}
=
\left(
k_{\mathrm{dom}}
+
\frac{K_D}{2}
\right)
\bmod K_D.
\end{equation*}

The compensation magnitude is determined from the difference between the
dominant direction and its opposite direction:

\begin{equation*}
r_{\mathrm{comp}}
=
\max\left(
0,
\gamma_K
\left[
L_s(k_{\mathrm{dom}})
-
L_s(k_{\mathrm{opp}})
\right]
\right),
\end{equation*}

where $\gamma_K$ determines the strength of the chromatic compensation.\\

The resulting difference measures the imbalance between the two opposing chromatic directions. A positive difference indicates an excess in the dominant direction and
determines the compensation magnitude, whereas a zero or negative difference
produces no compensation.\\

The centre angle of the selected bin is

\begin{equation*}
\phi_{\mathrm{dom}}
=
\frac{2\pi}{K_D}
\left(
k_{\mathrm{dom}}+\frac{1}{2}
\right).
\end{equation*}

The compensation direction is the opposite angular direction:

\begin{equation*}
\phi_{\mathrm{comp}}
=
\left(
\phi_{\mathrm{dom}}+\pi
\right)
\bmod 2\pi.
\end{equation*}

Using the orthonormal RGB chromatic axes

\begin{equation*}
\mathbf{E}_A
=
\frac{1}{\sqrt{2}}
\begin{bmatrix}
1\\
-1\\
0
\end{bmatrix},
\end{equation*}

and

\begin{equation*}
\mathbf{E}_B
=
\frac{1}{\sqrt{6}}
\begin{bmatrix}
1\\
1\\
-2
\end{bmatrix},
\end{equation*}

the compensation is converted into an RGB displacement vector:

\begin{equation*}
\Delta\mathbf{RGB}
=
r_{\mathrm{comp}}
\cos\left(\phi_{\mathrm{comp}}\right)
\mathbf{E}_A
+
r_{\mathrm{comp}}
\sin\left(\phi_{\mathrm{comp}}\right)
\mathbf{E}_B.
\end{equation*}

The same displacement is added to every pixel in the original image:

\begin{equation*}
\mathbf{RGB}_K(x,y)
=
\operatorname{clip}_{[0,1]}
\left(
\mathbf{RGB}(x,y)
+
\Delta\mathbf{RGB}
\right).
\end{equation*}

The compensation is therefore global: its direction and magnitude are
identical at every image location. Clipping keeps all compensated channel
values within the normalized RGB range.\\

After the compensation vector has been added and the resulting RGB values
have been clipped to $[0,1]$, the modified input is classified using the same
procedure as in the vector-colour domain. Achromaticity and chromatic
direction are determined from the compensated RGB values. Pixels classified as chromatic are assigned to one of the same $K_C$
chromatic direction classes. For
pixels classified as achromatic, the original processed luminance field $Y'$
is used to distinguish black, grey, and white. The chromatic compensation
therefore does not cause the luminance field to be recalculated. 

\subsubsection{Spatial Grouping and Cross-Domain Parent Relations}
\label{subsec:spatial_grouping}
Each processing domain produces a discrete pixel-class map. The luminance
domain contains $K_L$ luminance classes, while each colour domain contains
three achromatic classes and $K_C$ chromatic direction classes.\\

Spatial groups are formed independently within each class. Let
$\mathcal{P}_c$ denote the set of pixels assigned to class $c$. A group is
defined as a maximal connected subset of $\mathcal{P}_c$ under the selected
pixel-neighbourhood connectivity $\mathcal{C}$. Pixels assigned to the same
class therefore form the same group only when they are connected through
$\mathcal{C}$; spatially separated regions remain distinct groups even when
they belong to the same measurement class.\\

To reduce computational cost, connected components smaller than the relevant
domain-specific minimum area are rejected before the final group maps are
retained. Their pixels remain unassigned rather than being merged into a
neighbouring group.

Let $N_{\mathrm{valid}}$ denote the number of valid input pixels in the
current processing pass. For a complete image,
$N_{\mathrm{valid}}=HW$, where $H$ and $W$ are the image height and width.
During masked reprocessing, $N_{\mathrm{valid}}$ equals the number of valid
pixels belonging to the selected input mask.

For the luminance domain, the minimum accepted area is

\begin{equation*}
A_{\min}^{L}
=
\max\left(
A_0^L,
\left\lfloor
\frac{N_{\mathrm{valid}}}{\eta_L}
\right\rfloor
\right),
\end{equation*}

whereas for the vector-colour and colour-consistency domains it is

\begin{equation*}
A_{\min}^{C}
=
\max\left(
A_0^C,
\left\lfloor
\frac{N_{\mathrm{valid}}}{\eta_C}
\right\rfloor
\right).
\end{equation*}

Each retained connected component receives an independent group identifier. The resulting group maps are compared across domains. \\

Let $G_i$ and $G_j$ denote two groups originating from different domains, and $\left|G_k
\right|$ the spatial size of a group in pixels. Their spatial overlap is defined as

\begin{equation*}
A_{ij}
=
\left|
G_i\cap G_j
\right|.
\end{equation*}

A cross-domain relation is retained when the spatial overlap relative to the
smaller group satisfies

\begin{equation*}
\frac{
A_{ij}
}{
\min\left(
\lvert G_i\rvert,
\lvert G_j\rvert
\right)
}
\geq
\rho_{\min},
\end{equation*}

where $\rho_{\min}$ is the minimum required cross-domain overlap ratio.

When the sufficiently overlapping groups have different areas, the larger
group is assigned as the parent and the smaller group as its child. Thus,

\begin{equation*}
\lvert G_i\rvert
>
\lvert G_j\rvert
\quad
\Longrightarrow
\quad
G_i
\text{ is a parent of }
G_j,
\end{equation*}

whereas

\begin{equation*}
\lvert G_j\rvert
>
\lvert G_i\rvert
\quad
\Longrightarrow
\quad
G_j
\text{ is a parent of }
G_i.
\end{equation*}

No parent-child relation is created when the two groups have exactly equal
areas. The relative overlap requirement limits relations arising from small
incidental intersections while allowing the larger group to extend beyond
the spatial extent shared with the smaller group.\\

The parent and child terms refer only to this direct cross-domain
area-and-overlap relation. A child therefore represents a more
localized group that participates in the spatial organization associated
with the parent, but it is not an exclusive subdivision of
the parent and a parent may be related to multiple children, and a child may have multiple parents. A group is designated as a parent when it acts as the larger group in
at least one accepted parent-child relation. Child groups and groups without
an accepted parent-child relation remain available in the domain-specific
representation, but do not participate in the final output.

\subsubsection{Reprocessing of Selected Group Content}

DPG initially organizes the input relative to the measurement range and
spatial context of the complete observation. Measurement variation that is
weak in relation to the complete input may therefore remain internal to a
broader group. Reprocessing allows selected content to be evaluated again in
its own context without requiring the initial processing pass to expose every
available level of measurement variation simultaneously. It therefore
provides a mechanism for adjusting observational resolution while preserving
the broader groups already formed.\\

A retained parent group is selected using its mask, and the smallest
axis-aligned rectangle containing the mask defines the new processing area.
Only the original RGB measurements belonging to the selected mask are treated
as valid input; pixels inside the rectangle but outside the mask are excluded.
The selected measurements are then processed again using the same processing
domains and fixed method parameters. Because the input-relative ranges and
contextual relationships are recalculated from the selected content,
measurement variation that remained internal to the original parent may
support additional groups in the new processing pass.\\

The reprocessed groups are retained alongside the original parent rather than
replacing it or forming a required subdivision of it. A parent produced in a
new processing pass can itself be selected for subsequent reprocessing,
allowing observational resolution to be adjusted progressively.\\

\subsection{Experimental Methods}

The experiments comprised a qualitative analysis of the structures produced
by DPG and a quantitative evaluation on the BSDS500 dataset. The qualitative
analysis examined overlapping parent groups, their spatially related child
groups, and the additional organization exposed when selected group content
was reprocessed at a different observational resolution. The quantitative
evaluation first measured the contributions of the three processing domains
individually and in pairwise combinations, and then compared the complete DPG
configuration with three category-independent bottom-up grouping methods using
the same human-annotated BSDS500 reference regions.\\

The reported measures characterize complementary properties of the resulting
representation. Region covering and region recall quantify the availability
of groups corresponding to annotated regions. Boundary recall measures how
completely annotated transitions remain represented, whereas boundary
precision indicates how much of the produced boundary structure corresponds
to those annotations. Group count additionally characterizes the extent of
the retained proposal set.

\subsubsection{Dataset and Experimental Protocol}

The evaluation used the Berkeley Segmentation Dataset and Benchmark
(BSDS500) \cite{Arbelaez2011}. The dataset contains 500 natural images divided
into 200 training images, 100 validation images, and 200 test images. Each
image is accompanied by multiple different human annotations.\\

The human annotations provide useful references and are primarily organized around perceptually salient objects and other large image regions. Each annotation divides the image into a single set of mutually exclusive regions. It therefore does not exhaustively enumerate all visually supported subregions, overlapping organizations, illumination structures, shadows, markings, reflections, or textures.\\

\begin{figure}[!t]
    \centering

    \begin{minipage}[t]{0.49\textwidth}
        \centering
        \includegraphics[width=\linewidth]{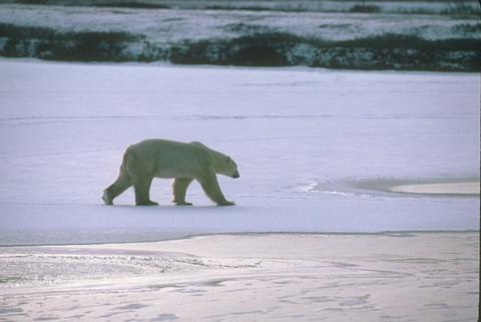}
    \end{minipage}
    \hfill
    \begin{minipage}[t]{0.49\textwidth}
        \centering
        \includegraphics[width=\linewidth]{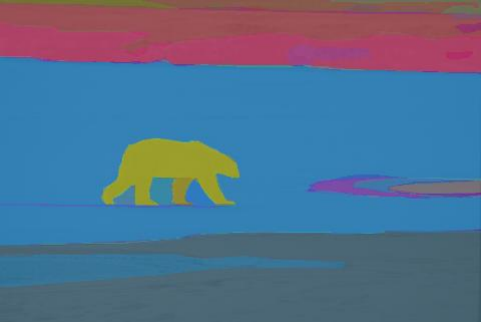}
    \end{minipage}

    \caption{Example from BSDS500. The original test image 100007 is shown on the left
    and visualization of the human-annotated segmentation is shown on
    the right. The annotation represents the principal mutually exclusive
    regions but does not enumerate all potentially overlapping or
    illumination-dependent visual structures. }
    \label{fig:example_annotation_of_dataset}
\end{figure}

An example is shown in Fig.~\ref{fig:example_annotation_of_dataset}. The human
annotation captures the principal regions, including the polar bear, water,
and surrounding ice. However, it does not represent all
illumination-dependent, shadow-related, or textured structures that may be
supported by the same sensor data.\\

BSDS500 provides a useful common reference for examining the extent to which the grouping hypotheses produced by DPG spatially correspond to the coarse scene structures identified by human observers. \\

All 200 test images were used in the statistical evaluation. All methods used identical decoded RGB arrays at the original dataset resolution. All human annotations were included on the evaluations: their regions were treated as separate references for the region metrics, their boundaries were combined for the boundary metrics, and each image constituted a single paired observation in the statistical analysis.

\subsubsection{Implementation and Execution Environment} \label{subsec:implementation_and_execution_enviroment}
The experiments were implemented in 64-bit Python 3.11 using NumPy 2.2.6,
SciPy 1.16.3, scikit-image 0.26.0, and Pillow 11.3.0. DPG and all comparison
methods were executed within the same software environment. No GPU acceleration was used.\\

The source code used to produce the DPG results and the supporting evaluation
and statistical-analysis material are publicly available as described in the
Code, Data, and Materials Availability statement.\\

The experiments were performed using the following hardware and operating
environment:

\begin{itemize}
    \item Processor Intel i7-10875H
    \item 8 CPU cores and 16 threads
    \item Ram 32 GB DDR4
    \item Operating system windows11 Version 25H2
\end{itemize}

Each image was evaluated in the same sequential execution environment.
For each conventional comparison method, runtime was measured as wall-clock
time from method entry to completion of the final group-mask list. The DPG
domain calculations were performed once per image to obtain all seven domain
configurations. The runtime reported for each DPG configuration was composed
from the measured input-preparation time, the domain calculations required by
that configuration, any required parent-group construction, and the formation
of the final proposal-mask list. Because the vector-colour and
colour-consistency calculations use the contextual luminance result, their
reported runtimes also include the required luminance calculation. File
loading, graphical visualization and metric calculation were
excluded from the reported processing time.\\

The DPG parameters were fixed before the BSDS500 test evaluation. The
input-relative luminance discretization was recalculated separately for every
input, but the number of luminance classes and the remaining algorithm
parameters were held constant across all test images. The cross-domain parent relation used
the overlap requirement defined in the implementation section, according to
which the shared area had to cover at least half of the smaller group. Pixels
left unassigned after small-component rejection remained unassigned and were
not absorbed into the nearest retained group.\\

The fixed DPG parameter values used throughout the evaluation were:
\begin{equation*}
\begin{aligned}
\alpha_R &= 0.2126, &
\alpha_G &= 0.7152, &
\alpha_B &= 0.0722, \\
n_Y &= 15, &
\alpha_{\tau} &= 0.042, &
\beta_{\tau} &= 0.85, \\
\lambda_Y &= 0.09, &
\gamma_{\mathrm{assim}} &= 0.56, &
\gamma_{\mathrm{contrast}} &= 0.20, \\
K_L &= 6, &
& \\
\tau_{\mathrm{ach}} &= 0.060, &
\tau_{\mathrm{dark}} &= 0.13, &
\tau_{\mathrm{light}} &= 0.68, \\
K_C &= 12, &
s_K &= 2, &
K_D &= 360, \\
\mathbf{h} &= (1,2,3,2,1), &
\gamma_K &= 0.75, \\
A_0^L &= 65, &
\eta_L &= 3500, \\
A_0^C &= 45, &
\eta_C &= 3200, \\
\rho_{\min} &= 0.5, &
\mathcal{C} &= \text{8-neighbour}.
\end{aligned}
\end{equation*}

For a single-domain configurations, the evaluated proposal set consisted of all
non-empty retained groups produced in that domain. For a two-domain
configurations, the cross-domain parent relations were reconstructed using only
the two selected domain maps, and the resulting parent groups formed the
evaluated proposal set. If no parent relation was produced for a two-domain
configuration, the non-empty groups from both selected domains were retained
as a fallback so that the configuration did not produce an artificial empty
proposal set. For the complete three-domain configuration, the evaluated
proposal set consisted of the parent groups reconstructed from all three
domain maps, as defined in the section \ref{subsec:spatial_grouping}.

\subsubsection{Comparison Methods and Parameter Selection}
\label{sec:comparison-methods}
DPG was compared with three category-independent bottom-up grouping methods:

\begin{itemize}
    \item \textbf{Felzenszwalb graph-based segmentation}
    \cite{Felzenszwalb2004}, which incrementally merges adjacent image
    components by comparing their internal variation with the difference
    between the components.

    \item \textbf{Quickshift mode-seeking segmentation}
    \cite{Vedaldi2008}, which groups pixels around local density modes in a joint
    colour-position feature space and provides a grouping principle distinct
    from adjacency-based component merging and boundary-driven region formation.
    The implementation converted the RGB input to CIELAB internally, used no
    additional Gaussian smoothing, and used the fixed random seed 42.
    
    \item \textbf{Watershed with boundary-RAG merging}, implemented using the
corresponding scikit-image operations \cite{VanDerWalt2014}, first forms an
input-derived marker-controlled watershed partition from the Sobel luminance
gradient \cite{Meyer1994} and subsequently merges adjacent regions using a
boundary-weighted region-adjacency graph \cite{Tremeau2000}. The selected
merge threshold produces one conventional mutually exclusive partition for
evaluation.
\end{itemize}

The comparison methods were selected to represent three different grouping principles. Felzenszwalb segmentation provides a graph-based bottom-up approach. The watershed-RAG configuration provides a boundary-oriented region-based
baseline. Quickshift groups pixels around local density modes in a joint colour-position feature space and consequently produces a characteristically different partition of the same input. Its inclusion therefore provides an informative contrast for the evaluation.\\

The parameters of the comparison methods were selected using only the 100
BSDS500 validation images. Once selected, the parameters were stored and used
without modification for the complete test-set evaluation and all
single-image visualizations.\\

For Felzenszwalb segmentation, the parameter search was calibrated towards
the published BSDS500 values of segmentation covering $0.52$, probabilistic
Rand index (PRI) $0.80$, and variation of information (VI) $2.21$
\cite{Xia2017}. These values were used as calibration targets for the present
implementation rather than as values that the implementation was expected to
reproduce exactly. During parameter search candidate configurations were ranked by the summed squared relative deviation from the three target values.
The selected configuration was

\begin{equation*}
\texttt{scale}=400,\qquad
\texttt{sigma}=1.0,\qquad
\texttt{min size}=100.
\end{equation*}

On the validation set, this configuration produced covering $0.515$, PRI
$0.770$, and VI $2.427$. \\

No directly corresponding published BSDS500 target values were available for
the exact Quickshift implementation or the watershed-RAG composition used
here. Their parameters were therefore selected by maximizing segmentation
covering on the validation set. PRI and VI were used only as
deterministic secondary criteria when candidate covering values were
effectively similar. The resulting Quickshift parameters were

\begin{equation*}
\texttt{kernel size}=7,\qquad
\texttt{max dist}=20,\qquad
\texttt{ratio}=0.5,
\end{equation*}

and the selected watershed-RAG parameters were

\begin{equation*}
\texttt{minima depth}=0.04,\qquad
\texttt{merge threshold}=0.075.
\end{equation*}

\subsubsection{Evaluation Metrics}

Because DPG produces a set of potentially overlapping masks rather than one
exclusive image partition, the quantitative evaluation was formulated as a
proposal-to-annotation comparison. All metrics were first calculated
separately for each test image.\\

For image $i$, let $\mathcal{G}_i$ denote the set of human-annotated regions
collected from all available annotations of that image, with each annotated
region treated as a separate reference. Let $\mathcal{P}_i$ denote the set of
group proposals produced by the evaluated method for the same image. The
spatial correspondence between an annotated region $G\in\mathcal{G}_i$ and a
proposal $P\in\mathcal{P}_i$ was measured using \emph{intersection over union}:

\begin{equation*}
\operatorname{IoU}(G,P)
=
\frac{|G\cap P|}{|G\cup P|}.
\end{equation*}

\paragraph{Region covering.}
For each test image, region covering was calculated as the
annotated-region-area-weighted mean of the best proposal IoU obtained for
each reference region:

\begin{equation*}
\operatorname{Covering}_i
=
\frac{
\displaystyle
\sum_{G\in\mathcal{G}_i}
|G|
\max_{P\in\mathcal{P}_i}
\operatorname{IoU}(G,P)
}{
\displaystyle
\sum_{G\in\mathcal{G}_i}|G|
}.
\end{equation*}

Thus, each annotated region contributes its best available proposal match to
the image-level covering score, with larger annotated regions receiving
greater weight. One covering value is obtained for each test image.

\paragraph{Region recall.}
For each test image, region recall was the proportion of annotated regions for
which at least one proposal reached the selected IoU threshold $t$:

\begin{equation*}
\operatorname{Recall}_{i,t}
=
\frac{
\left|
\left\{
G\in\mathcal{G}_i
\; \middle| \;
\max_{P\in\mathcal{P}_i}
\operatorname{IoU}(G,P)
\geq t
\right\}
\right|
}{
|\mathcal{G}_i|
}.
\end{equation*}

Each annotated region contributes equally to this measure. Recall was reported
at $t=0.50$ and $t=0.75$, producing two region-recall values for each test
image.

\paragraph{Boundary precision, recall, and F1 score.}
Boundary measures were also calculated separately for each test image. A
boundary map was extracted from every proposal mask, and the boundaries of all
proposals in $\mathcal{P}_i$ were combined into one predicted boundary map for
image $i$. The boundaries from all available human annotations of the same
image were combined into one reference boundary map. Thus, unlike the region
metrics, the individual human annotations were combined before the boundary
metrics were calculated.\\

A predicted and reference boundary pixel were considered corresponding when a
boundary pixel from the other map occurred within a tolerance of $0.75\%$ of
the image diagonal, rounded to the nearest pixel with a minimum tolerance of
one pixel.\\

Let $\mathcal{B}_{P,i}$ and $\mathcal{B}_{G,i}$ denote the predicted and
reference boundary-pixel sets for image $i$, and let
$\mathcal{B}_{P,i}^{\mathrm{matched}}$ and
$\mathcal{B}_{G,i}^{\mathrm{matched}}$ denote the subsets having a
corresponding boundary within the allowed tolerance. Boundary precision and
recall are then

\begin{equation*}
P_{B,i}
=
\frac{
|\mathcal{B}_{P,i}^{\mathrm{matched}}|
}{
|\mathcal{B}_{P,i}|
},
\end{equation*}

and

\begin{equation*}
R_{B,i}
=
\frac{
|\mathcal{B}_{G,i}^{\mathrm{matched}}|
}{
|\mathcal{B}_{G,i}|
}.
\end{equation*}

Their harmonic mean is

\begin{equation*}
F_{1,B,i}
=
\frac{2P_{B,i}R_{B,i}}
{P_{B,i}+R_{B,i}}.
\end{equation*}

One boundary-precision, boundary-recall, and boundary-F1 value is therefore
obtained for each method and test image.

\paragraph{Group count, runtime, and dataset-level reporting.}
Group count was defined as the number of masks in the evaluated proposal set
for each image. Runtime was recorded per image using the execution-time
definition given in Section~\ref{subsec:implementation_and_execution_enviroment}.\\

For every quantitative metric, the method-level result reported in the paper
is the mean and sample standard deviation of the 200 image-level values.
Higher covering, region recall, boundary precision, boundary recall, and
boundary F1 indicate greater correspondence with the human annotations.
Lower runtime indicates faster execution. Group count describes the size of
the evaluated proposal set and is not treated as an accuracy measure.

\subsubsection{Paired Statistical Analysis}

The paired statistical analysis was applied to the image-level quantitative
results defined above. Each of the 200 BSDS500 test images therefore
contributed one value per evaluated method for region covering, recall at IoU
$0.50$, recall at IoU $0.75$, boundary precision, boundary recall, boundary
F1, group count, and runtime.\\

Two comparison families were analysed separately. In the domain evaluation,
the complete three-domain DPG configuration was compared with each of the
three single-domain and three two-domain configurations. In the conventional
method comparison, complete DPG was compared separately with Felzenszwalb,
Quickshift, and watershed-RAG.\\

For metric $m$, comparison method or configuration $c$, and test image $i$,
the paired difference was defined as

\begin{equation*}
d_{m,c,i}
=
m_{\mathrm{DPG},i}
-
m_{c,i}.
\end{equation*}

Positive differences therefore indicate larger DPG values and negative
differences smaller DPG values. For the region and boundary correspondence
metrics, a positive value indicates higher reference correspondence. For
runtime, a negative value indicates faster DPG execution. Group-count
differences describe only differences in proposal-set size.\\

Systematic paired differences were evaluated using two-sided Wilcoxon
signed-rank tests \cite{Wilcoxon1945}. The mean paired difference was
reported together with a percentile-based 95\% confidence interval estimated
from 10,000 paired bootstrap resamples \cite{Efron1987}, using the fixed
random seed 20260718. Matched-pairs rank-biserial correlation was reported as
the paired effect-size measure \cite{Kerby2014}.\\

Wilcoxon $p$-values were adjusted using Holm's procedure separately for each
metric and comparison family \cite{Holm1979}. The six DPG component
comparisons formed one family and the three conventional-method comparisons
formed the other. Statistical significance was assessed at the
Holm-adjusted level $\alpha=0.05$.\\

Complete paired results, including method means and sample standard
deviations, mean paired differences, bootstrap confidence intervals,
Holm-adjusted $p$-values, and rank-biserial effect sizes, are reported in
Appendix~\ref{app:statistical-results}.

\subsubsection{Qualitative Experimental Protocol}

The qualitative evaluation used two captures of the same scene under
different natural illumination conditions while keeping the camera position
and principal scene content similar. The scene contained both natural and
manufactured structures with variation in colour, luminance, surface
appearance, and spatial extent. These images were used to examine the grouping
produced by the DPG domains, the resulting parent-child and overlapping group
relations, and the effect of reprocessing selected group content.\\

The images were captured with a GoPro HERO8 camera in GPR raw format and
subsequently converted to DNG using Adobe DNG Converter. Raw capture was used
to limit the influence of in-camera image processing. The camera settings were
Output: RAW, Lens: Wide, Shutter: Auto, EV Compensation: 0, White Balance:
Native, ISO Minimum: 100, ISO Maximum: 200, Sharpness: Low, and Colour: Flat.
These settings were kept fixed between the two captures, apart from the
automatically determined exposure time.\\

\section{Results}

The qualitative examples show that DPG can retain multiple spatially
overlapping groups from the same observation and that reprocessing can expose
additional measurement-supported organization without replacing the broader
group from which it was obtained. In the BSDS500 evaluation, the complete DPG
configuration retained region correspondence comparable to conventional
bottom-up grouping while producing a higher boundary recall.\\

\subsection{Qualitative Results}

\begin{figure}[!t]
    \centering

    \subfloat[Felzenszwalb\label{fig:lighted_felzenszwalb}]{%
    \begin{minipage}[t]{0.32\textwidth}
        \centering
        \includegraphics[width=\linewidth]
        {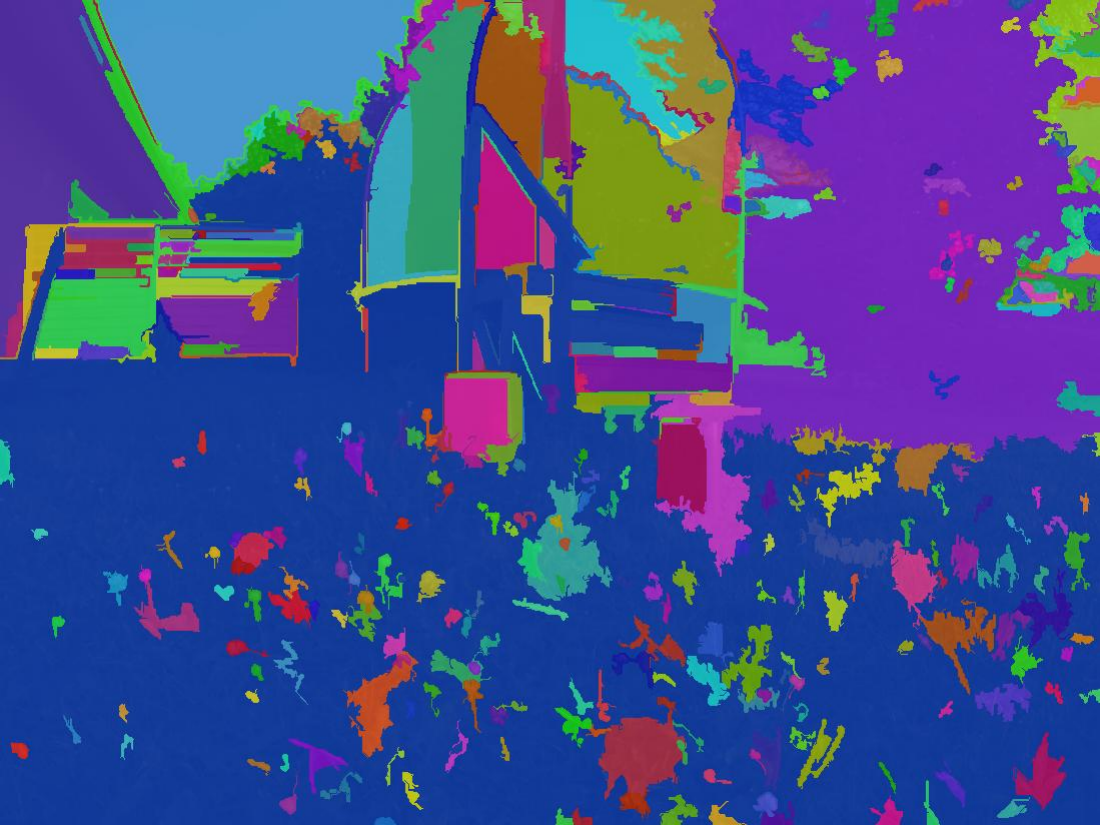}
    \end{minipage}}
    \hfill
    \subfloat[Watershed with boundary-RAG merging\label{fig:lighted_watershed}]{%
    \begin{minipage}[t]{0.32\textwidth}
        \centering
        \includegraphics[width=\linewidth]
        {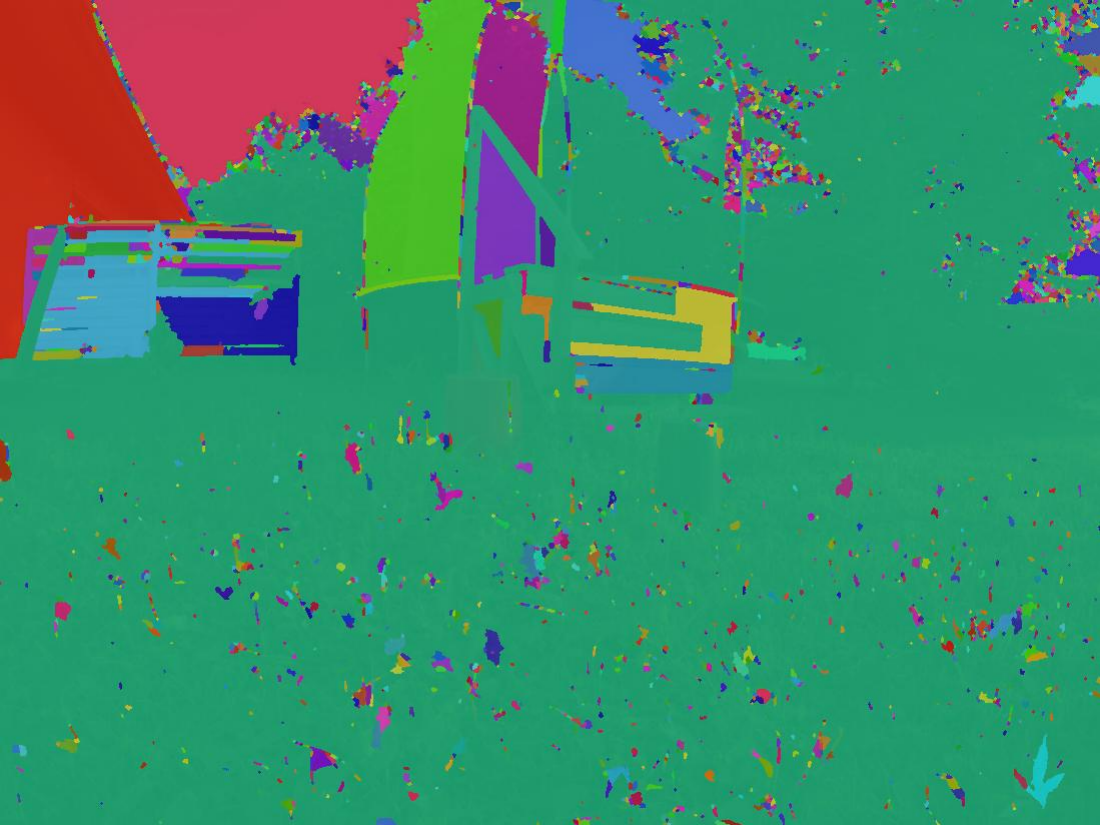}
    \end{minipage}}
    \hfill
    \subfloat[Quickshift\label{fig:lighted_quickshift}]{%
    \begin{minipage}[t]{0.32\textwidth}
        \centering
        \includegraphics[width=\linewidth]
        {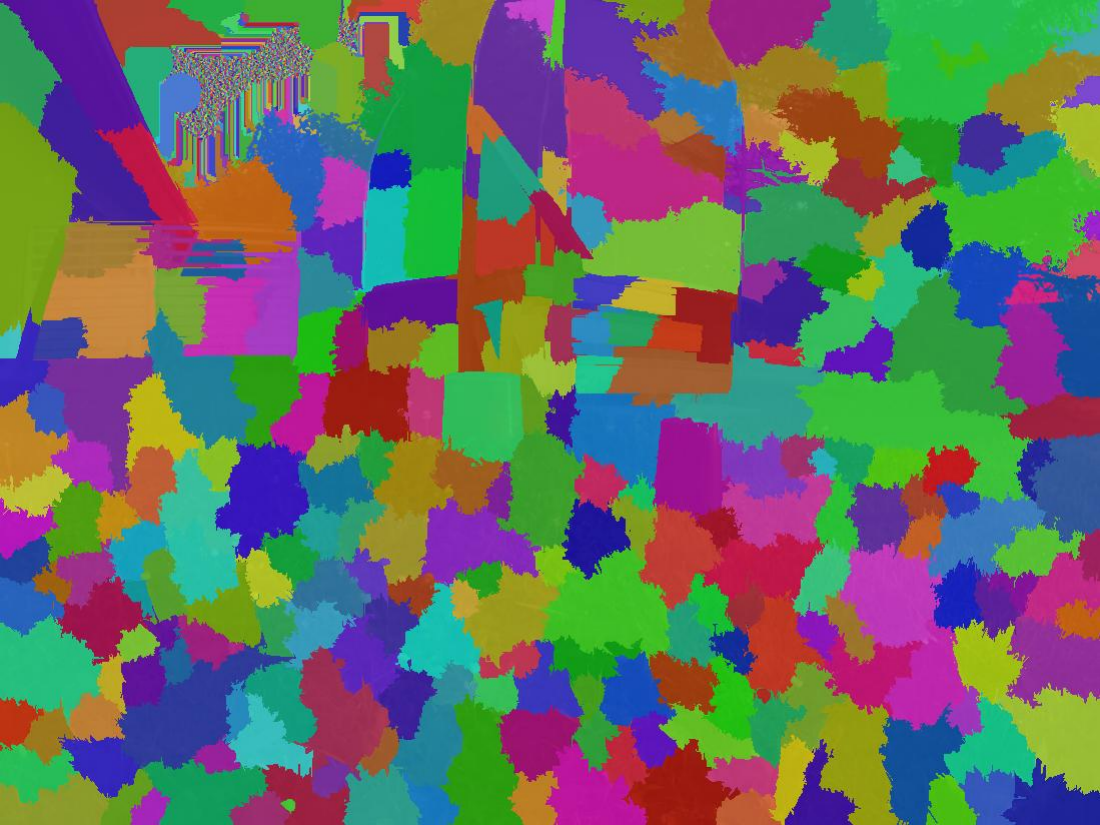}
    \end{minipage}}

    \vspace{0.6em}

    \subfloat[DPG\label{fig:lighted_dpg}]{%
    \begin{minipage}[t]{0.44\textwidth}
        \centering
        \includegraphics[width=\linewidth]
        {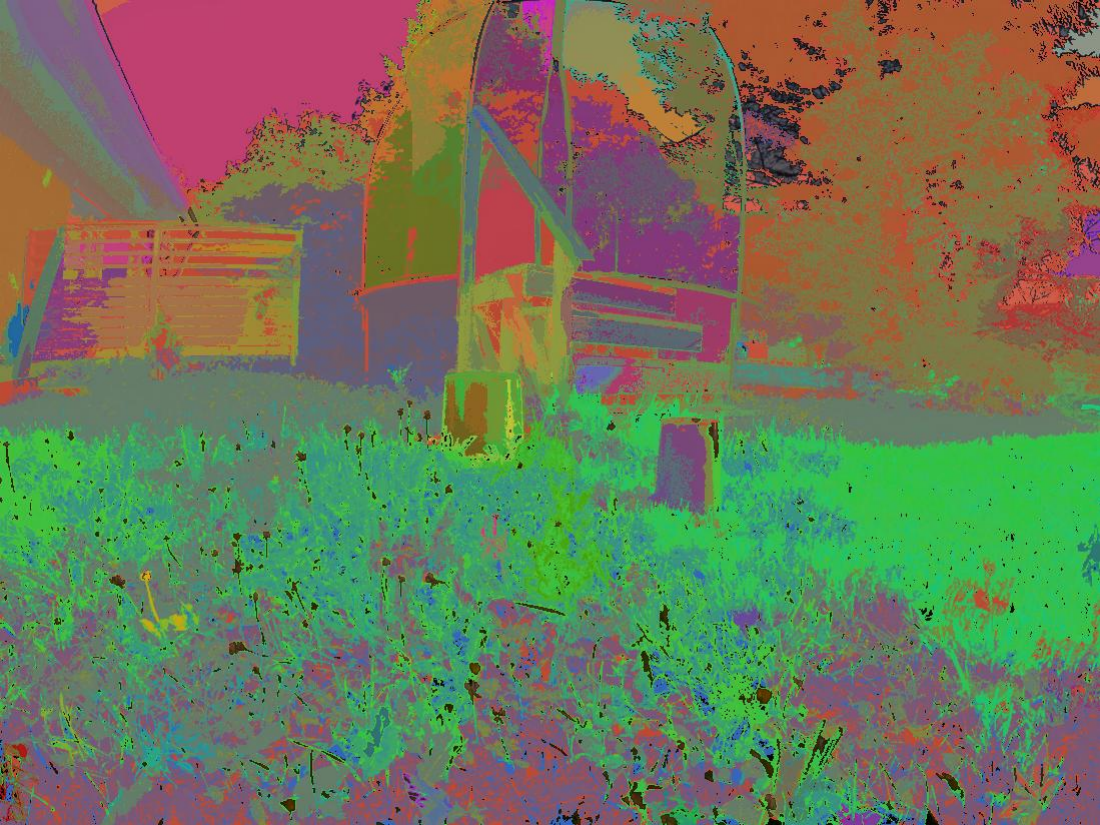}
    \end{minipage}}
    \hspace{0.04\textwidth}
    \subfloat[Original image\label{fig:lighted_original}]{%
    \begin{minipage}[t]{0.44\textwidth}
        \centering
        \includegraphics[width=\linewidth]
        {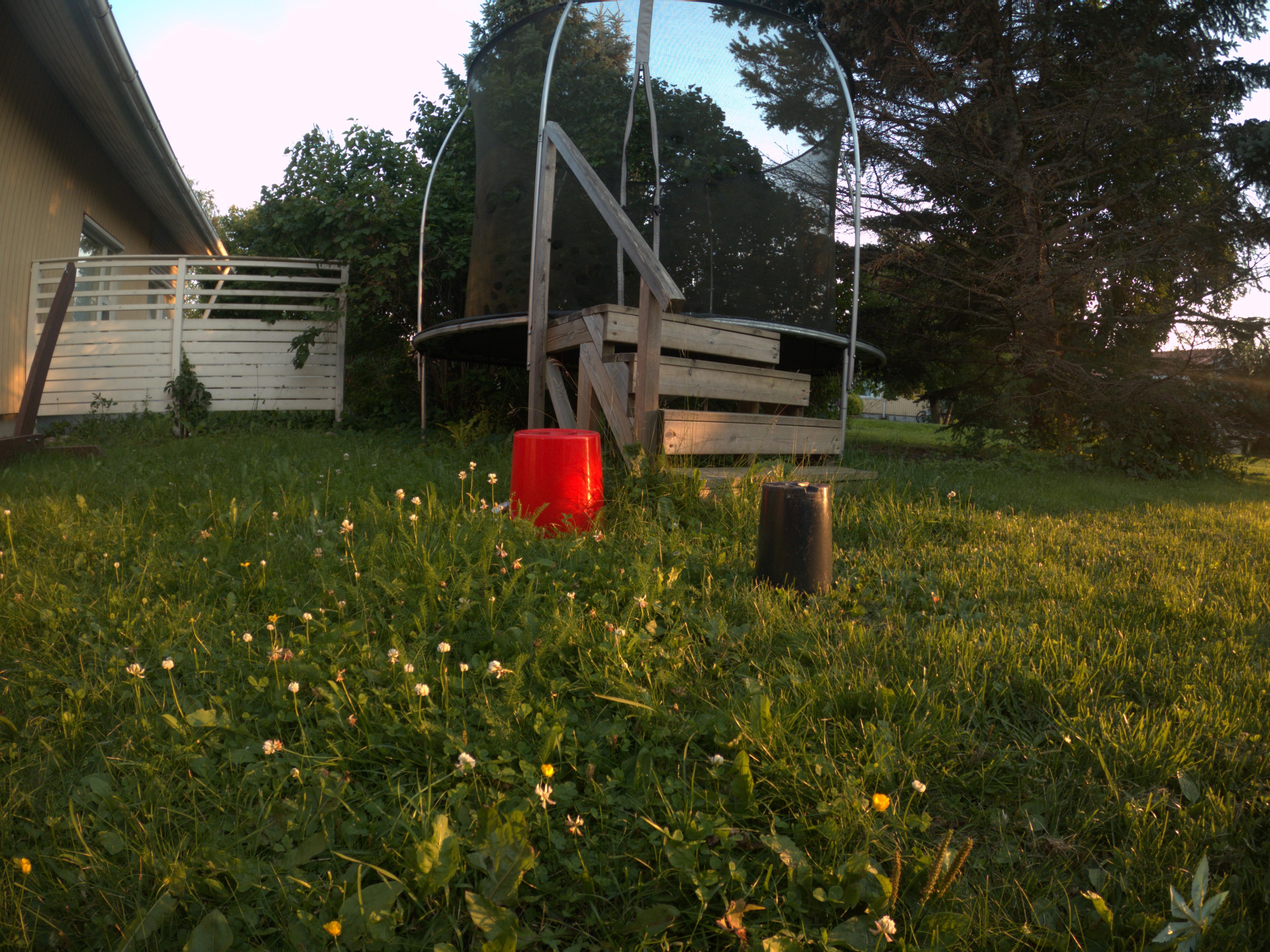}
    \end{minipage}}

    \caption{Qualitative comparison for the brighter capture, GOPR1789.
    The top row shows the mutually exclusive partitions produced by the three
    comparison methods. The bottom row shows the simplified visualization of the
    complete DPG group set and the original input image.}
    \label{fig:lighted_method_comparison}
\end{figure}

Figure~\ref{fig:lighted_method_comparison} compares the grouping results for
the brighter of the two captures. All four methods produce groups associated
with several prominent scene structures, including the building, trampoline,
containers, and vegetation. The three comparison methods represent the image
as mutually exclusive partitions. Felzenszwalb retains several distinct
structures while fragmenting parts of the detailed vegetation, Quickshift
produces a dense local partition, and watershed-RAG merges much of the
background into broader regions.\\

The DPG visualization differs because its groups may overlap. A single colour
cannot display all groups present at the same image location, so the shown
DPG result is a simplified visualization of groups formed through different
processing domains and at different spatial extents. Descriptive terms used
below, such as bucket, tree, or illumination-associated group, are assigned
after processing only to identify the illustrated structures.

\begin{figure}[!t]
    \centering

    \subfloat[The selected parent group in the input image.\label{fig:bucket_parent_image}]{%
    \begin{minipage}[t]{0.32\textwidth}
        \centering
        \includegraphics[width=\linewidth]
        {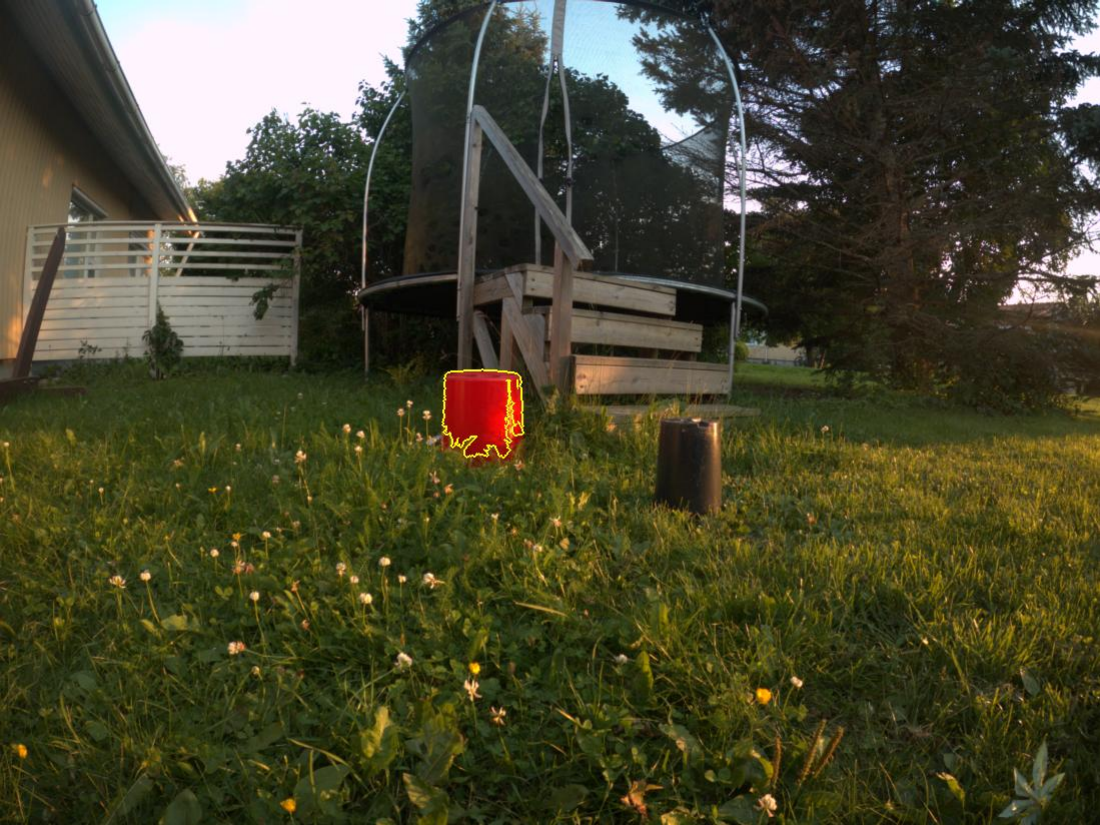}
    \end{minipage}}
    \hfill
    \subfloat[The parent group shown separately.\label{fig:bucket_parent_mask}]{%
    \begin{minipage}[t]{0.32\textwidth}
        \centering
        \includegraphics[width=\linewidth]
        {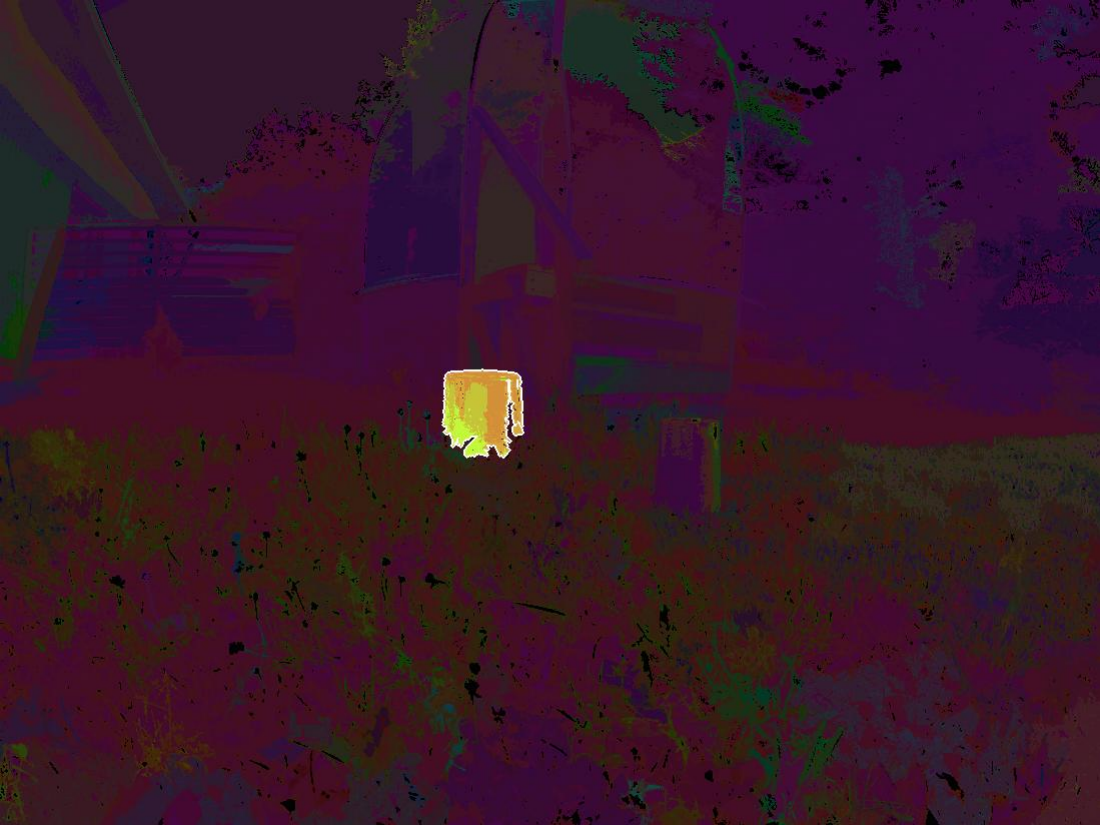}
    \end{minipage}}
    \hfill
    \subfloat[Child groups contained by the selected parent.\label{fig:bucket_children}]{%
    \begin{minipage}[t]{0.32\textwidth}
        \centering
        \includegraphics[width=\linewidth]
        {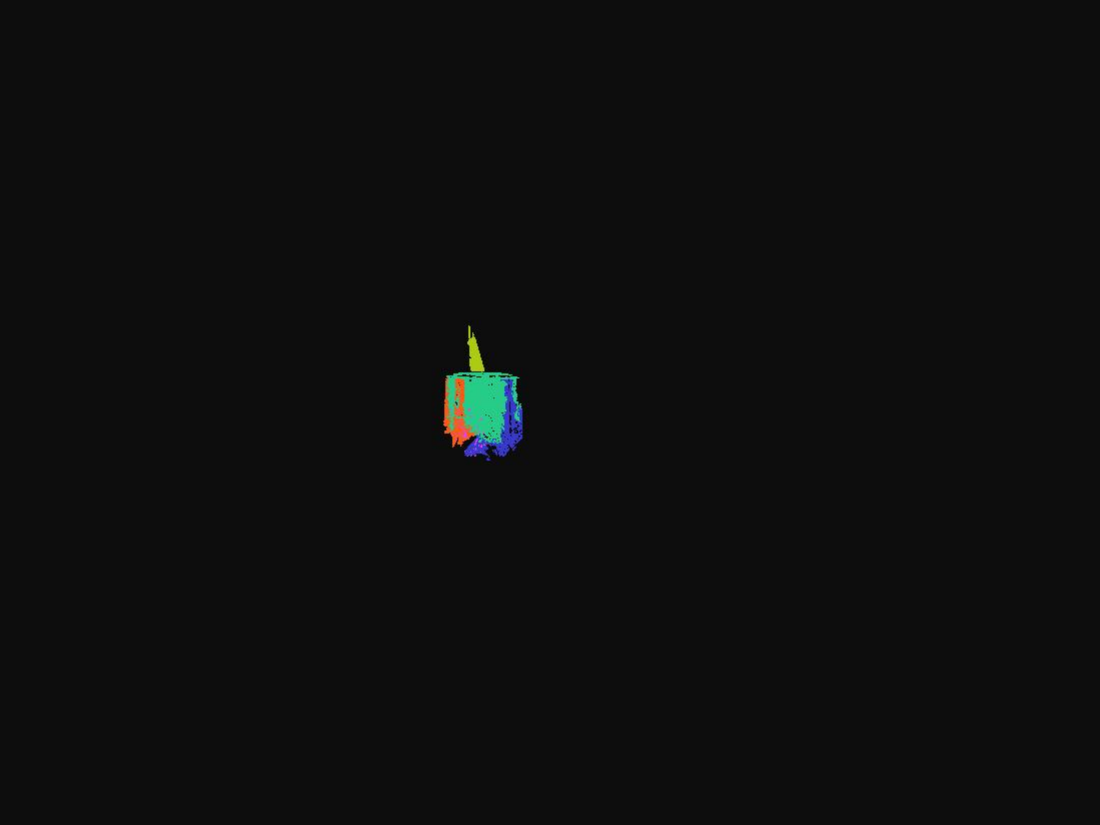}
    \end{minipage}}

    \caption{Parent-child organization associated with the red bucket.
    The parent retains most of the bucket within one group despite internal
    measurement variation, while its child groups represent more localized
    regions.}
    \label{fig:lighted_bucket_hierarchy}
\end{figure}

Figure~\ref{fig:lighted_bucket_hierarchy} shows a selected parent group that
covers most of the region subsequently described as the red bucket, together
with its spatially related child groups. The RGB measurements within the
parent vary with illumination, shading, the upper rim, and local surface
structure.\\

The broader parent remains available as one group despite this internal
variation, while the associated child groups represent more localized
measurement-supported regions within the same spatial organization.

\begin{figure}[!p]
    \centering
    \subfloat[The larger tree-associated parent group in the input image.\label{fig:lighted_tree_large_image}]{%
    \begin{minipage}[t]{0.36\textwidth}
        \centering
        \includegraphics[width=\linewidth]
        {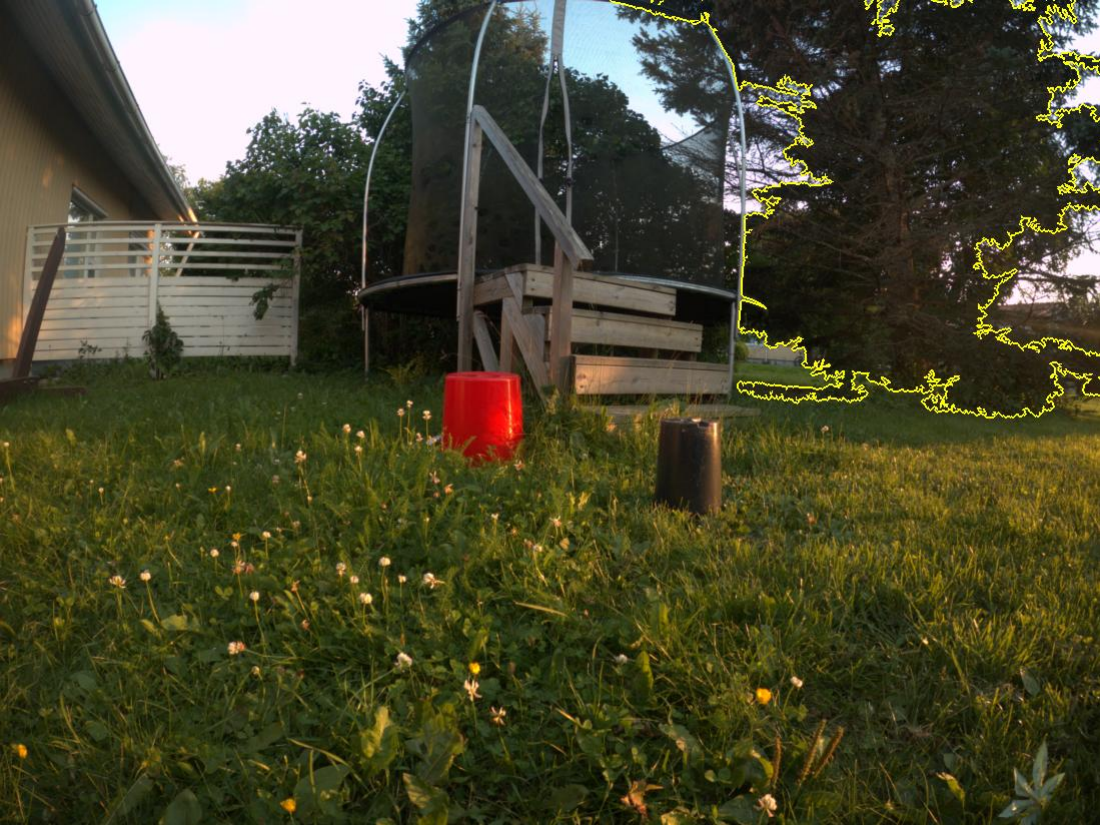}
    \end{minipage}}
    \hfill
    \subfloat[The larger parent group shown separately.\label{fig:lighted_tree_large_mask}]{%
    \begin{minipage}[t]{0.36\textwidth}
        \centering
        \includegraphics[width=\linewidth]
        {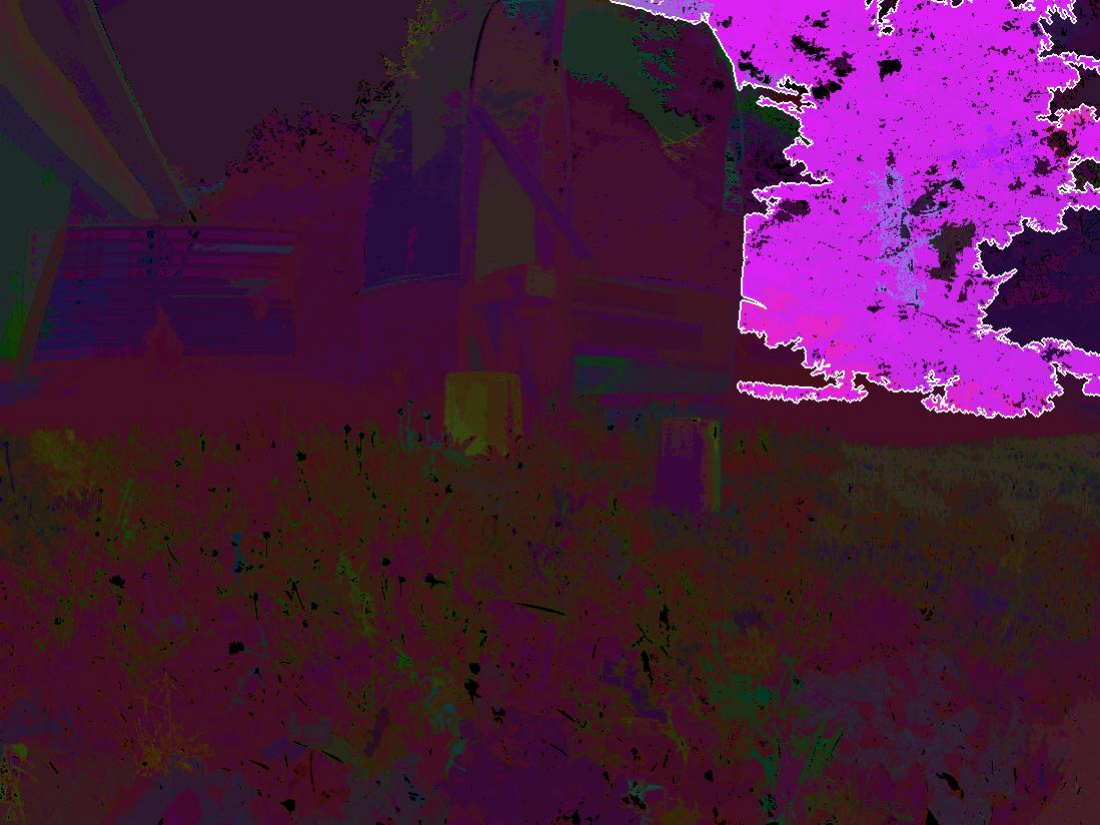}
    \end{minipage}} 
    \vspace{0.3em}
    \subfloat[A smaller tree-associated parent group.\label{fig:lighted_tree_small_image}]{%
    \begin{minipage}[t]{0.36\textwidth}
        \centering
        \includegraphics[width=\linewidth]
        {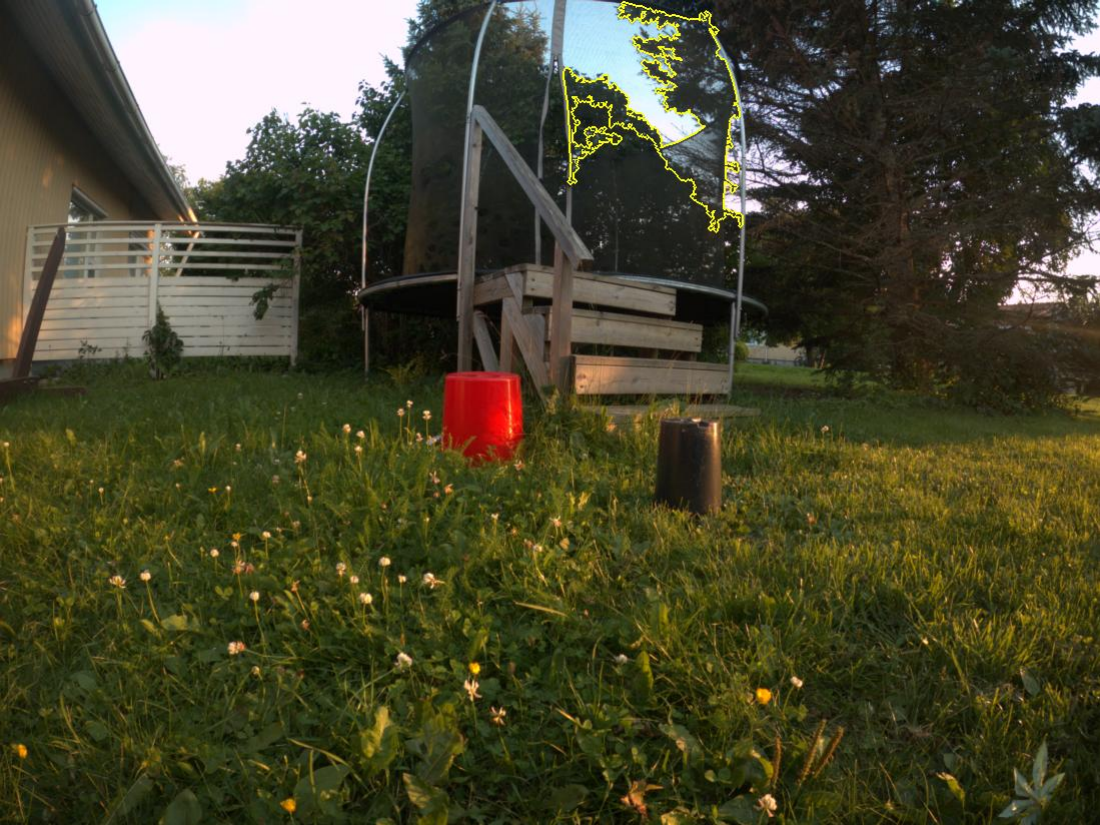}
    \end{minipage}}
    \hfill
    \subfloat[The smaller parent group shown separately.\label{fig:lighted_tree_small_mask}]{%
    \begin{minipage}[t]{0.36\textwidth}
        \centering
        \includegraphics[width=\linewidth]
        {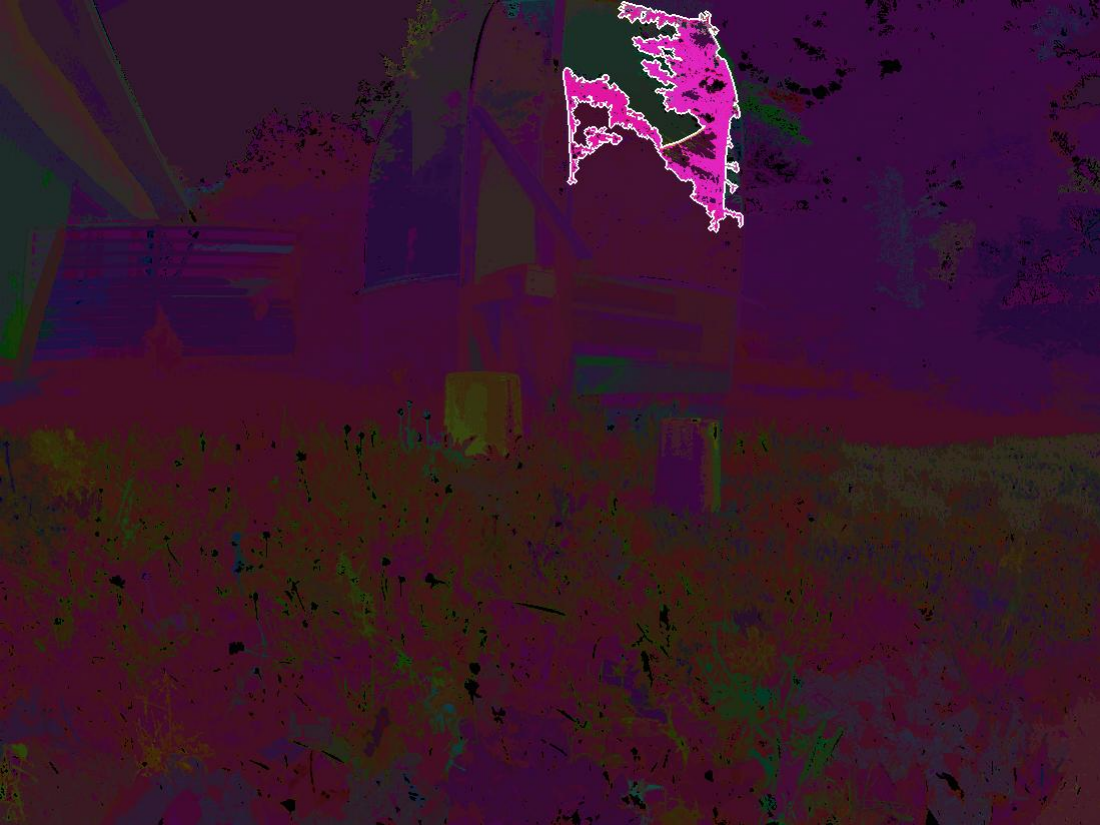}
    \end{minipage}}
    
    \vspace{0.3em}
    
    \subfloat[A group obtained by reprocessing the smaller parent group.\label{fig:lighted_tree_rerun_image}]{%
    \begin{minipage}[t]{0.36\textwidth}
        \centering
        \includegraphics[width=\linewidth]
        {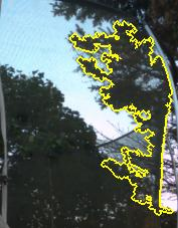}
    \end{minipage}}
    \hfill
    \subfloat[The reprocessed group shown separately.\label{fig:lighted_tree_rerun_mask}]{%
    \begin{minipage}[t]{0.36\textwidth}
        \centering
        \includegraphics[width=\linewidth]
        {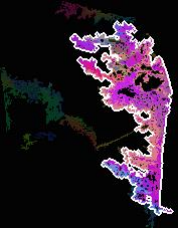}
    \end{minipage}}

    \caption{Example of grouping at multiple levels of spatial extent.
    The displayed groups were selected after processing based on their
    spatial correspondence with the tree structures. The original processing produces both
    a larger tree-associated parent group and a more spatially restricted
    group behind the trampoline. Reprocessing the restricted group exposes
    additional internal organization at a finer observation level.}
    \label{fig:lighted_tree_hierarchy}
\end{figure}

Figure~\ref{fig:lighted_tree_hierarchy} shows two parent groups associated
with the same tree structure at different spatial extents. One covers a
substantial part of the vegetation on the right-hand side of the image,
whereas the other is restricted to the region behind the trampoline.\\

Reprocessing the more restricted parent produces an additional group
corresponding to a smaller subset of the previously grouped foliage. The
original groups remain available, so the result demonstrates a change in
observational resolution without imposing a single exclusive hierarchy.
\FloatBarrier

\begin{figure}[!t]
    \centering

    \subfloat[The selected group outlined in the input image.\label{fig:light_rays_image}]{%
    \begin{minipage}[t]{0.45\textwidth}
        \centering
        \includegraphics[width=\linewidth]
        {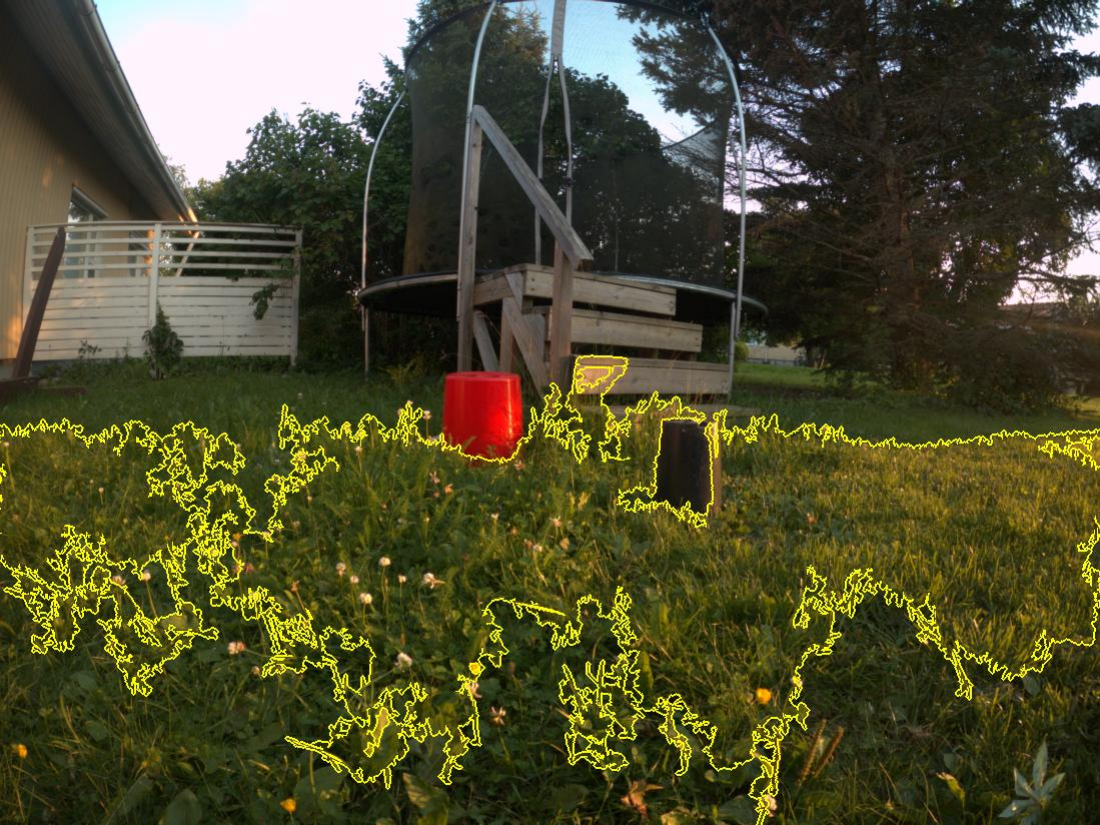}
    \end{minipage}}
    \hfill
    \subfloat[The illumination-associated group shown separately.\label{fig:light_rays_mask}]{%
    \begin{minipage}[t]{0.45\textwidth}
        \centering
        \includegraphics[width=\linewidth]
        {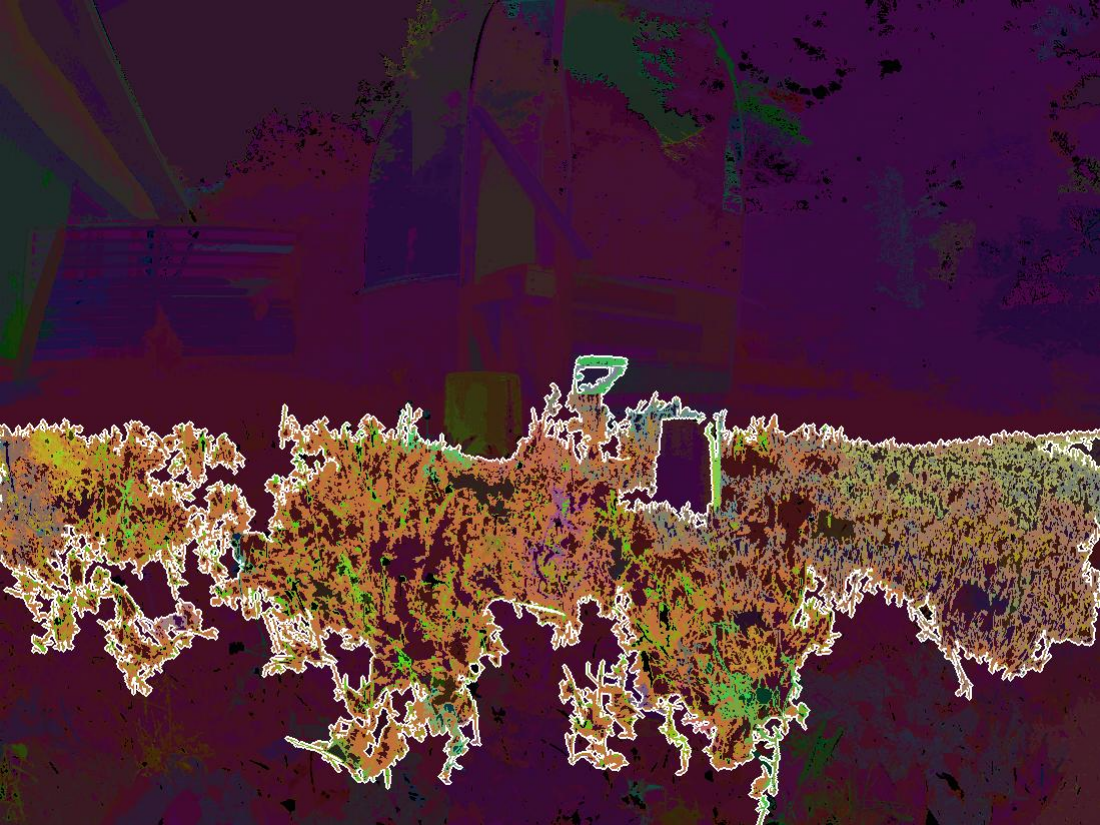}
    \end{minipage}}

    \caption{A grouping hypothesis associated with the illuminated foreground
    structure. The group extends across spatially detailed vegetation and
    does not correspond to a single semantic object or a conventional
    mutually exclusive scene segment.}
    \label{fig:lighted_rays_group}
\end{figure}

Figure~\ref{fig:lighted_rays_group} shows a parent group extending across the
illuminated foreground vegetation rather than following the extent of a
single semantic object. The group can coexist and overlap with groups
supported by other measurement relationships, illustrating how DPG can retain
spatial organization that crosses an object-centred partition of the scene.\\

\begin{figure}[!t]
    \centering

    \subfloat[Felzenszwalb\label{fig:dark_felzenszwalb}]{%
    \begin{minipage}[t]{0.32\textwidth}
        \centering
        \includegraphics[width=\linewidth]
        {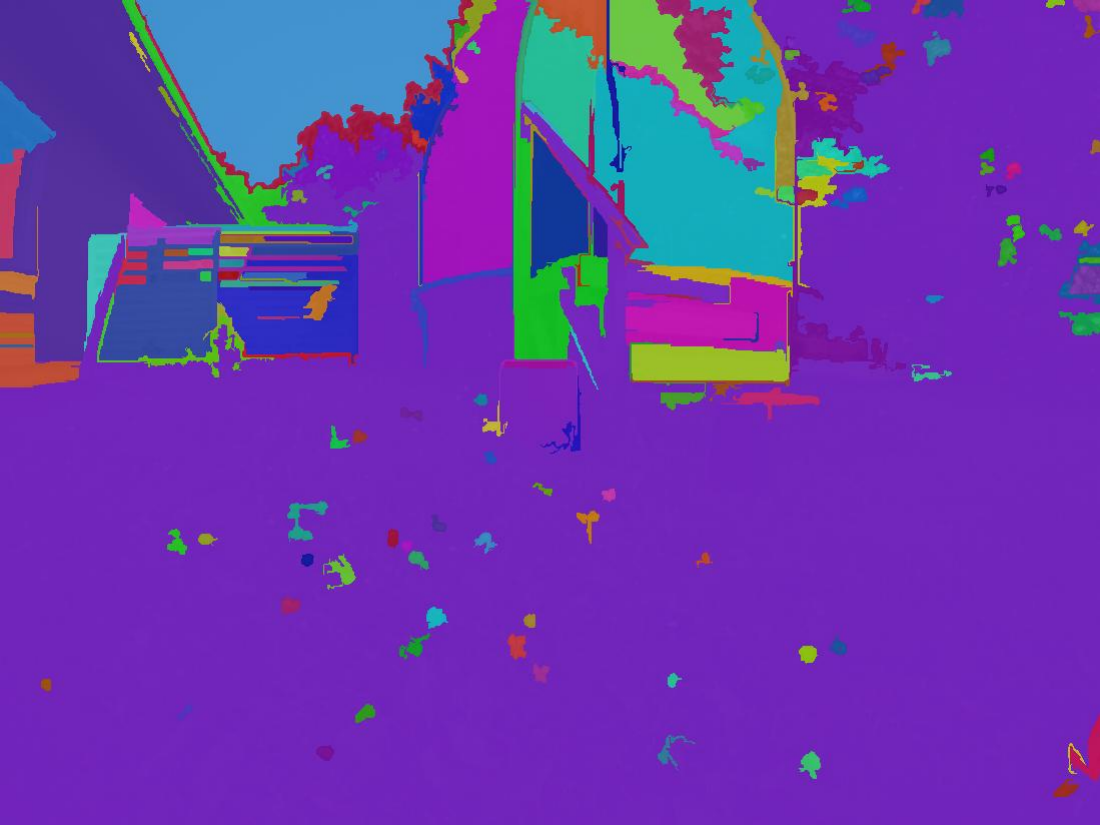}
    \end{minipage}}
    \hfill
    \subfloat[Watershed with boundary-RAG merging\label{fig:dark_watershed}]{%
    \begin{minipage}[t]{0.32\textwidth}
        \centering
        \includegraphics[width=\linewidth]
        {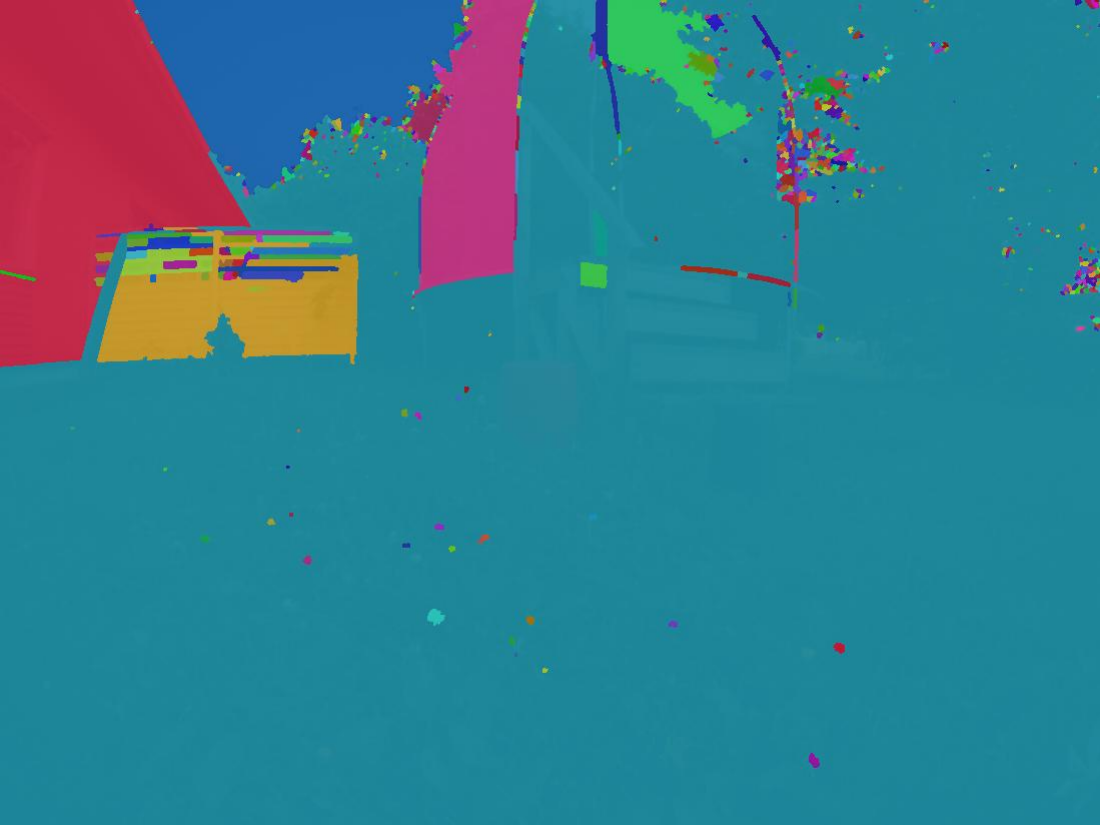}
    \end{minipage}}
    \hfill
    \subfloat[Quickshift\label{fig:dark_quickshift}]{%
    \begin{minipage}[t]{0.32\textwidth}
        \centering
        \includegraphics[width=\linewidth]
        {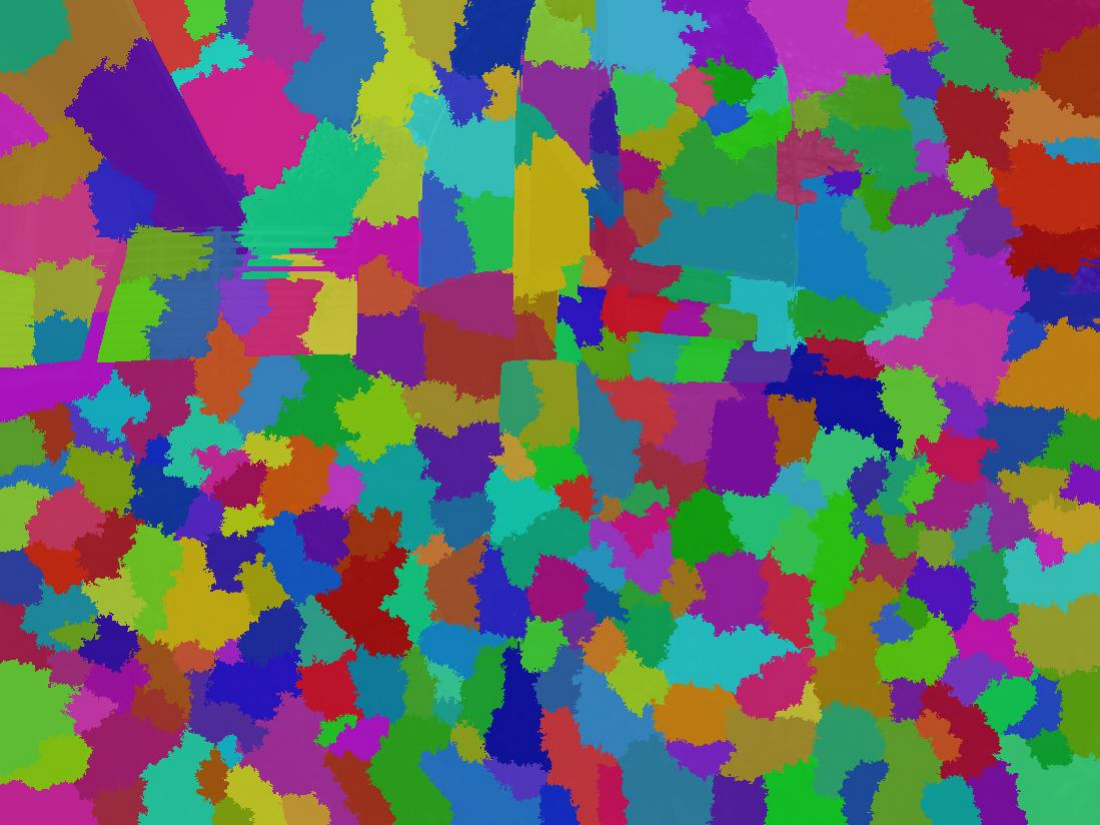}
    \end{minipage}}

    \vspace{0.6em}

    \subfloat[DPG\label{fig:dark_dpg}]{%
    \begin{minipage}[t]{0.44\textwidth}
        \centering
        \includegraphics[width=\linewidth]
        {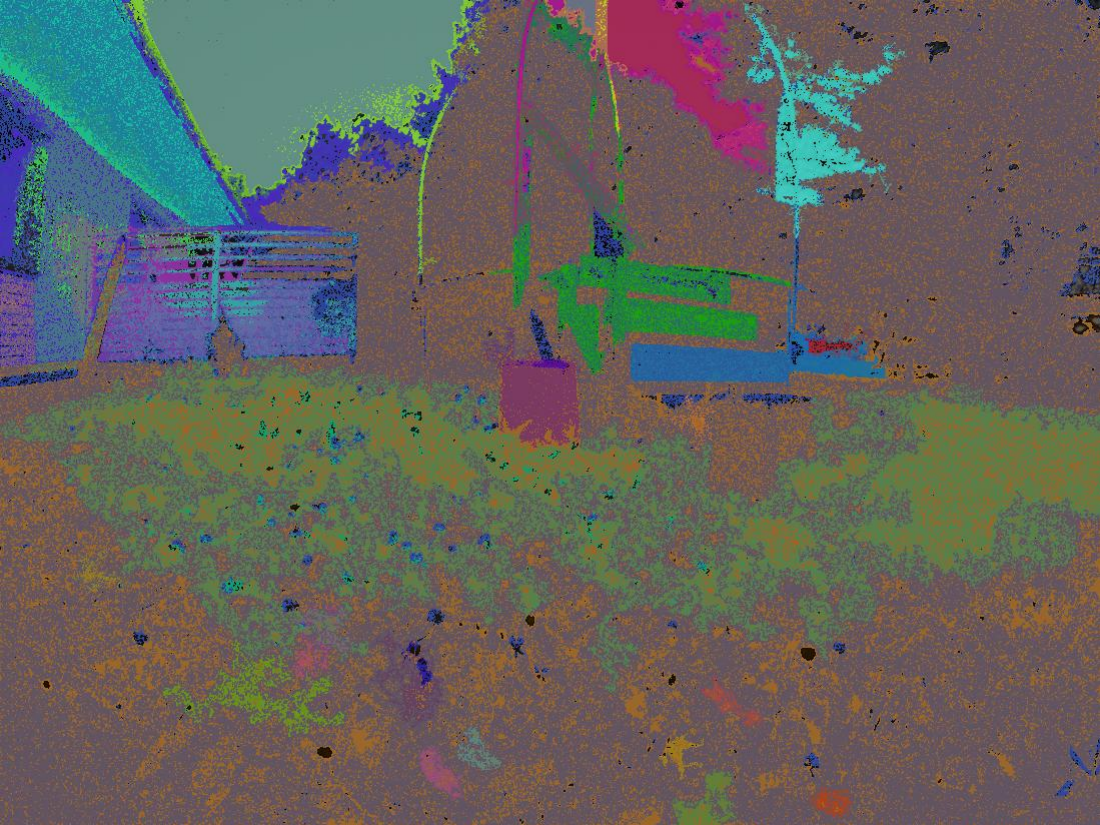}
    \end{minipage}}
    \hspace{0.04\textwidth}
    \subfloat[Original image\label{fig:dark_original}]{%
    \begin{minipage}[t]{0.44\textwidth}
        \centering
        \includegraphics[width=\linewidth]
        {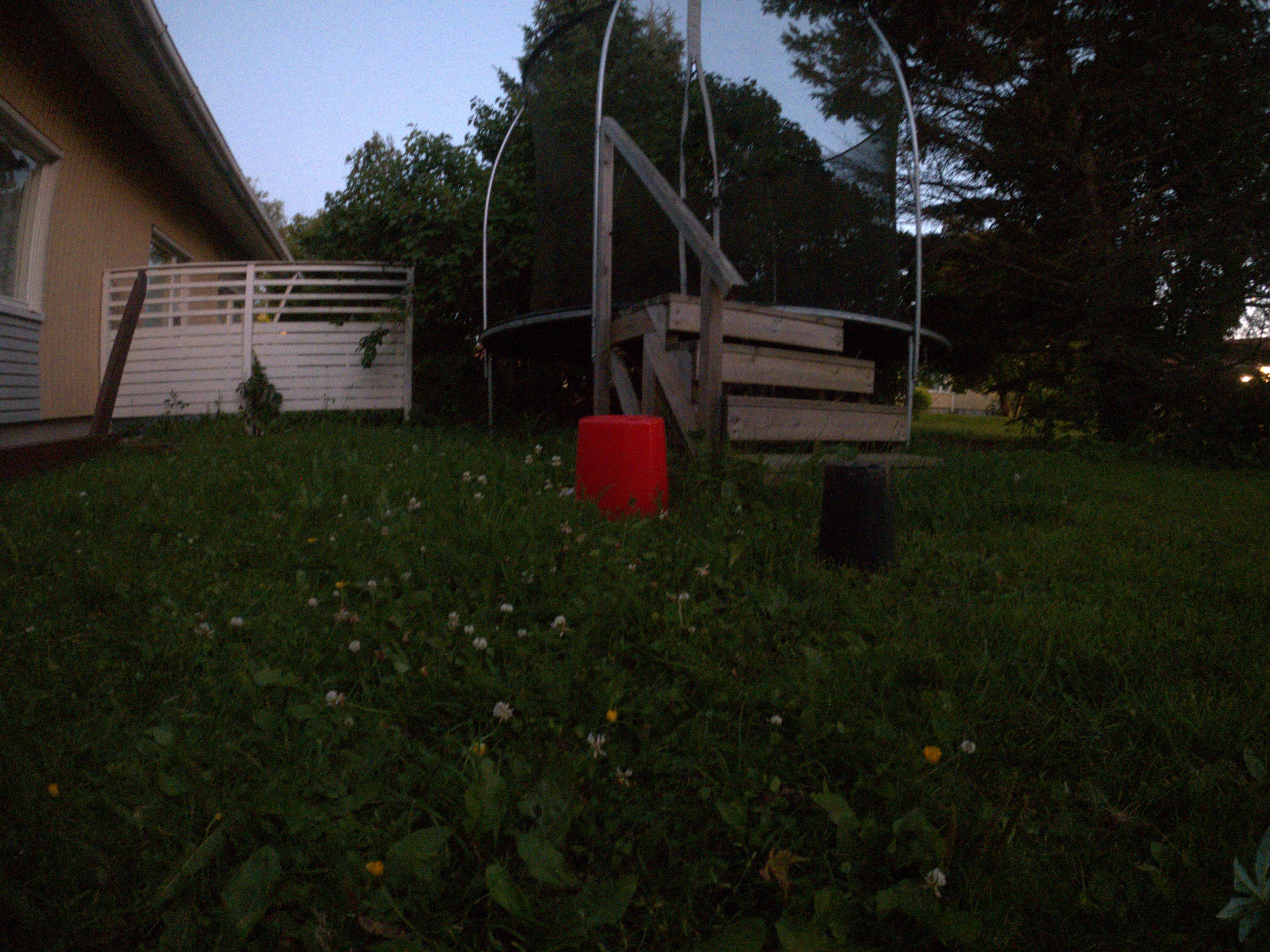}
    \end{minipage}}

    \caption{Qualitative comparison for the lower-luminance capture, GOPR1792.
    The top row shows the mutually exclusive partitions produced by the three
    comparison methods. The bottom row shows the simplified visualization of the
    complete DPG group set and the original input image.}
    \label{fig:dark_method_comparison}
\end{figure}

Figure~\ref{fig:dark_method_comparison} compares the methods for the
lower-luminance capture. Major structures such as the building, fence,
trampoline, and red container remain represented to different degrees, while
separation within the dark foreground is reduced.\\

Felzenszwalb retains several strongly separated structures but merges much of
the low-contrast foreground into broader regions. Quickshift again produces a
dense local partition, whereas watershed-RAG merges most of the dark
foreground into one dominant region. The DPG result retains both broader
scene-associated groups and more localized variation within the foreground.\\

\begin{figure}[!p]
    \centering

    \subfloat[The selected broad parent group in the original image.\label{fig:greens_parent_image}]{%
    \begin{minipage}[t]{0.45\textwidth}
        \centering
        \includegraphics[width=\linewidth]
        {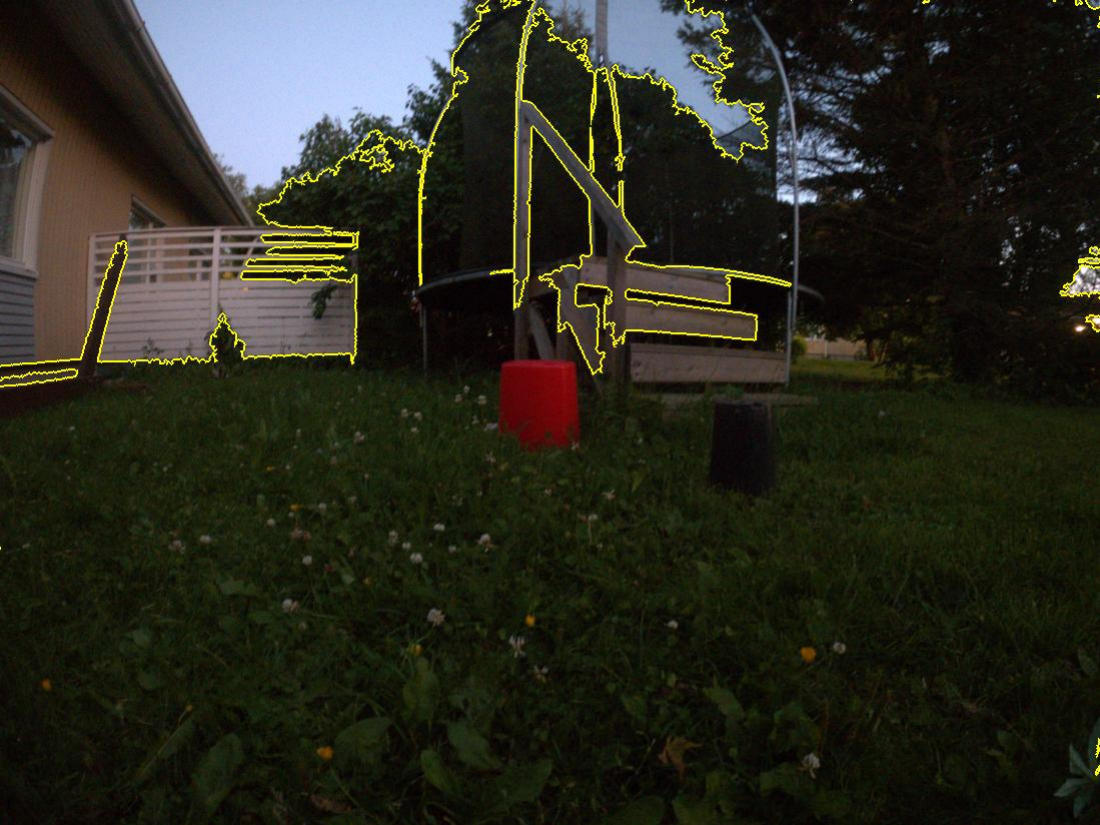}
    \end{minipage}}
    \hfill
    \subfloat[The selected parent group shown separately.\label{fig:greens_parent_group}]{%
    \begin{minipage}[t]{0.45\textwidth}
        \centering
        \includegraphics[width=\linewidth]
        {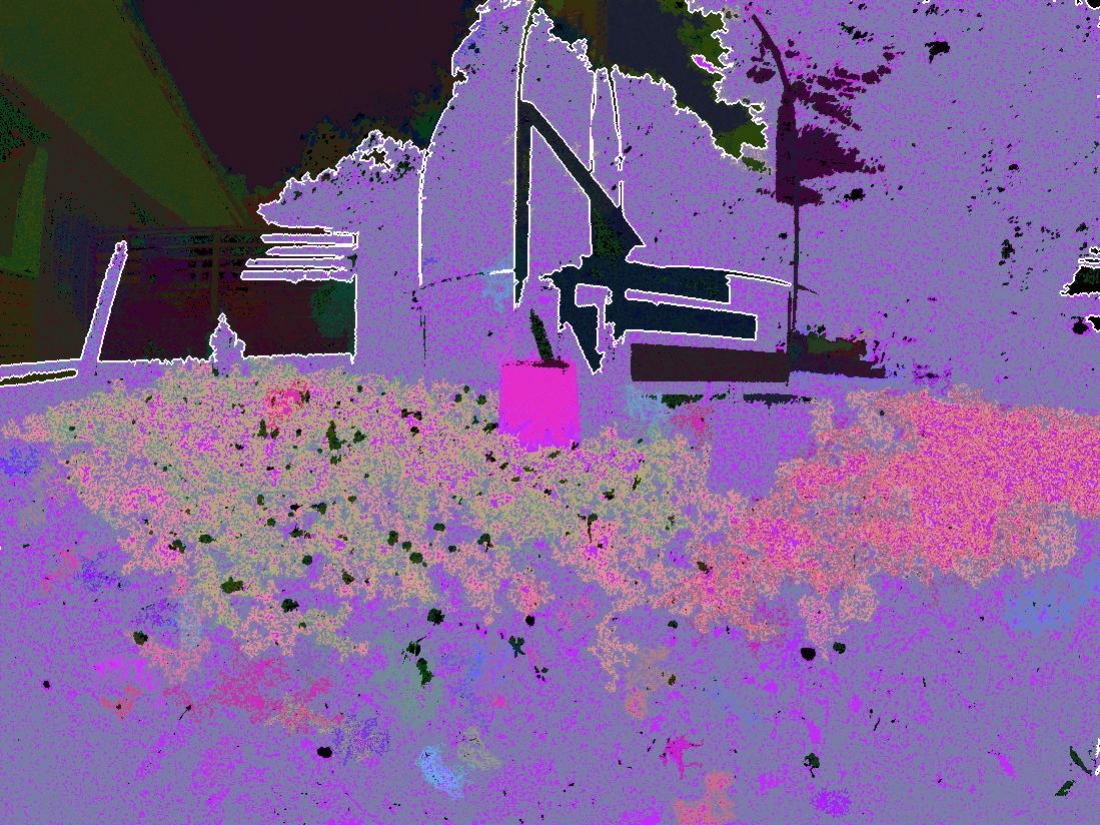}
    \end{minipage}}

    \vspace{0.8em}

    \subfloat[A tree-associated parent obtained by reprocessing the selected group.\label{fig:greens_rerun_tree_image}]{%
    \begin{minipage}[t]{0.45\textwidth}
        \centering
        \includegraphics[width=\linewidth]
        {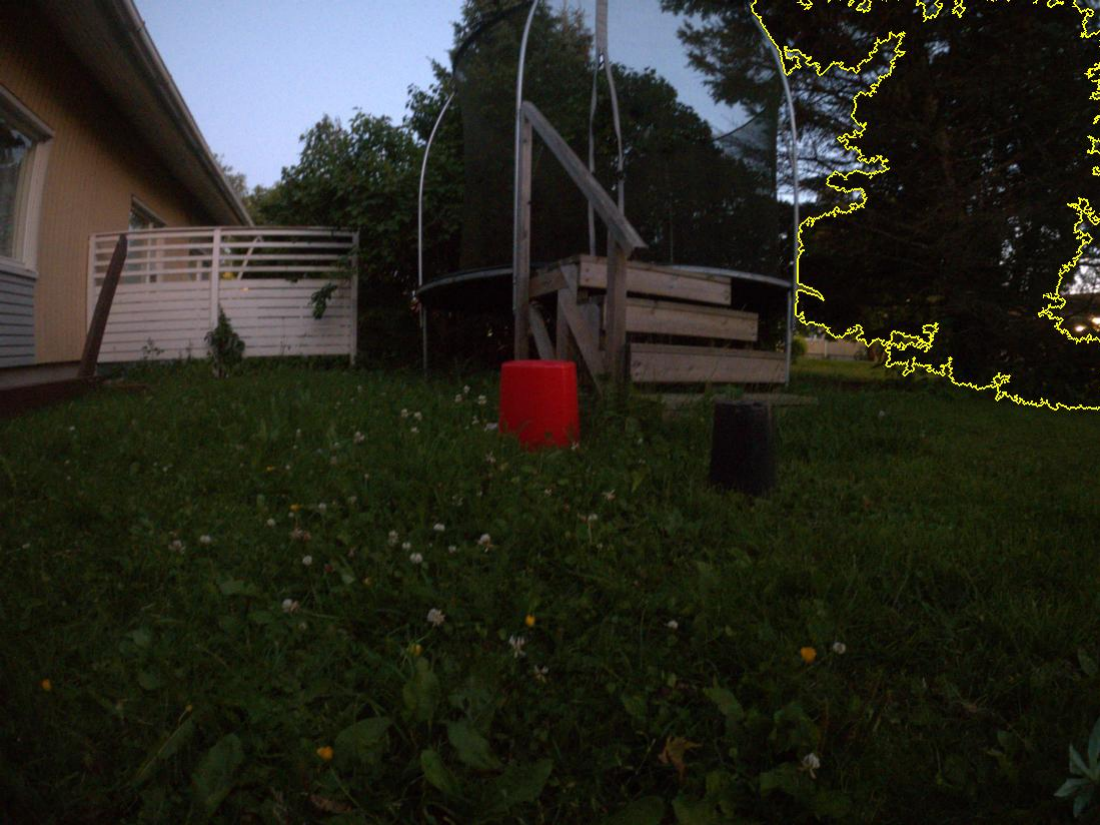}
    \end{minipage}}
    \hfill
    \subfloat[The tree-associated parent in the reprocessed representation.\label{fig:greens_rerun_tree_group}]{%
    \begin{minipage}[t]{0.45\textwidth}
        \centering
        \includegraphics[width=\linewidth]
        {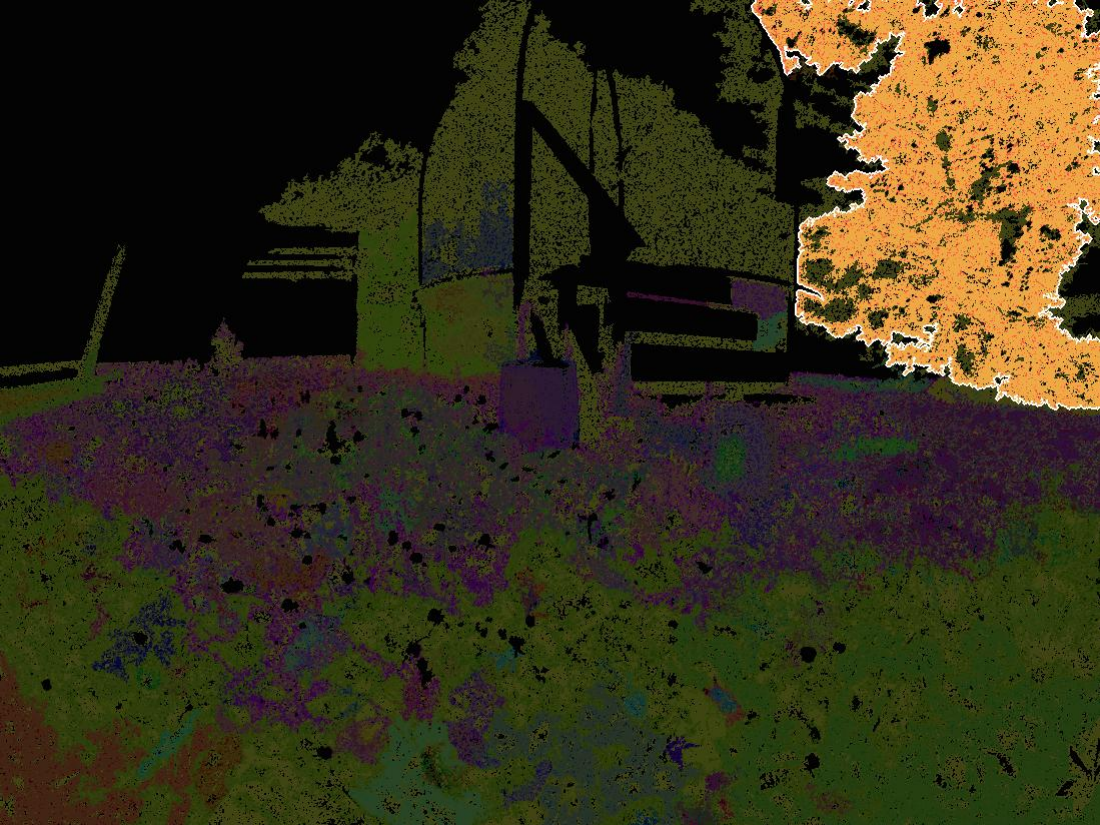}
    \end{minipage}}

    \vspace{0.8em}

    \subfloat[A bucket-associated parent obtained by reprocessing the selected group.\label{fig:greens_rerun_bucket_image}]{%
    \begin{minipage}[t]{0.45\textwidth}
        \centering
        \includegraphics[width=\linewidth]
        {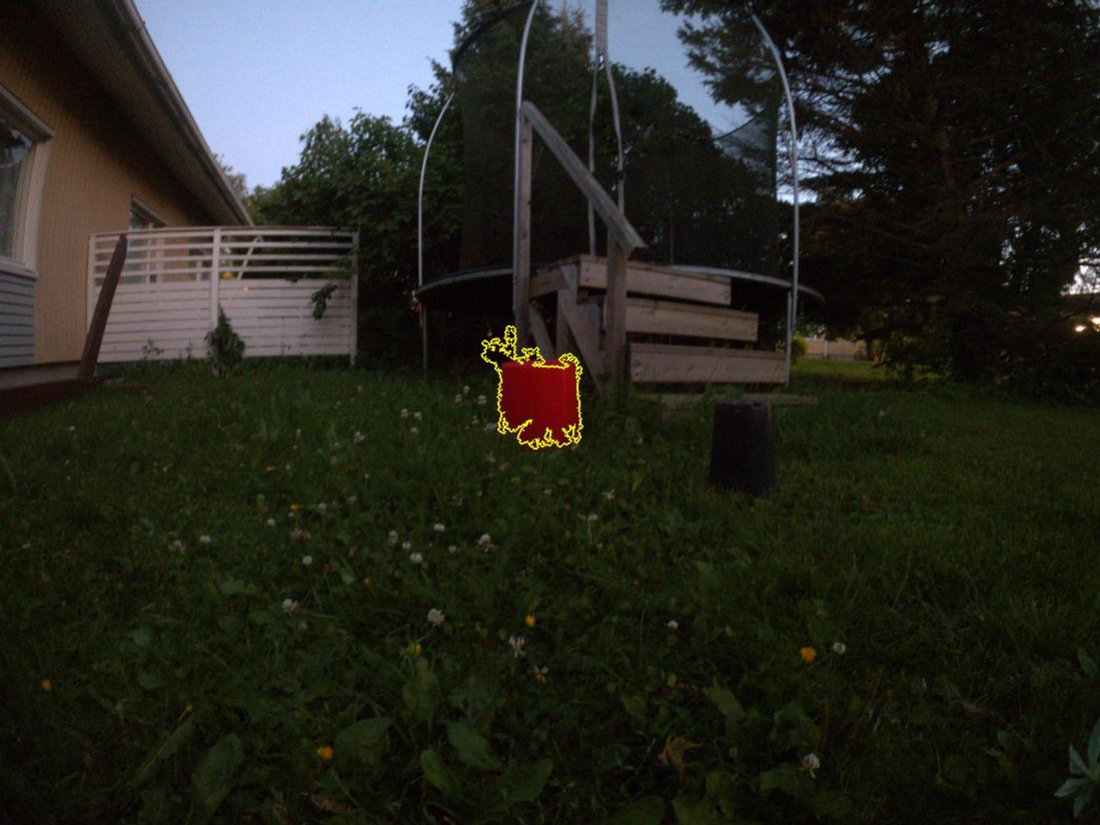}
    \end{minipage}}
    \hfill
    \subfloat[The bucket-associated parent in the reprocessed representation.\label{fig:greens_rerun_bucket_group}]{%
    \begin{minipage}[t]{0.45\textwidth}
        \centering
        \includegraphics[width=\linewidth]
        {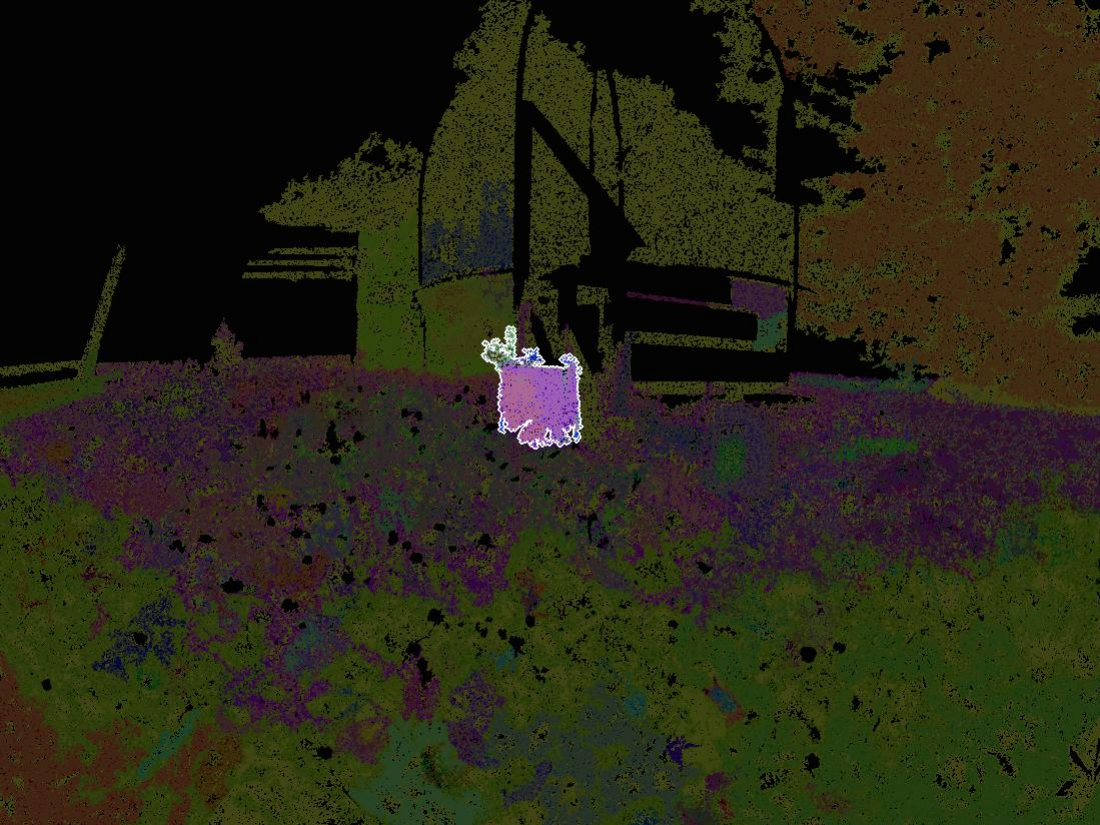}
    \end{minipage}}

    \caption{Reprocessing of a broad parent group under lower-luminance
    conditions. The initial parent contains substantial internal variation.
    Reprocessing makes more spatially restricted parent groups associated with
    the tree and red bucket separately available.}
    \label{fig:greens_low_light_reprocessing}
\end{figure}

Figure~\ref{fig:greens_low_light_reprocessing} shows reprocessing of a broad
parent group from the lower-luminance image. In the initial result, this
parent extends across a large part of the green-associated foreground and
retains substantial internal measurement variation.\\

After reprocessing, more localized parent groups associated with the tree and
the red bucket become separately available. These groups do not replace the
initial parent or form an exhaustive subdivision of it, demonstrating the
intended change in observational resolution within the selected content.

\FloatBarrier

\begin{figure}[!t]
    \centering

    \subfloat[The selected black-bucket-associated parent outlined in the original image.\label{fig:black_bucket_rerun_image}]{%
    \begin{minipage}[t]{0.49\textwidth}
        \centering
        \includegraphics[width=\linewidth]
        {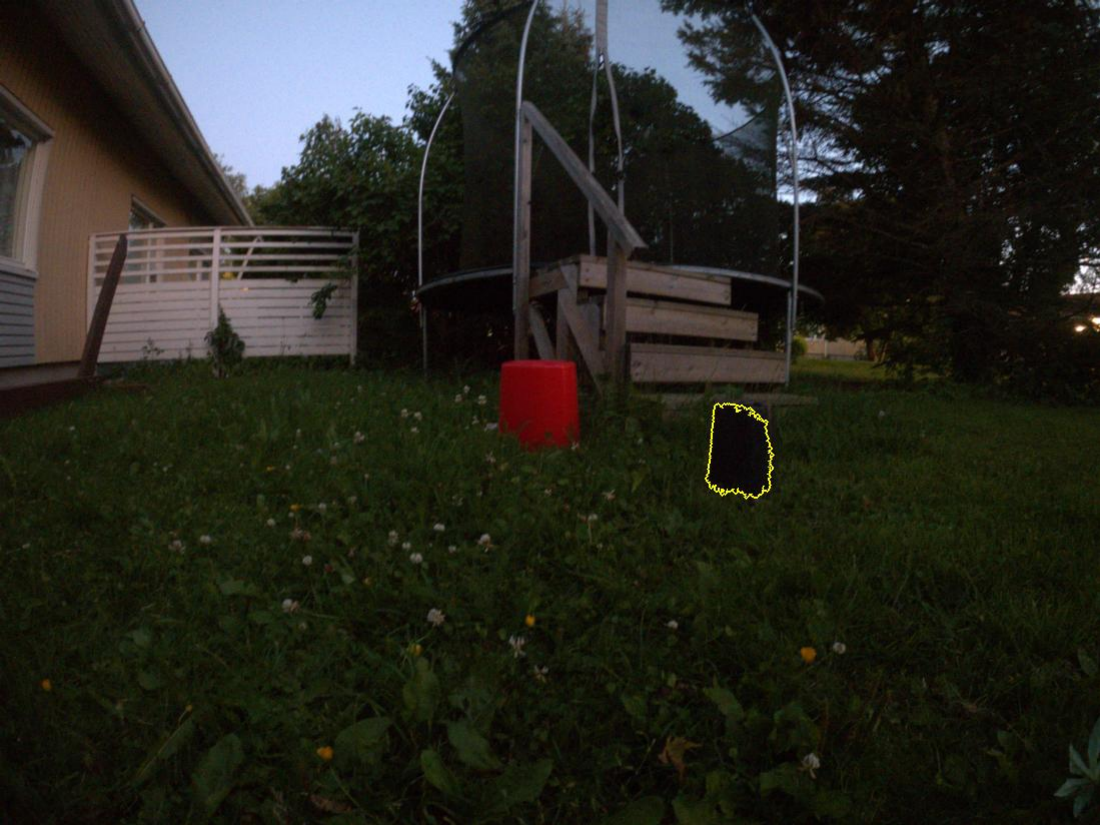}
    \end{minipage}}
    \hfill
    \subfloat[The parent group in the reprocessed representation.\label{fig:black_bucket_rerun_group}]{%
    \begin{minipage}[t]{0.49\textwidth}
        \centering
        \includegraphics[width=\linewidth]
        {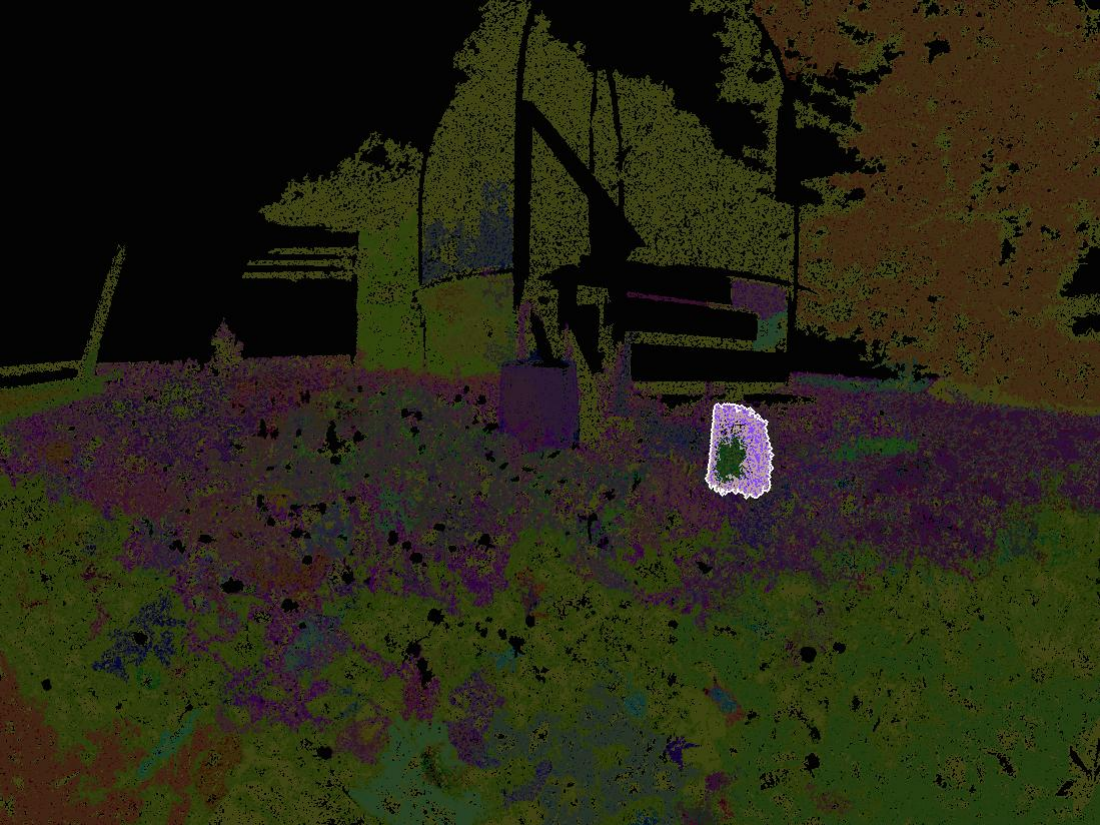}
    \end{minipage}}

    \caption{Black-container-associated parent obtained by reprocessing the
    broad parent in Figure~\ref{fig:greens_low_light_reprocessing}. Despite weak
    separation from parts of the surrounding low-luminance vegetation, most of
    the visible container extent forms a distinct group in the restricted
    observation context.}
    \label{fig:black_bucket_low_light_reprocessing}
\end{figure}

The same reprocessed output contains the black-container-associated parent
shown in Figure~\ref{fig:black_bucket_low_light_reprocessing}. Unlike the
strongly chromatic red bucket, the black container is nearly achromatic and
has relatively low luminance, with weaker measurement separation from parts
of the surrounding shaded vegetation. Nevertheless, the resulting group
follows most of its visible external extent.\\

Together, the red- and black-container examples show that reprocessing can
make localized groups available from both strongly chromatic and weaker
luminance-associated structure within the same broad initial parent.\\

The two captures also illustrate how the same visible structure can be grouped
differently as its recorded measurements change. In the lower-luminance image,
where illumination across the white fence is comparatively uniform, DPG,
Felzenszwalb, and watershed-RAG each retain most of its visible extent within
a single group or region. Under the brighter but less uniform illumination,
the corresponding results show greater fragmentation of the same fence.
This provides a direct example of grouping changing with the available
measurements even though the scene structure itself remains similar.

\FloatBarrier
\subsection{Domain Evaluation}

\begin{figure}[!p]
    \centering

    \subfloat[Region covering\label{fig:dpg_domain_covering}]{%
    \begin{minipage}[t]{0.485\textwidth}
        \centering
        \includegraphics[width=\linewidth]
        {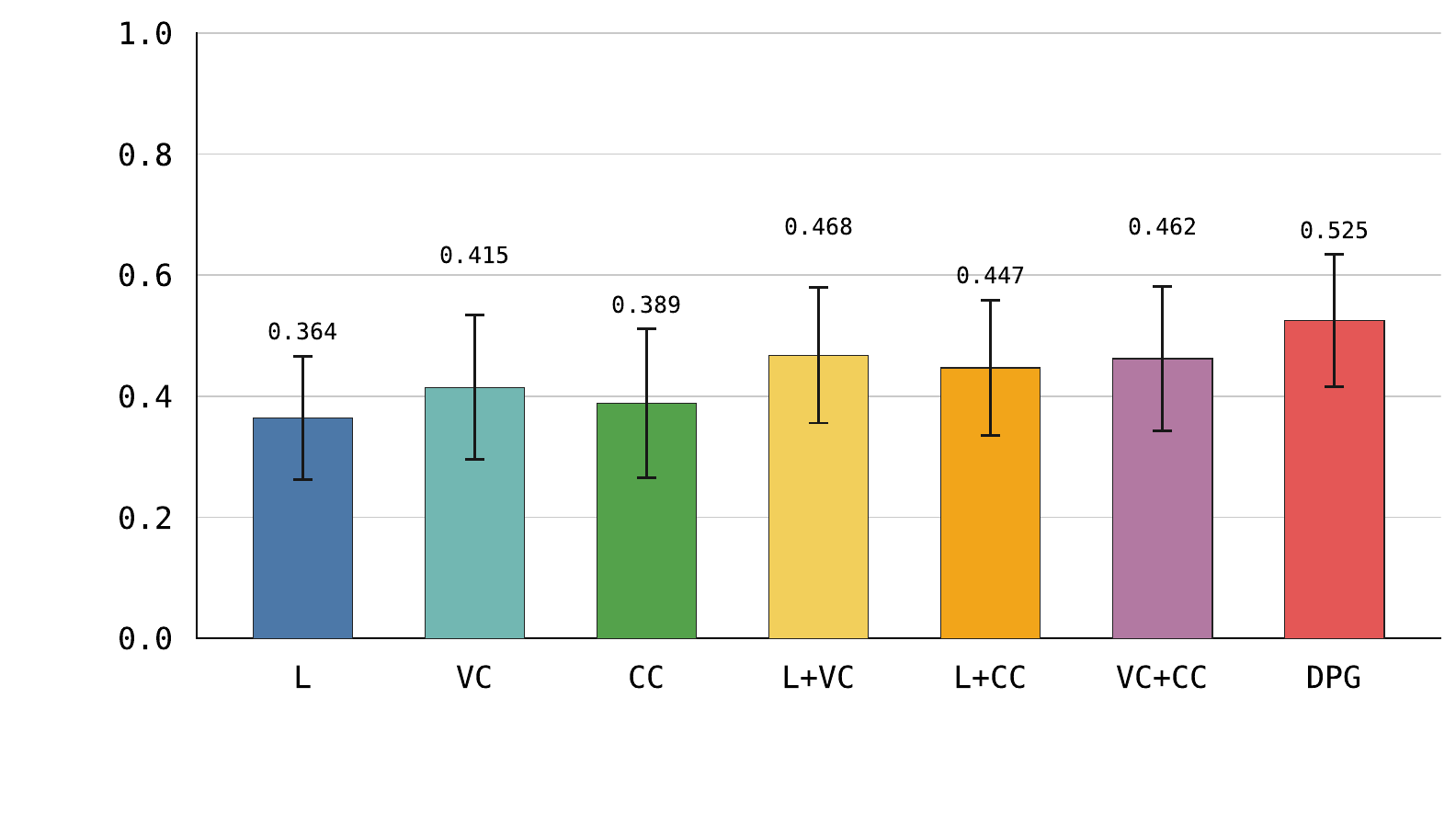}
    \end{minipage}}
    \hfill
    \subfloat[Region recall at IoU 0.50\label{fig:dpg_domain_recall_050}]{%
    \begin{minipage}[t]{0.485\textwidth}
        \centering
        \includegraphics[width=\linewidth]
        {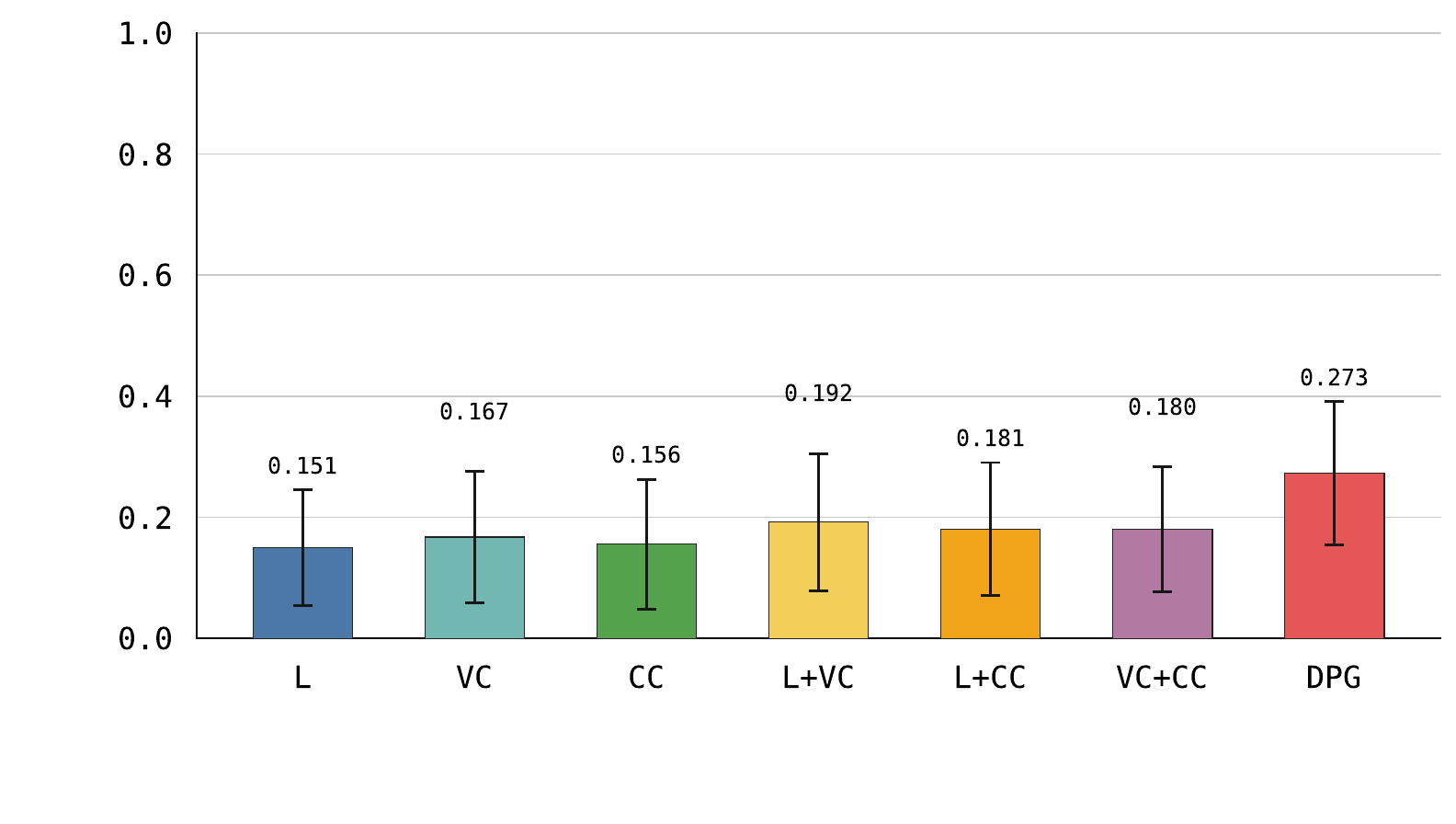}
    \end{minipage}}

    \vspace{0.4em}

    \subfloat[Region recall at IoU 0.75\label{fig:dpg_domain_recall_075}]{%
    \begin{minipage}[t]{0.485\textwidth}
        \centering
        \includegraphics[width=\linewidth]
        {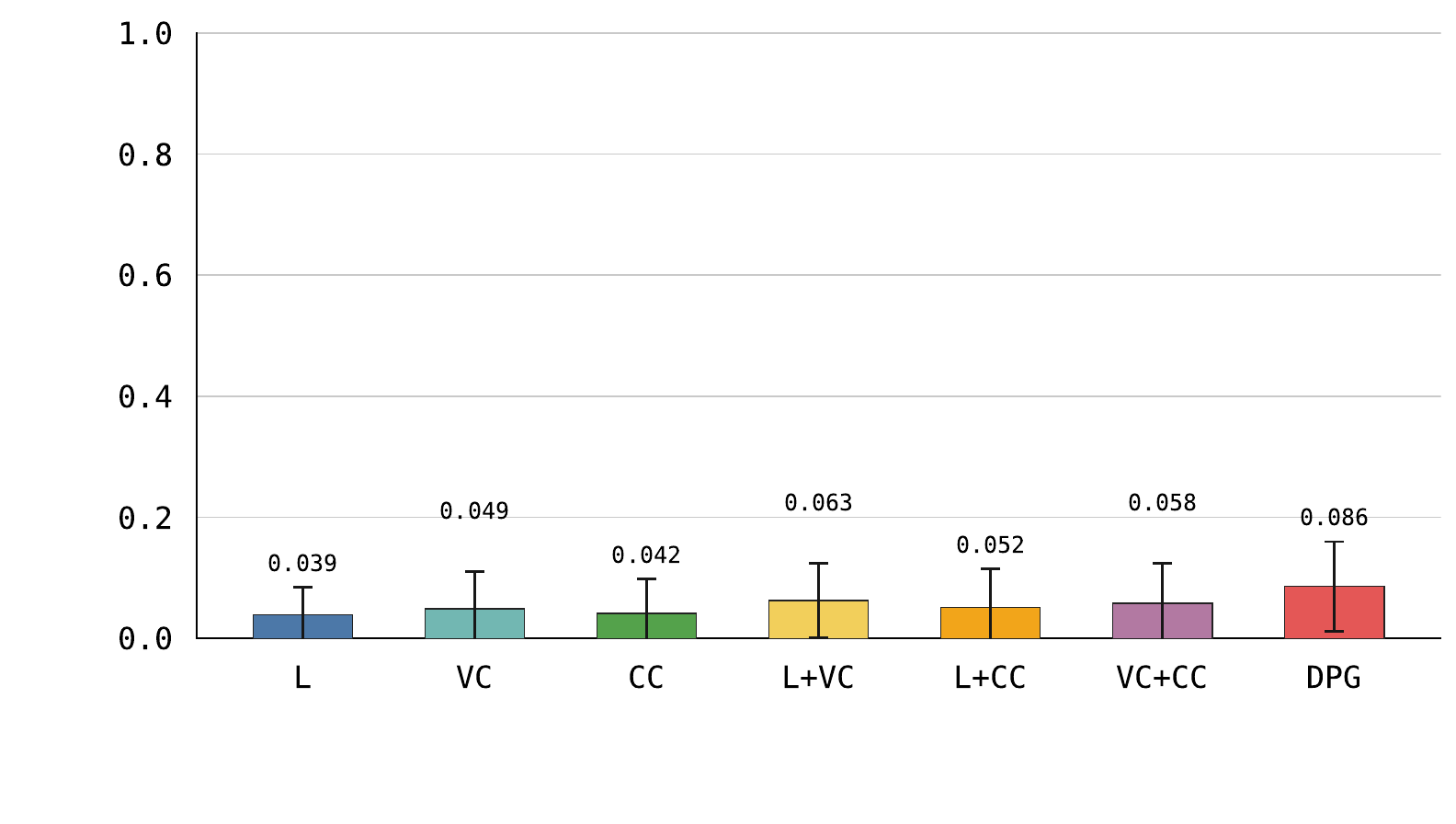}
    \end{minipage}}
    \hfill
    \subfloat[Boundary precision\label{fig:dpg_domain_boundary_precision}]{%
    \begin{minipage}[t]{0.485\textwidth}
        \centering
        \includegraphics[width=\linewidth]
        {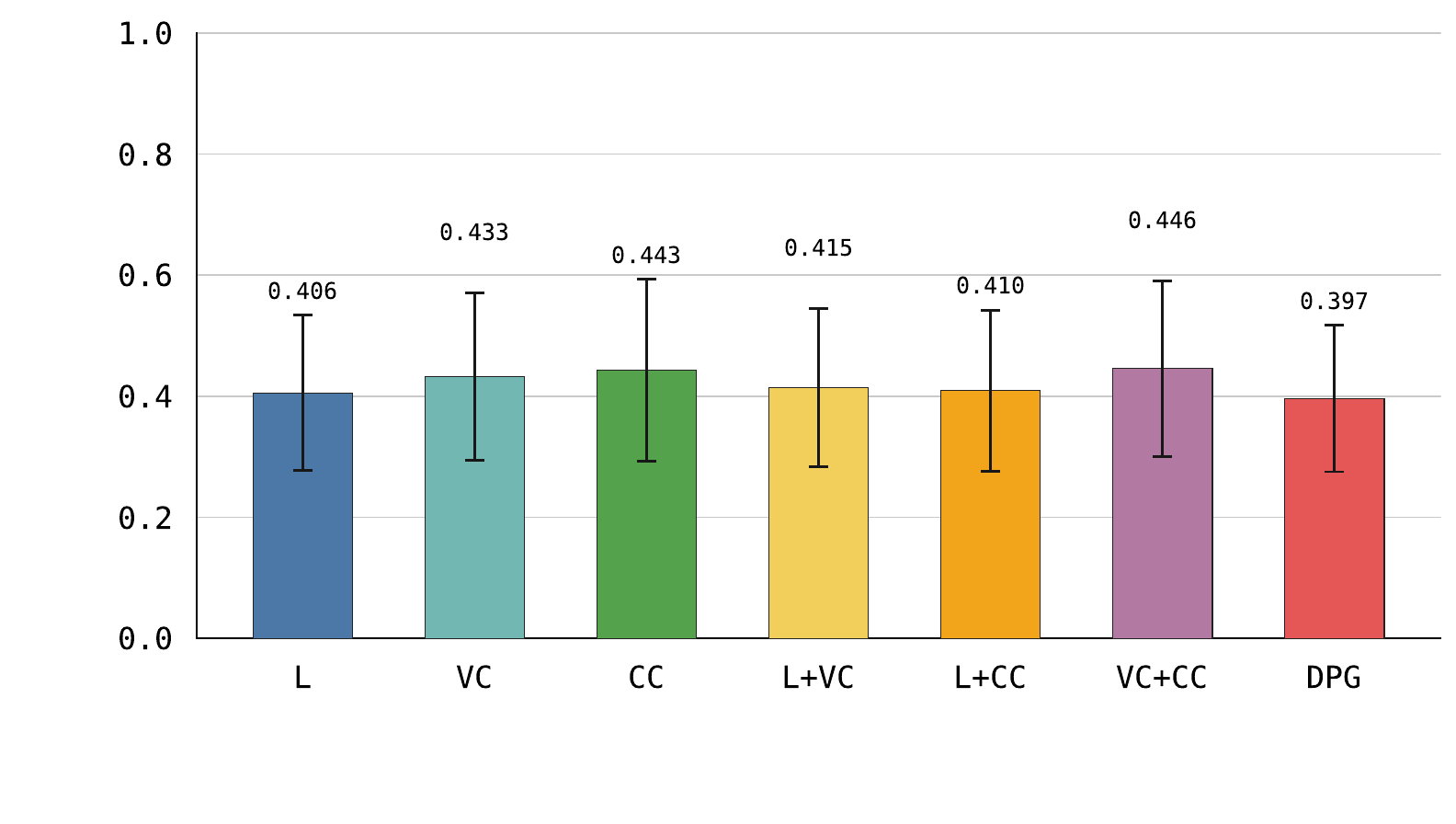}
    \end{minipage}}

    \vspace{0.4em}

    \subfloat[Boundary recall\label{fig:dpg_domain_boundary_recall}]{%
    \begin{minipage}[t]{0.485\textwidth}
        \centering
        \includegraphics[width=\linewidth]
        {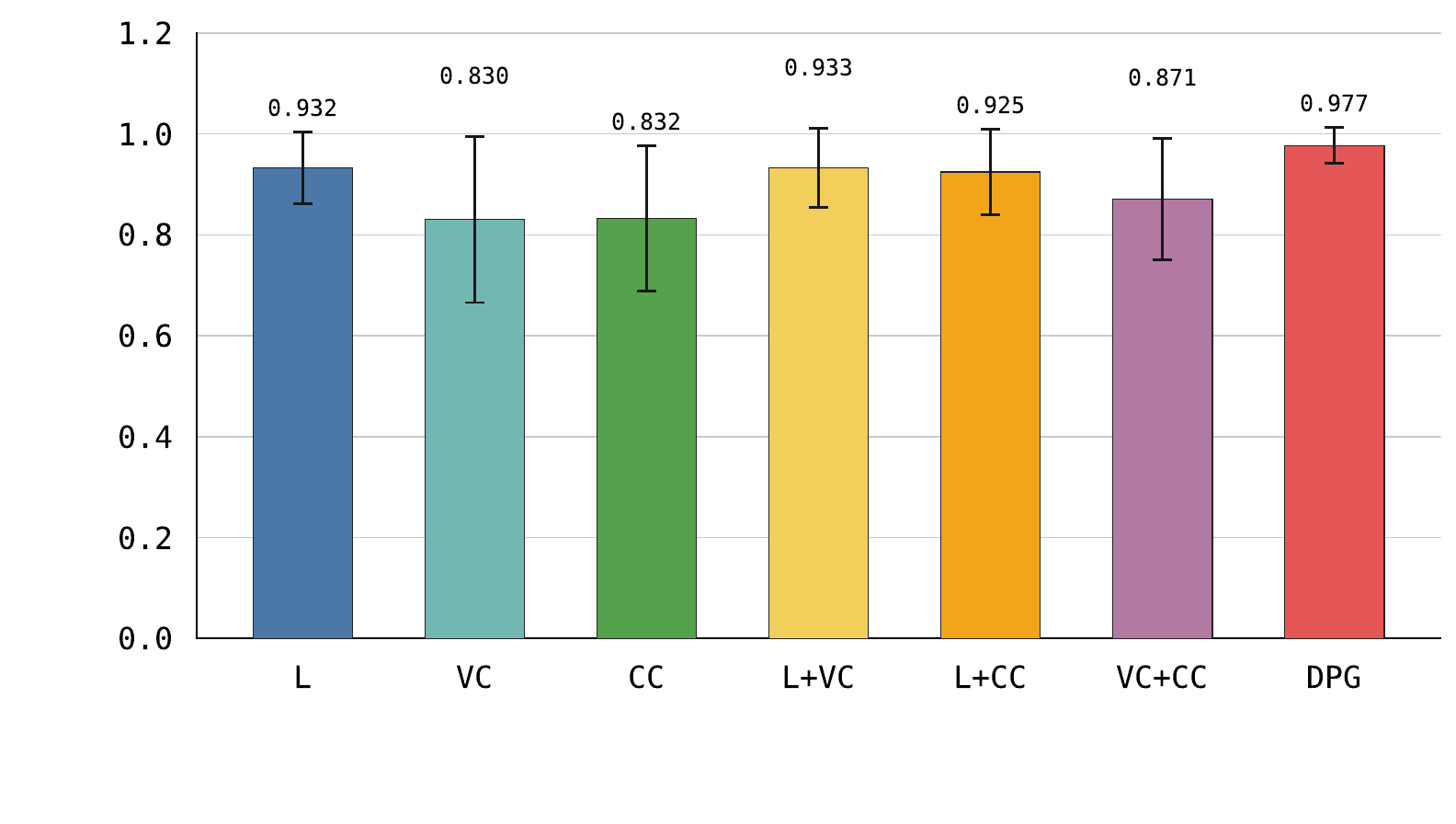}
    \end{minipage}}
    \hfill
    \subfloat[Boundary F1\label{fig:dpg_domain_boundary_f1}]{%
    \begin{minipage}[t]{0.485\textwidth}
        \centering
        \includegraphics[width=\linewidth]
        {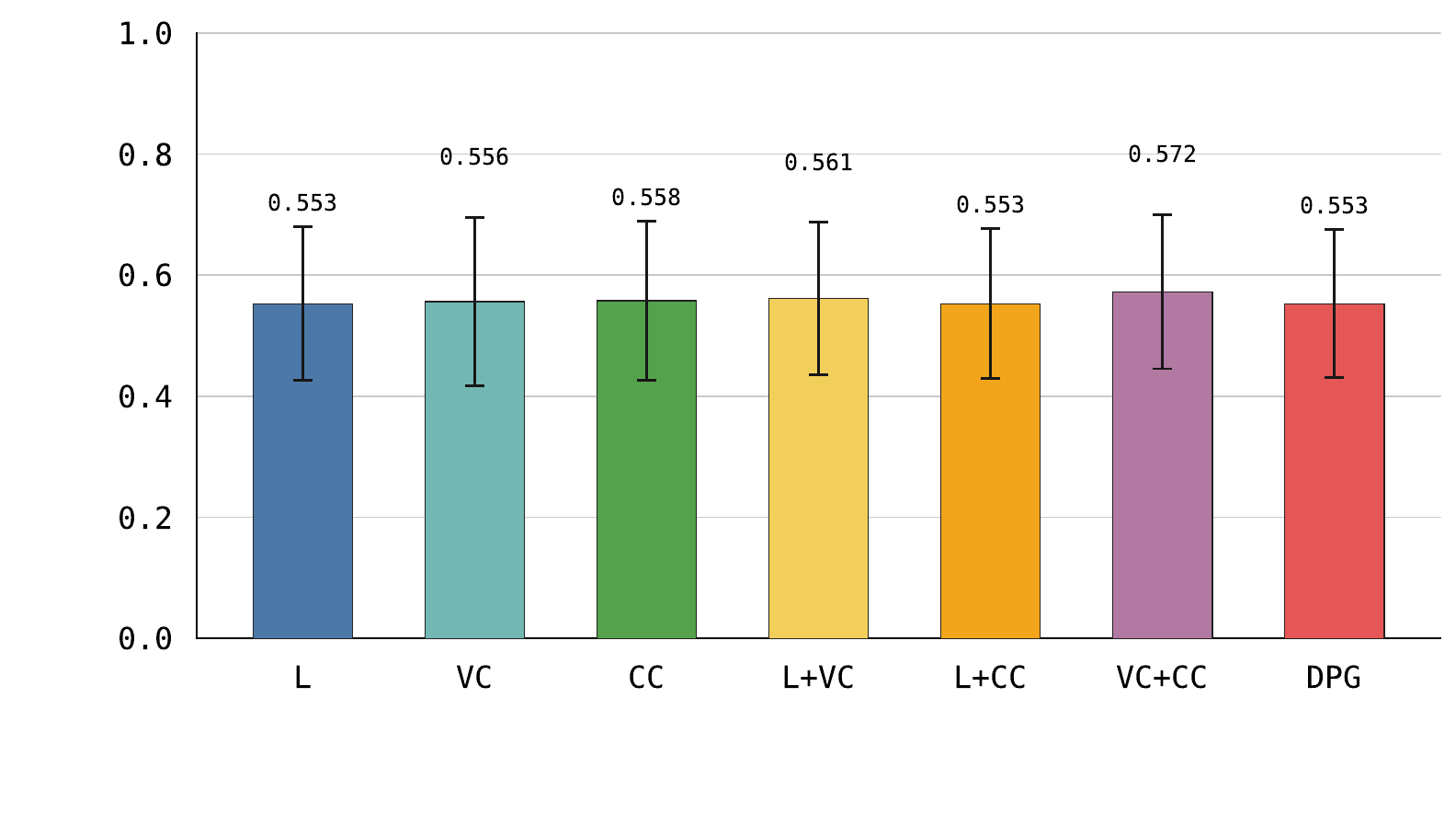}
    \end{minipage}}

    \vspace{0.4em}

    \subfloat[Number of groups\label{fig:dpg_domain_group_count}]{%
    \begin{minipage}[t]{0.485\textwidth}
        \centering
        \includegraphics[width=\linewidth]
        {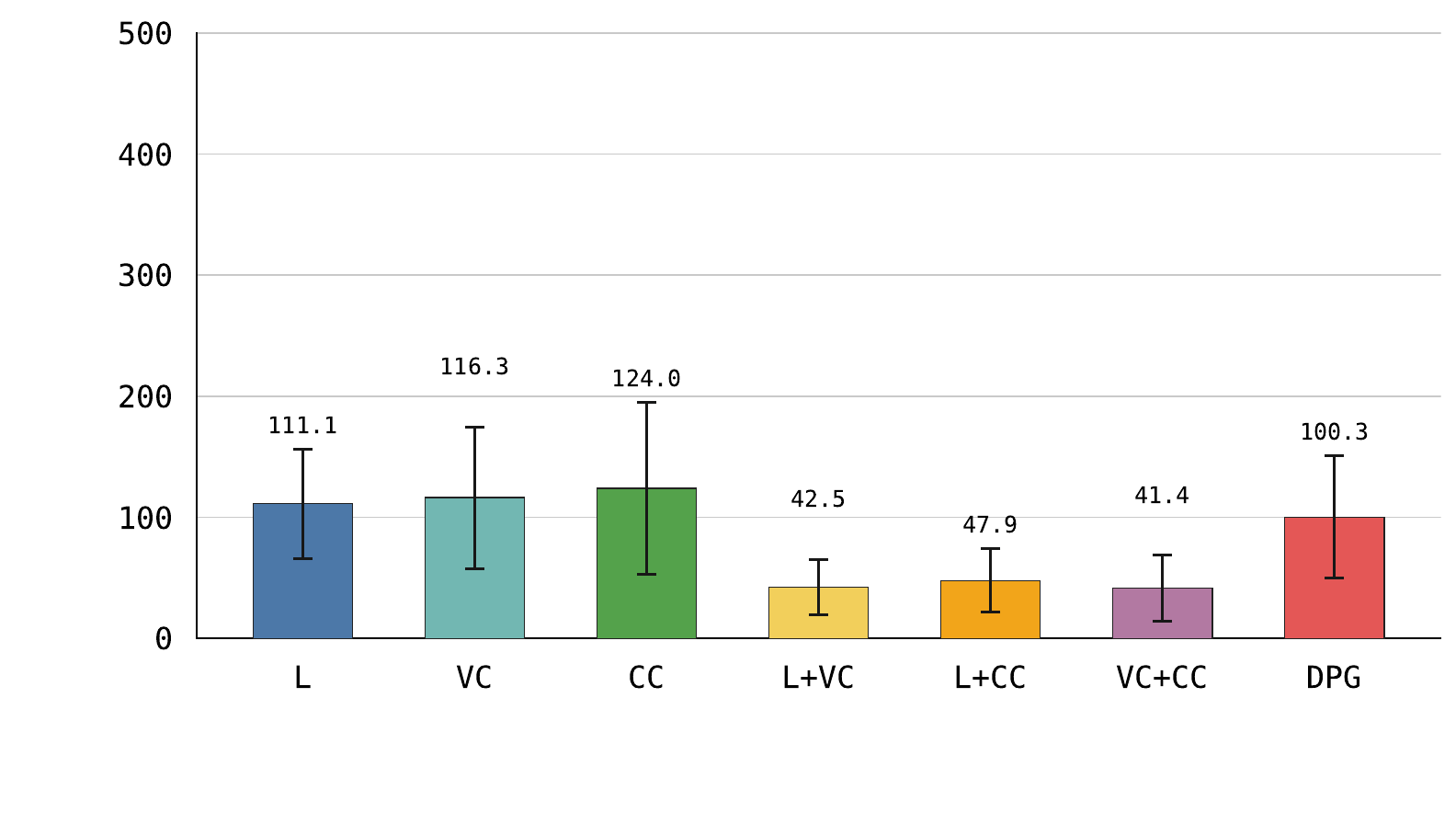}
    \end{minipage}}
    \hfill
    \subfloat[Runtime\label{fig:dpg_domain_runtime}]{%
    \begin{minipage}[t]{0.485\textwidth}
        \centering
        \includegraphics[width=\linewidth]
        {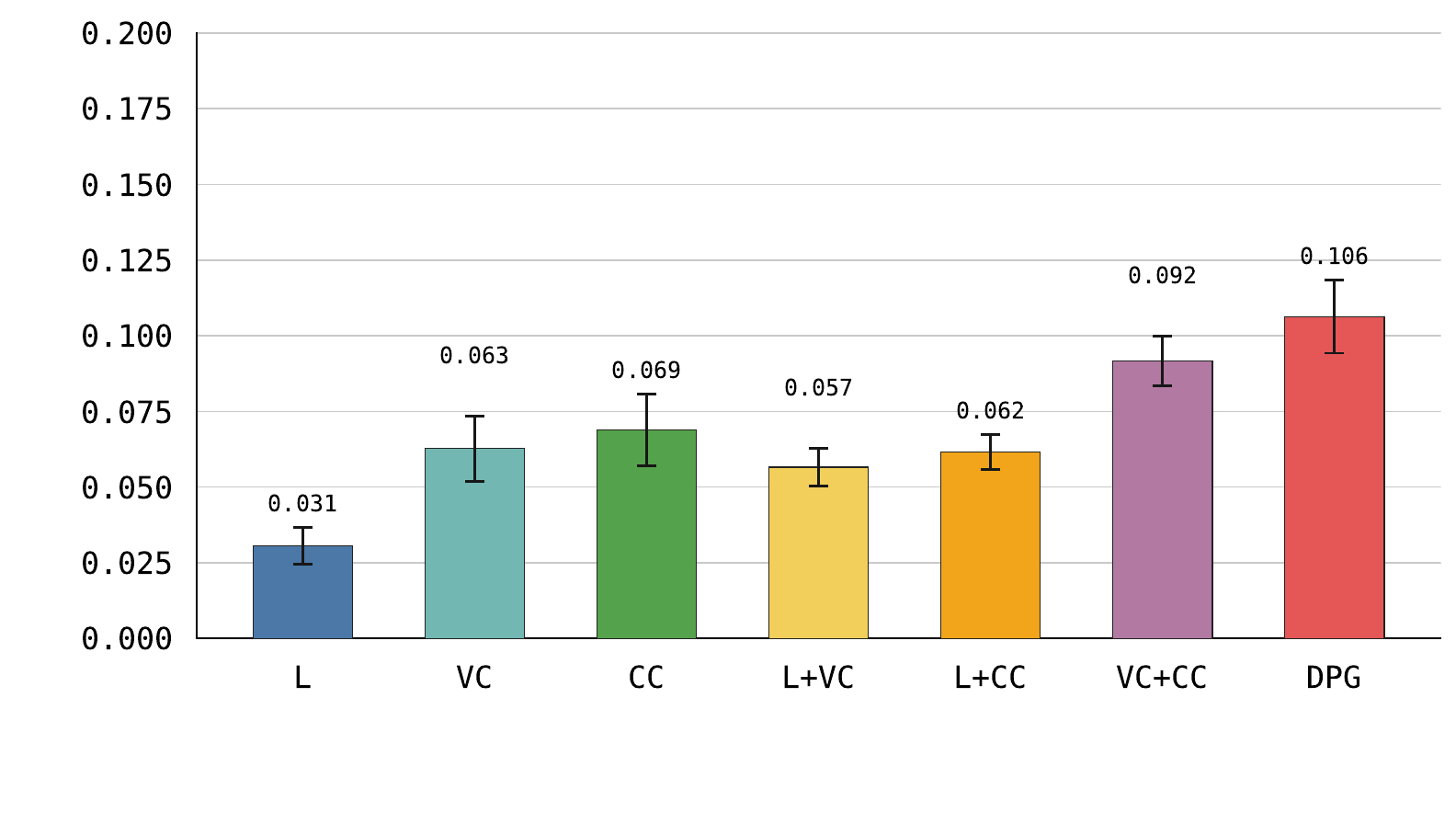}
    \end{minipage}}

    \caption{Comparison of the individual DPG domains, their pairwise
    combinations, and the complete three-domain DPG method over the 200
    BSDS500 test images. Each bar shows the mean of the 200 image-level
    values for the corresponding metric, and the error bars show the sample
    standard deviation across those images. L denotes the luminance domain,
    VC the vector-colour domain, and CC the colour-consistency domain. DPG
    denotes the complete combination of all three domains.}
    \label{fig:dpg_domain_comparison}
\end{figure}

Figure~\ref{fig:dpg_domain_comparison} summarizes the individual-domain,
pairwise-domain, and complete DPG results. Complete paired comparisons between
the three-domain DPG configuration and each component configuration are
reported in Table~\ref{tab:dpg-component-paired-statistics} in
Appendix~\ref{app:statistical-results}. \\

Among the individual domains, VC obtained the highest region covering
($0.4146 \pm 0.1191$), recall at IoU $0.50$
($0.1674 \pm 0.1081$), and recall at IoU $0.75$
($0.0489 \pm 0.0614$). CC obtained the highest boundary precision
($0.4433 \pm 0.1506$) and boundary F1
($0.5579 \pm 0.1311$), whereas L obtained the highest boundary recall
($0.9324 \pm 0.0713$).\\

Among the pairwise configurations, L+VC obtained the highest region covering
($0.4678 \pm 0.1120$), recall at IoU $0.50$
($0.1919 \pm 0.1133$), recall at IoU $0.75$
($0.0627 \pm 0.0622$), and boundary recall
($0.9325 \pm 0.0785$). VC+CC obtained the highest pairwise boundary precision
($0.4459 \pm 0.1450$) and boundary F1
($0.5725 \pm 0.1271$).\\

The complete DPG configuration obtained the highest region covering
($0.5251 \pm 0.1095$), recall at IoU $0.50$
($0.2729 \pm 0.1190$), and recall at IoU $0.75$
($0.0859 \pm 0.0739$). Relative to L+VC, which had the highest values among
the pairwise configurations for all three region measures, the mean paired
differences were $0.0573$ for covering, $0.0809$ for recall at IoU $0.50$,
and $0.0232$ for recall at IoU $0.75$. The complete configuration was
significantly higher in each region measure than every individual-domain and
pairwise-domain configuration after within-metric Holm correction
($p_{\mathrm{H}}<0.05$ for all comparisons).\\

The complete configuration obtained the highest boundary recall
($0.9772 \pm 0.0357$) and the lowest boundary precision
($0.3965 \pm 0.1213$) among the evaluated DPG configurations. Both differences
were statistically significant relative to every individual-domain and
pairwise-domain configuration ($p_{\mathrm{H}}<0.05$ for all comparisons).

Its boundary F1 was $0.5530 \pm 0.1219$. This value was significantly lower
than for L+VC ($0.5612 \pm 0.1260$,
$p_{\mathrm{H}}=0.0165729$) and VC+CC
($0.5725 \pm 0.1271$,
$p_{\mathrm{H}}=1.99702 \times 10^{-9}$), but did not differ significantly
from L, VC, CC, or L+CC.\\

The complete configuration produced $100.31 \pm 50.35$ groups per image.
The individual-domain configurations produced between
$111.14 \pm 45.31$ groups per image for L and
$124.02 \pm 70.85$ for CC, whereas the pairwise configurations produced
between $41.44 \pm 27.15$ for VC+CC and
$47.89 \pm 26.23$ for L+CC. The complete DPG runtime was
$0.1064 \pm 0.0121$ seconds per image.\\

Runtime does not increase monotonically with the number of domains retained
in the output because the configurations share processing stages and differ
in how their final proposal sets are constructed. VC and CC already use the
luminance processing result, even when L is not retained as an output domain.
In addition, single-domain configurations retain all connected components
from the selected domain, whereas multi-domain configurations retain the
parent groups formed through cross-domain relations. Consequently, for
example, L+VC produced fewer proposals
($42.52 \pm 22.75$) and had a lower measured runtime
($0.0566 \pm 0.0063$ s) than VC alone
($116.35 \pm 58.47$ proposals,
$0.0627 \pm 0.0108$ s).

\FloatBarrier
\subsection{Comparison with Conventional Methods}

\begin{figure}[!p]
    \centering

    \subfloat[Region covering\label{fig:dpg_baseline_covering}]{%
    \begin{minipage}[t]{0.485\textwidth}
        \centering
        \includegraphics[width=\linewidth]
        {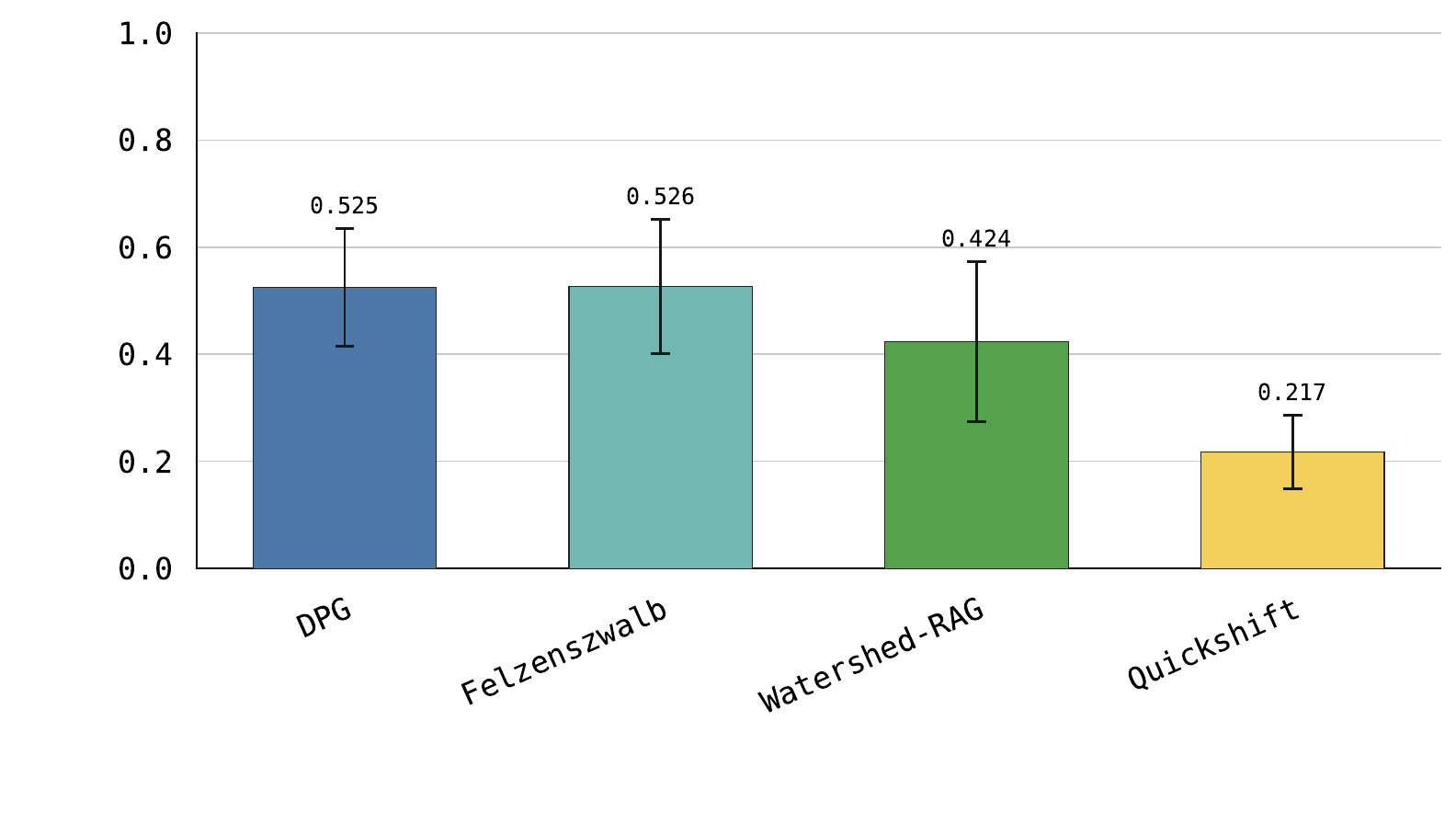}
    \end{minipage}}
    \hfill
    \subfloat[Region recall at IoU 0.50\label{fig:dpg_baseline_recall_050}]{%
    \begin{minipage}[t]{0.485\textwidth}
        \centering
        \includegraphics[width=\linewidth]
        {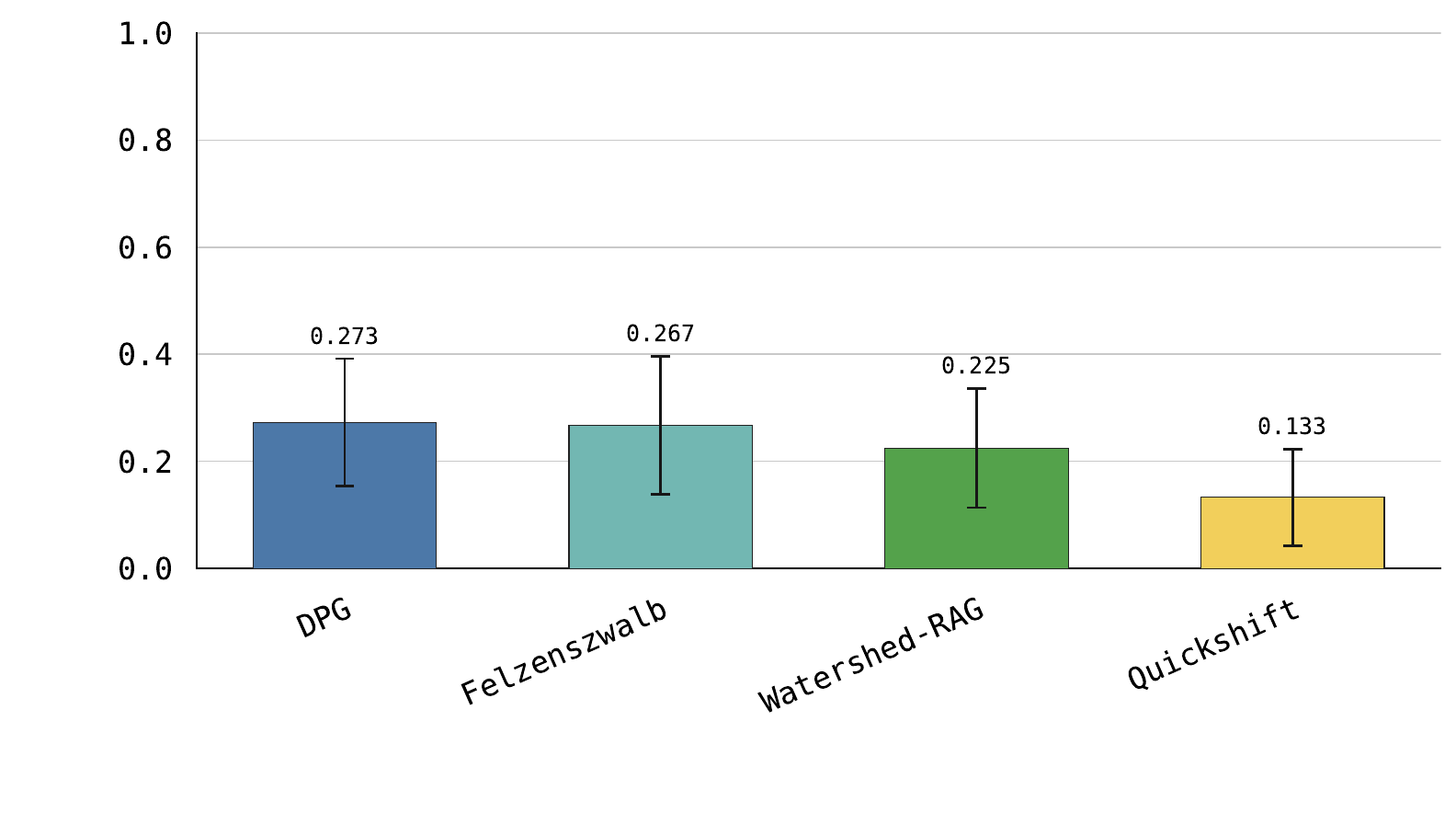}
    \end{minipage}}

    \vspace{0.4em}

    \subfloat[Region recall at IoU 0.75\label{fig:dpg_baseline_recall_075}]{%
    \begin{minipage}[t]{0.485\textwidth}
        \centering
        \includegraphics[width=\linewidth]
        {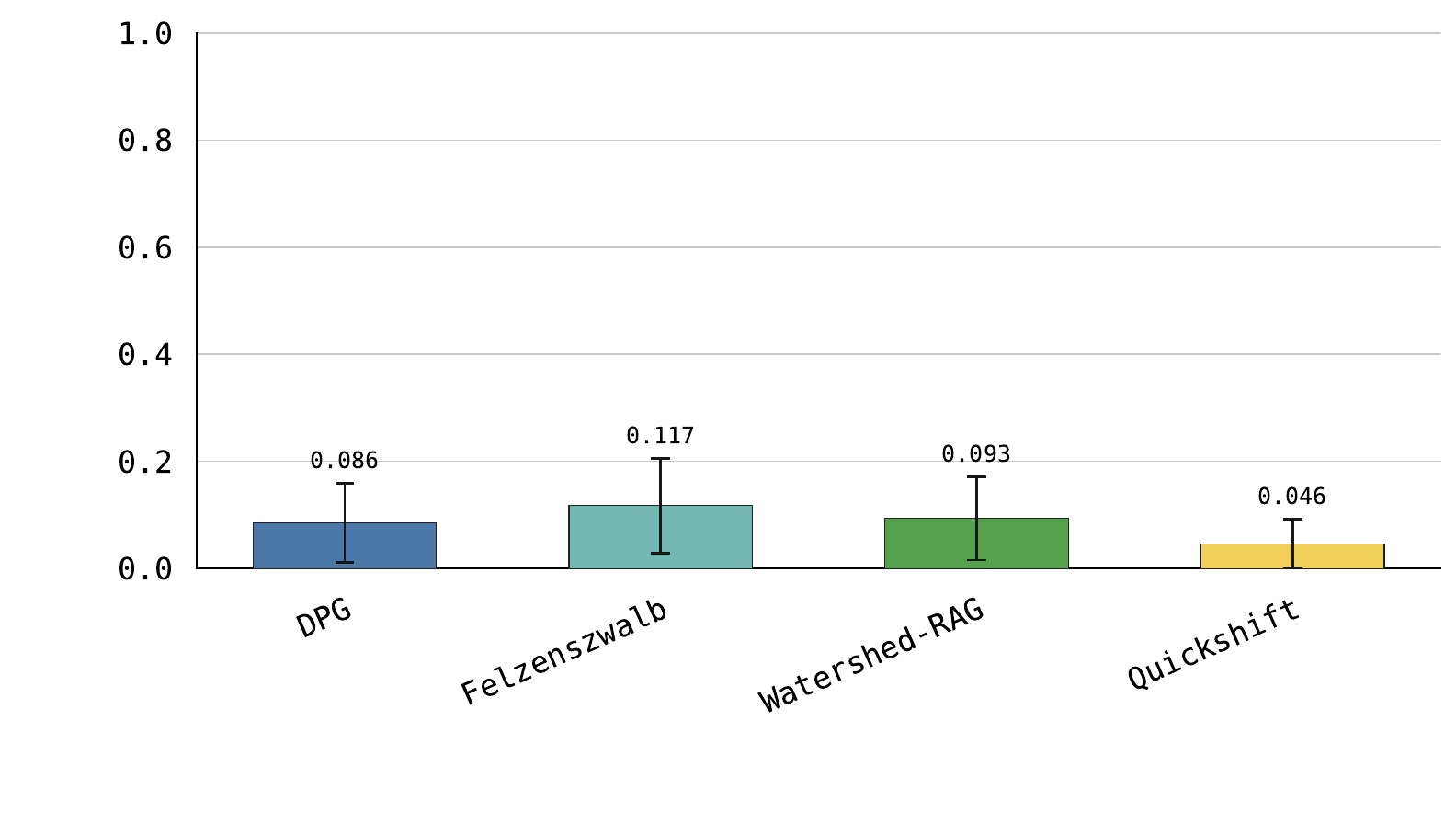}
    \end{minipage}}
    \hfill
    \subfloat[Boundary precision\label{fig:dpg_baseline_boundary_precision}]{%
    \begin{minipage}[t]{0.485\textwidth}
        \centering
        \includegraphics[width=\linewidth]
        {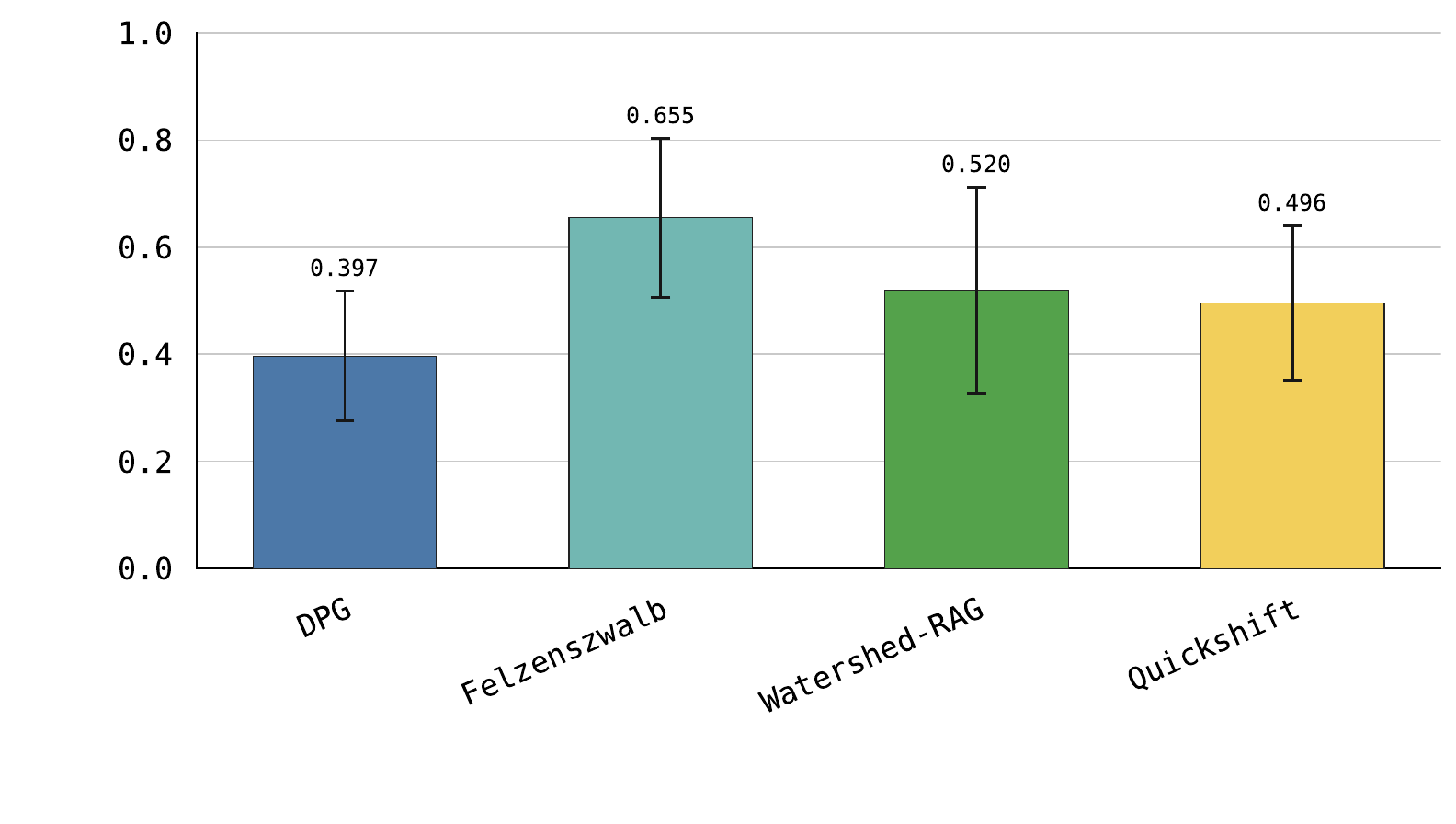}
    \end{minipage}}

    \vspace{0.4em}

    \subfloat[Boundary recall\label{fig:dpg_baseline_boundary_recall}]{%
    \begin{minipage}[t]{0.485\textwidth}
        \centering
        \includegraphics[width=\linewidth]
        {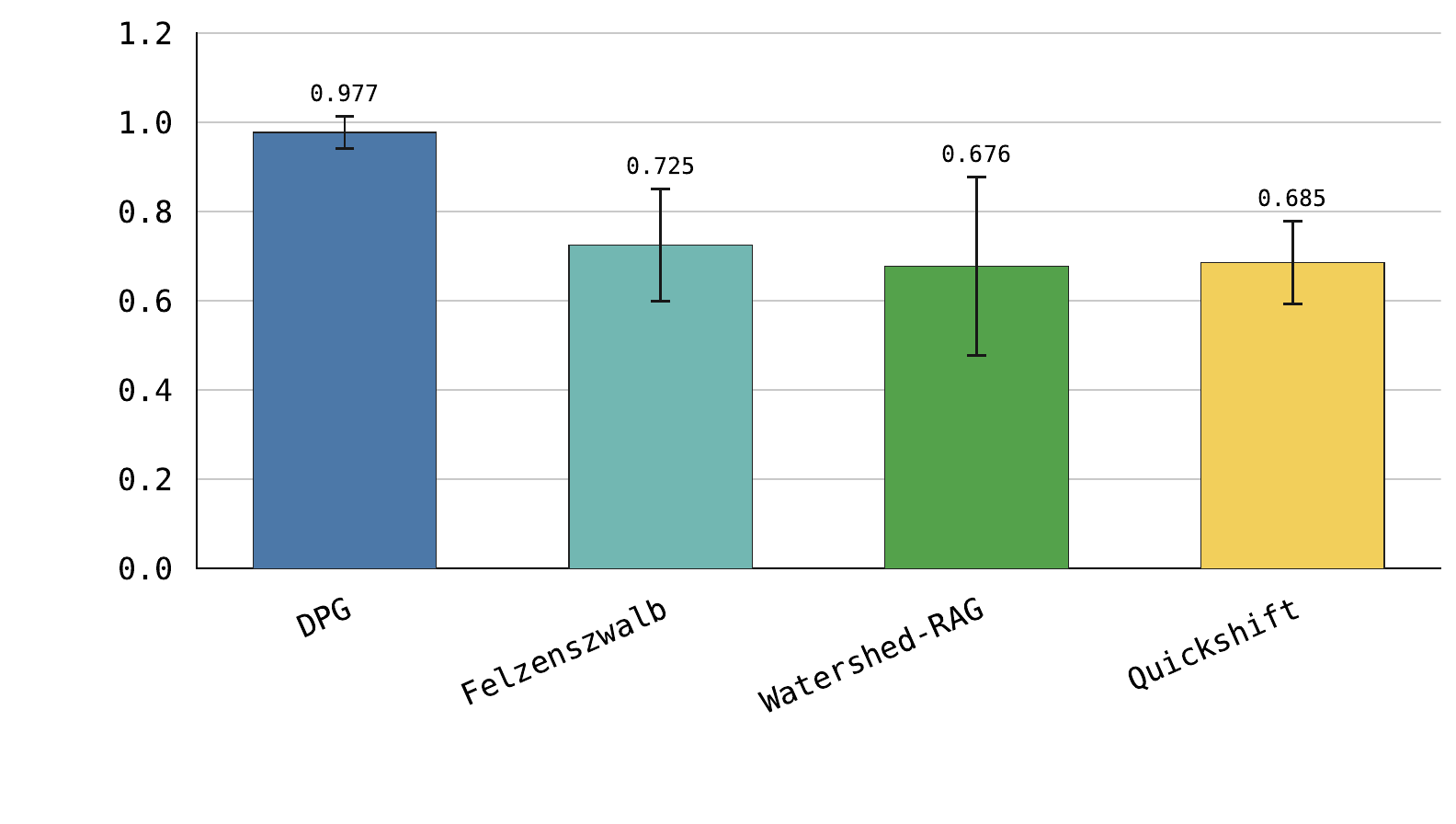}
    \end{minipage}}
    \hfill
    \subfloat[Boundary F1\label{fig:dpg_baseline_boundary_f1}]{%
    \begin{minipage}[t]{0.485\textwidth}
        \centering
        \includegraphics[width=\linewidth]
        {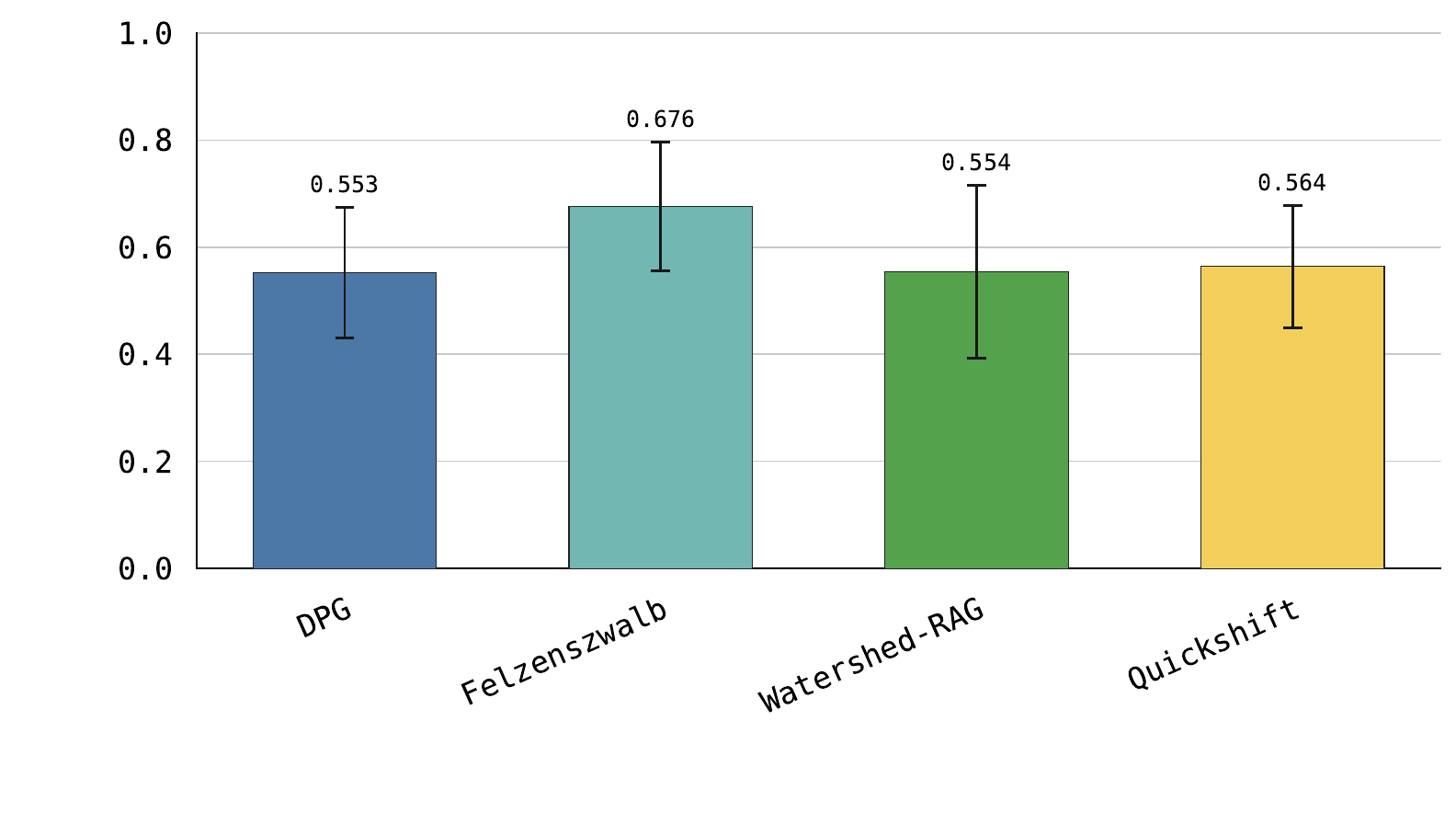}
    \end{minipage}}

    \vspace{0.4em}

    \subfloat[Number of groups\label{fig:dpg_baseline_group_count}]{%
    \begin{minipage}[t]{0.485\textwidth}
        \centering
        \includegraphics[width=\linewidth]
        {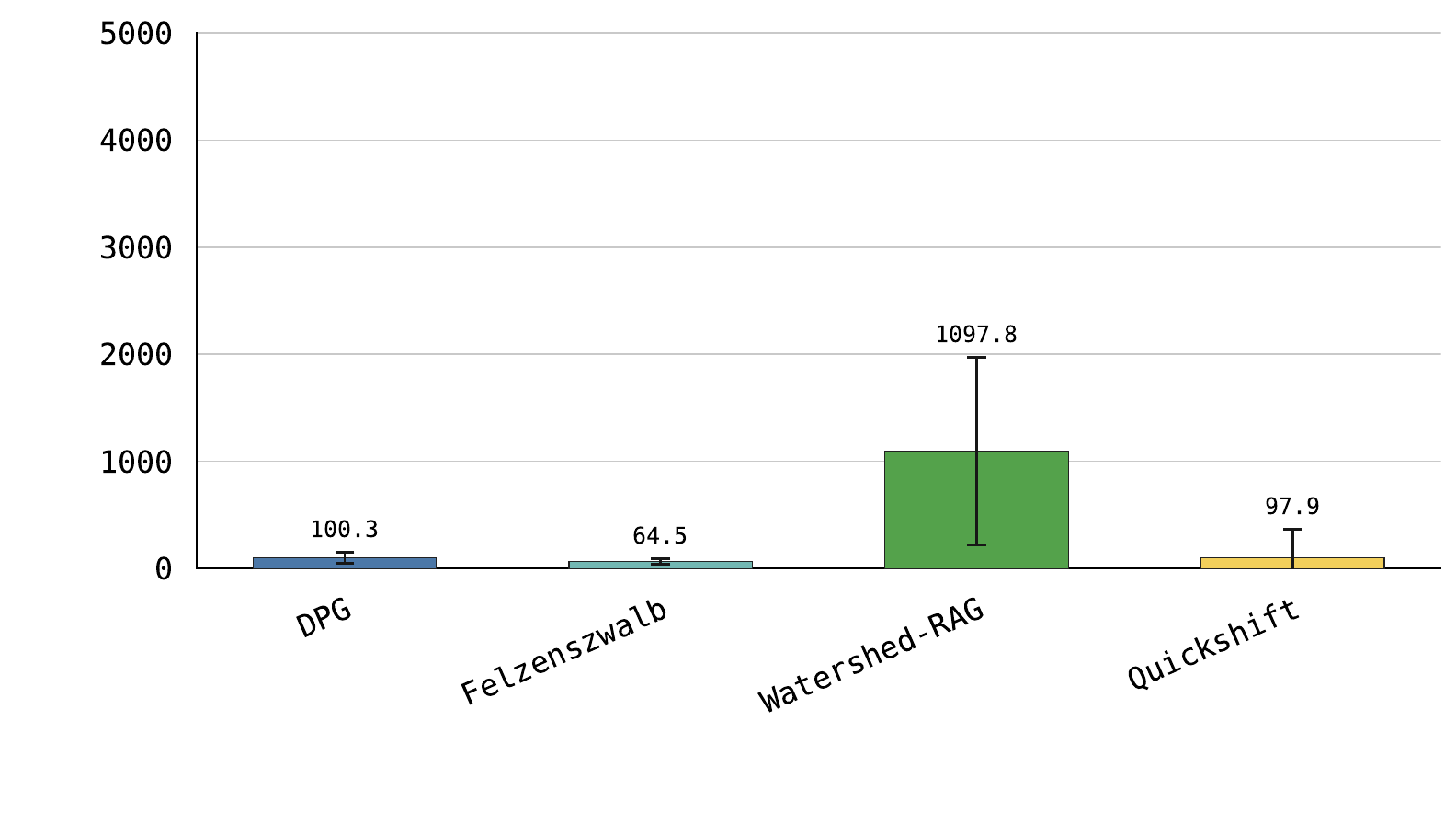}
    \end{minipage}}
    \hfill
    \subfloat[Runtime\label{fig:dpg_baseline_runtime}]{%
    \begin{minipage}[t]{0.485\textwidth}
        \centering
        \includegraphics[width=\linewidth]
        {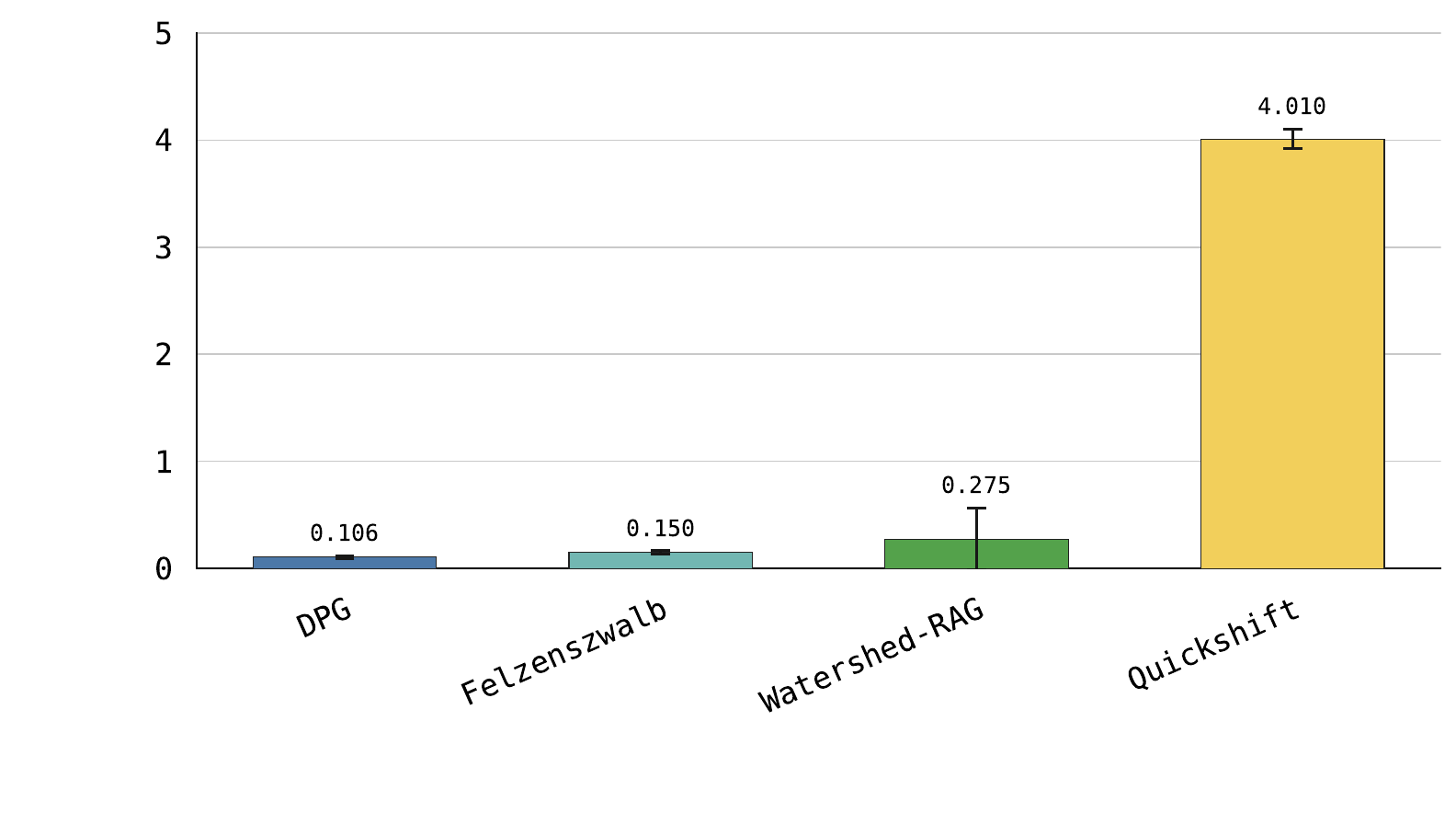}
    \end{minipage}}

    \caption{Comparison of the complete DPG method with Felzenszwalb
    segmentation, watershed with boundary-RAG merging, and Quickshift over
    the 200 BSDS500 test images. Each bar shows the mean of the 200
    image-level values for the corresponding metric, and the error bars show
    the sample standard deviation across those images.}
    \label{fig:dpg_baseline_comparison}
\end{figure}

Figure~\ref{fig:dpg_baseline_comparison} compares the complete DPG
configuration with the three conventional grouping methods. Complete paired
statistics, including mean differences, bootstrap confidence intervals,
Holm-adjusted $p$-values, and effect sizes, are reported in
Table~\ref{tab:dpg-baseline-paired-statistics} in
Appendix~\ref{app:statistical-results}. Statistical significance reported
below refers to these Holm-adjusted paired comparisons. 
DPG and Felzenszwalb obtained similar mean region covering values,
$0.5251 \pm 0.1095$ and $0.5260 \pm 0.1255$, respectively. Watershed-RAG
obtained $0.4238 \pm 0.1493$ and Quickshift
$0.2171 \pm 0.0689$. The DPG--Felzenszwalb difference was not statistically
significant, whereas DPG covering was significantly higher than for
watershed-RAG and Quickshift.\\

At IoU $0.50$, DPG obtained the highest mean region recall
($0.2729 \pm 0.1190$), followed by Felzenszwalb
($0.2674 \pm 0.1293$), watershed-RAG
($0.2247 \pm 0.1111$), and Quickshift
($0.1327 \pm 0.0900$). DPG did not differ significantly from Felzenszwalb,
but its recall was significantly higher than for watershed-RAG and
Quickshift.\\

At IoU $0.75$, Felzenszwalb obtained the highest mean region recall
($0.1173 \pm 0.0883$), followed by watershed-RAG
($0.0932 \pm 0.0775$), DPG
($0.0859 \pm 0.0739$), and Quickshift
($0.0461 \pm 0.0465$). DPG recall was significantly lower than for
Felzenszwalb, did not differ significantly from watershed-RAG, and was
significantly higher than for Quickshift.\\

DPG obtained the highest boundary recall
($0.9772 \pm 0.0357$), compared with
$0.7247 \pm 0.1251$ for Felzenszwalb,
$0.6765 \pm 0.2000$ for watershed-RAG, and
$0.6854 \pm 0.0928$ for Quickshift. DPG boundary recall was significantly
higher than for all three comparison methods.\\

Conversely, DPG obtained the lowest boundary precision
($0.3965 \pm 0.1213$), compared with
$0.6548 \pm 0.1486$ for Felzenszwalb,
$0.5198 \pm 0.1923$ for watershed-RAG, and
$0.4959 \pm 0.1448$ for Quickshift. DPG boundary precision was significantly
lower than for all three comparison methods.\\

Boundary F1 was $0.5530 \pm 0.1219$ for DPG,
$0.6764 \pm 0.1200$ for Felzenszwalb,
$0.5539 \pm 0.1613$ for watershed-RAG, and
$0.5641 \pm 0.1143$ for Quickshift. DPG F1 was significantly lower than for
Felzenszwalb and Quickshift, but did not differ significantly from
watershed-RAG.\\

DPG produced $100.31 \pm 50.35$ groups per image, compared with
$64.50 \pm 23.83$ for Felzenszwalb,
$1097.82 \pm 874.51$ for watershed-RAG, and
$97.89 \pm 266.33$ for Quickshift. DPG and Quickshift therefore had similar
mean group counts but substantially different between-image variation,
whereas watershed-RAG produced a much larger mean group count.\\

DPG had the lowest mean runtime at
$0.1064 \pm 0.0121$ seconds per image, compared with
$0.1504 \pm 0.0145$ seconds for Felzenszwalb,
$0.2749 \pm 0.2868$ seconds for watershed-RAG, and
$4.0096 \pm 0.0922$ seconds for Quickshift. DPG runtime was significantly
lower than for all three comparison methods.

\FloatBarrier
\section{Discussion}
\begin{figure}[!t]
    \centering
    \includegraphics[width=0.9\linewidth]
    {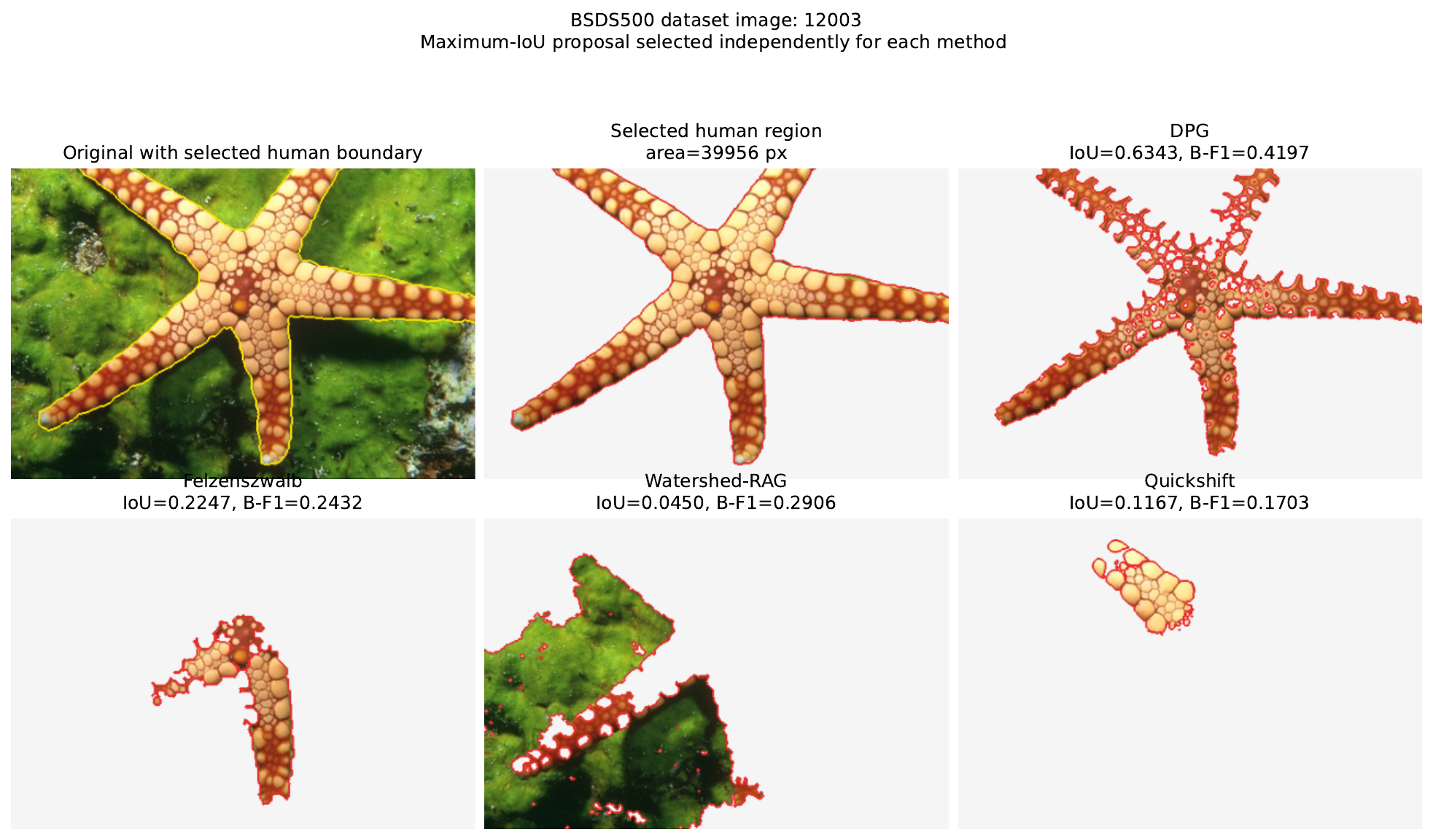}
    \caption{Illustrative training-set example using image 12003 and region 2
    from annotation 1. The top row shows the human-annotated region within
    the original image, the region separately, and the maximum-IoU DPG
    proposal. The bottom row shows the maximum-IoU proposals selected
    independently for the comparison methods. The IoU and B-F1 values shown
    in the panels are calculated only between the selected human region and
    the displayed proposal; they are not the image-level metrics used in the
    200-image test-set evaluation. The DPG proposal has the highest IoU with the selected reference region in
this illustrative example.}
    \label{fig:image12003}
\end{figure}

The results support the use of DPG as an intermediate visual representation
that retains multiple measurement-supported organizations of the same sensor
input without reducing them immediately to a single exclusive partition.
Across BSDS500, the complete DPG representation retained substantial
correspondence with the human-annotated regions, while the qualitative
examples showed that domain-specific, spatially overlapping groups can remain
available simultaneously. Figure~\ref{fig:image12003} provides an illustrative
example of this region correspondence for one selected training-set region.\\

\subsection{Complementary Contributions of the Domains}

The domain evaluation indicates that the three processing domains contribute
complementary grouping information. VC provided the strongest
individual-domain region correspondence, L the highest individual-domain
boundary recall, and CC the highest individual-domain boundary precision and
F1. Combining all three domains significantly increased each of the three
region measures relative to every individual and pairwise configuration and
also produced the highest boundary recall. This increase was accompanied by
lower boundary precision, while boundary F1 did not improve over the strongest
component configurations. The complete representation therefore expands the
annotation-corresponding structure available in the proposal set rather than
uniformly improving every segmentation-style measure.\\

\subsection{Relation to Conventional Segmentation Methods}

The comparison with conventional methods further separates the availability
of a corresponding group from the accuracy of its agreement with a particular
reference region. DPG and Felzenszwalb achieved similar region covering and
did not differ significantly in recall at IoU $0.50$. At the stricter IoU
$0.75$ threshold, however, Felzenszwalb achieved significantly higher recall.
DPG therefore retained approximately corresponding groups at a frequency
comparable to the graph-based partition at the lower threshold, while
Felzenszwalb more often produced the closer spatial agreement required by the
stricter threshold.\\

The boundary results reflect a corresponding trade-off. DPG recovered a larger
proportion of the annotated boundaries than the comparison methods but had
lower boundary precision. This is important when interpreting the result
because the evaluated DPG boundary map is the union of the boundaries of its
overlapping parent groups. A boundary without a nearby BSDS500 annotation
reduces precision regardless of whether it follows a visually distinguishable
but unannotated structure or an unsupported division.\\

\subsection{Interpretation of the DPG Representation}

The reference annotations also represent several alternative human
segmentations of each image but do not exhaustively describe the overlapping
subregions, illumination structures, markings, reflections, or textures that
DPG may retain. The quantitative evaluation therefore establishes
correspondence with the annotated scene regions and boundaries, but cannot by
itself determine the perceptual relevance of every additional DPG group or
boundary.\\

\begin{figure}[!t]
    \centering
    \includegraphics[width=0.9\linewidth]
    {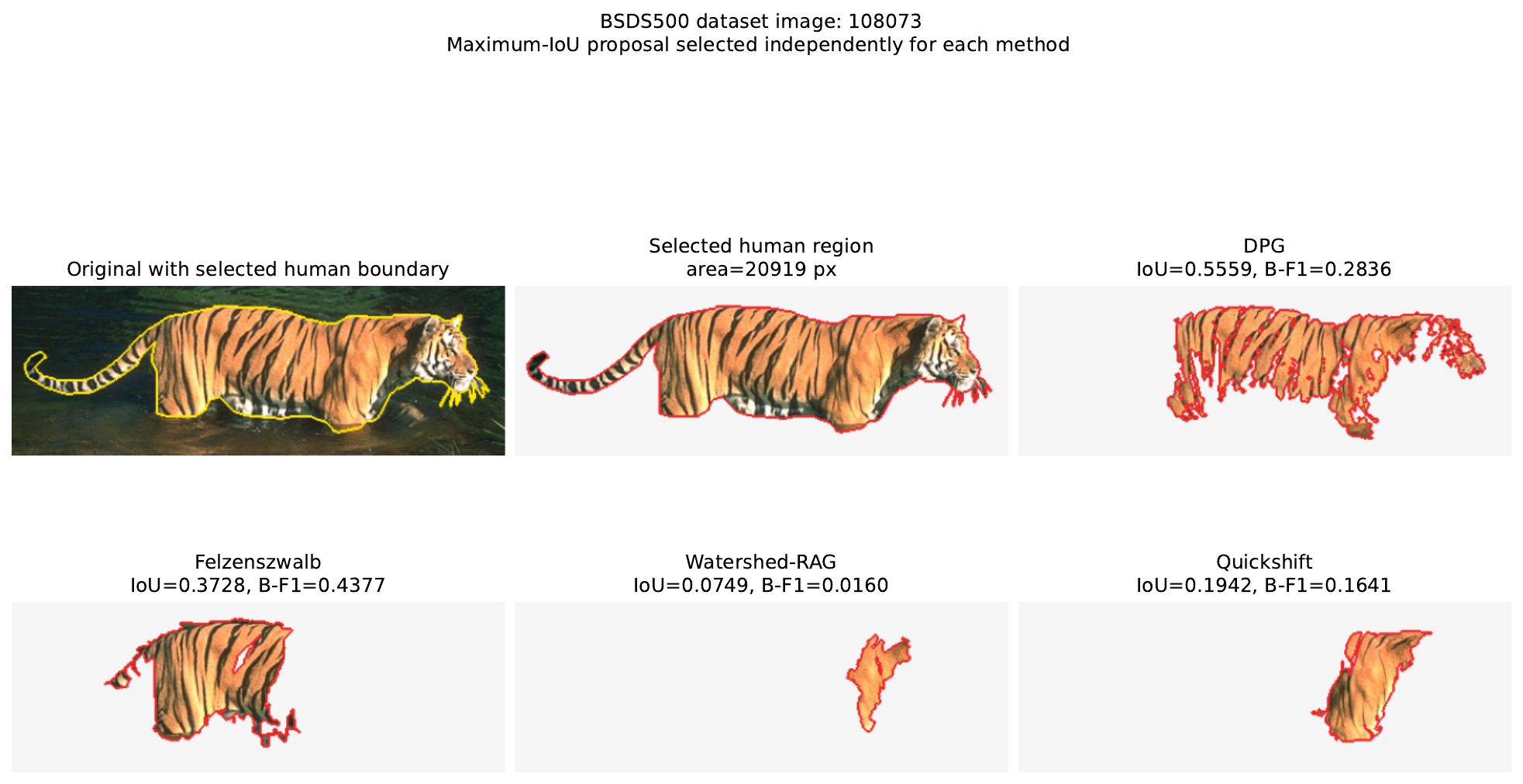}
    \caption{Illustrative training-set example using image 108073 and region 3
    from annotation 1. The top row shows the human-annotated region within
    the original image, the region separately, and the maximum-IoU DPG
    proposal. The bottom row shows the maximum-IoU proposals selected
    independently for the comparison methods. The IoU and B-F1 values shown
    in the panels are selected-region/proposal-pair measures and are not the
    image-level metrics used in the 200-image test-set evaluation. The
    selected DPG proposal follows the directly illuminated part of the tiger
    within the annotated region.}
    \label{fig:image108073}
\end{figure}

Figure~\ref{fig:image108073} illustrates this distinction for one selected
training-set region. The human annotation represents the broader tiger region,
whereas the selected DPG proposal follows the directly illuminated part within
it. The example shows how a measurement-supported subregion can be available
in the DPG representation even when it does not correspond to the complete
annotated region.\\

The qualitative results complement the dataset evaluation by showing the
organization retained within the DPG representation. Parent and child groups
can represent related structure at different spatial extents, and groups
supported by different measurement relationships can overlap rather than
forming one exclusive partition. Reprocessing additionally makes more
localized structure available without replacing the broader groups formed in
the earlier pass. Together, these examples demonstrate the two central
representational properties examined in this work: overlapping
domain-supported grouping and adjustable observational resolution.\\

\begin{figure}[!t]
    \centering
    \includegraphics[width=0.9\linewidth]
    {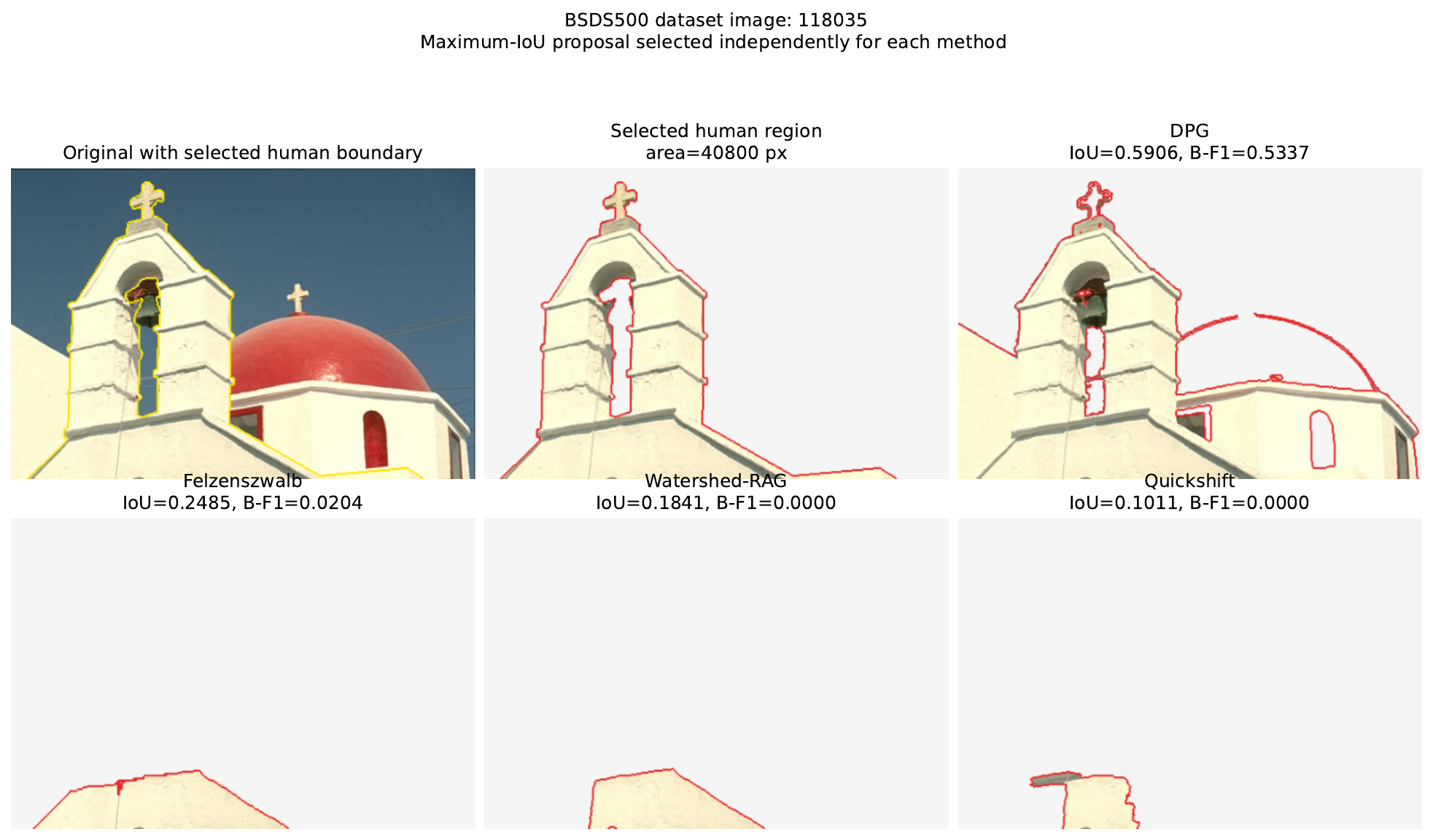}
    \caption{Illustrative training-set example using image 118035 and region 2
    from annotation 1. The top row shows the human-annotated region within
    the original image, the region separately, and the maximum-IoU DPG
    proposal. The bottom row shows the maximum-IoU proposals selected
    independently for the comparison methods. The IoU and B-F1 values shown
    in the panels are selected-region/proposal-pair measures and are not the
    image-level metrics used in the 200-image test-set evaluation. The DPG
    proposal contains the complete white building as one group.}
    \label{fig:image118035}
\end{figure}

Figure~\ref{fig:image118035} provides a separate example of a broad DPG group:
the selected proposal retains the complete white building within one group.
The qualitative reprocessing experiments show how such broad groups can
subsequently be examined at a different observational resolution while the
earlier group remains available. In the present experiments, the parent groups
used for reprocessing were selected manually; selection of which content
should be reprocessed automatically remains outside the scope of this work.\\

\subsection{Computational Characteristics}

Retaining overlapping groups did not result in an exceptionally large
proposal set in the present implementation: the complete DPG configuration
produced a mean group count similar to Quickshift, substantially fewer groups
than watershed-RAG, and more than Felzenszwalb. DPG also had the lowest
measured runtime among the evaluated methods in the common CPU execution
environment. These results show that the additional representational
structure examined here was obtained without a corresponding increase in
runtime relative to the comparison methods.\\

\subsection{Limitations and Future Work}

The present evaluation is limited to static RGB images and the three
implemented processing domains. The DPG parameters were selected qualitatively
by the method developer using development images that were separate from both
the BSDS500 test set and the qualitative experimental images, after which the
test-set configuration was kept fixed. This development process relied on one
evaluator. In addition, the BSDS500 annotations do not provide exhaustive
reference labels for the additional overlapping or measurement-supported
structures that DPG may produce. A multiple-observer evaluation would therefore
be useful for assessing how consistently such additional groups correspond to
visually distinguishable structure.\\

A further extension is temporal association. The present groups are formed
independently from individual images, whereas continuous observations could
provide additional evidence about persistence and relationships between
sensor-grounded groups. Associating DPG groups across time could allow later
processes to learn which structures recur, change together, or become relevant
through interaction. Combined with automatic selection for reprocessing, this
could extend observational resolution from the manually demonstrated operation
used here toward an active observation mechanism.\\

\section{Conclusions}

Overall, the results show that a non-exclusive, multi-domain grouping
representation can retain substantial correspondence with conventional
human-annotated scene regions while preserving additional
measurement-supported organization in the same observation. DPG forms these
groups without assigning semantic labels or object categories and without
requiring the image to be reduced to one mutually exclusive partition. The
retained parent groups, their cross-domain spatial relations, and the ability
to reprocess selected content provide an intermediate sensor-grounded
representation on which later processes could develop interpretations and
learn the relevance of the available perceptual units.

\clearpage
\input{appendix_statistics_complete}

\subsection*{Code, Data, and Materials Availability}
The source code used to produce the DPG results, together with the evaluation
and statistical-analysis code and the qualitative evaluation images used in
this study, is publicly available at \url{https://github.com/tjsauk/dpg-overlapping-visual-grouping}.
The BSDS500 dataset used for the quantitative evaluation is publicly available
from its original source \cite{Arbelaez2011}.

\bibliography{references}
\bibliographystyle{unsrt}

\end{document}

%% file: appendix_statistics_complete.tex

\appendix
\renewcommand{\theHsection}{appendix.\Alph{section}}
\clearpage
\onecolumn

\section{Complete paired statistical results}
\label{app:statistical-results}

This appendix reports the complete paired statistical results for the
component-configuration evaluation and for the comparison of complete DPG
with the three conventional grouping methods. All comparisons use the same
200 BSDS500 test images. For each metric and comparison, the tables report
the mean and sample standard deviation across images, the mean paired
difference, its 95\% paired-bootstrap confidence interval, the within-metric
Holm-adjusted Wilcoxon $p$-value, and the matched-pairs rank-biserial
correlation.\\

The paired difference $\Delta$ is defined as complete DPG minus the comparison
configuration or method. Positive values therefore indicate a larger DPG
value and negative values a smaller DPG value. For runtime, a negative
difference indicates faster DPG execution. Group count describes proposal-set
size and is not interpreted as an accuracy measure.

\clearpage
\begin{landscape}
\scriptsize
\setlength{\tabcolsep}{4pt}
\begin{longtable}{@{}llrrrrrr@{}}
\caption{Paired comparisons between complete DPG and the six component
configurations. For each metric, $\Delta$ is the mean paired image-level
difference defined as complete DPG minus the comparison configuration.
CI denotes the 95\% paired-bootstrap confidence interval,
$p_{\mathrm{H}}$ the within-metric Holm-adjusted Wilcoxon $p$-value, and
$r_{\mathrm{rb}}$ the matched-pairs rank-biserial correlation.}
\label{tab:dpg-component-paired-statistics}\\
\toprule
Metric & Comparison & Complete DPG mean $\pm$ SD & Comparison mean $\pm$ SD & $\Delta$ & 95\% CI & $p_{\mathrm{H}}$ & $r_{\mathrm{rb}}$ \\
\midrule
\endfirsthead
\multicolumn{8}{c}{\tablename\ \thetable\ -- continued} \\
\toprule
Metric & Comparison & Complete DPG mean $\pm$ SD & Comparison mean $\pm$ SD & $\Delta$ & 95\% CI & $p_{\mathrm{H}}$ & $r_{\mathrm{rb}}$ \\
\midrule
\endhead
\midrule
\multicolumn{8}{r}{Continued on next page} \\
\endfoot
\bottomrule
\endlastfoot
Covering & L & 0.5251 $\pm$ 0.1095 & 0.3644 $\pm$ 0.1021 & 0.1607 & [0.1484, 0.1737] & $8.61688\times 10^{-34}$ & 0.994 \\
Covering & VC & 0.5251 $\pm$ 0.1095 & 0.4146 $\pm$ 0.1191 & 0.1104 & [0.1003, 0.1212] & $8.61688\times 10^{-34}$ & 1.000 \\
Covering & CC & 0.5251 $\pm$ 0.1095 & 0.3888 $\pm$ 0.1229 & 0.1363 & [0.1240, 0.1489] & $8.61688\times 10^{-34}$ & 1.000 \\
Covering & L+VC & 0.5251 $\pm$ 0.1095 & 0.4678 $\pm$ 0.1120 & 0.0573 & [0.0497, 0.0654] & $8.61688\times 10^{-34}$ & 1.000 \\
Covering & L+CC & 0.5251 $\pm$ 0.1095 & 0.4469 $\pm$ 0.1122 & 0.0782 & [0.0692, 0.0874] & $8.61688\times 10^{-34}$ & 1.000 \\
Covering & VC+CC & 0.5251 $\pm$ 0.1095 & 0.4621 $\pm$ 0.1187 & 0.0629 & [0.0552, 0.0711] & $8.61688\times 10^{-34}$ & 1.000 \\
\addlinespace[2pt]
Recall@0.50 & L & 0.2729 $\pm$ 0.1190 & 0.1506 $\pm$ 0.0955 & 0.1223 & [0.1100, 0.1352] & $1.24817\times 10^{-32}$ & 0.995 \\
Recall@0.50 & VC & 0.2729 $\pm$ 0.1190 & 0.1674 $\pm$ 0.1081 & 0.1055 & [0.0943, 0.1173] & $4.95889\times 10^{-31}$ & 0.974 \\
Recall@0.50 & CC & 0.2729 $\pm$ 0.1190 & 0.1558 $\pm$ 0.1072 & 0.1170 & [0.1041, 0.1312] & $2.61845\times 10^{-31}$ & 0.974 \\
Recall@0.50 & L+VC & 0.2729 $\pm$ 0.1190 & 0.1919 $\pm$ 0.1133 & 0.0809 & [0.0721, 0.0904] & $6.66298\times 10^{-32}$ & 1.000 \\
Recall@0.50 & L+CC & 0.2729 $\pm$ 0.1190 & 0.1810 $\pm$ 0.1095 & 0.0919 & [0.0814, 0.1029] & $4.95889\times 10^{-31}$ & 1.000 \\
Recall@0.50 & VC+CC & 0.2729 $\pm$ 0.1190 & 0.1800 $\pm$ 0.1033 & 0.0929 & [0.0831, 0.1032] & $1.65492\times 10^{-31}$ & 1.000 \\
\addlinespace[2pt]
Recall@0.75 & L & 0.0859 $\pm$ 0.0739 & 0.0388 $\pm$ 0.0462 & 0.0471 & [0.0392, 0.0555] & $1.16569\times 10^{-23}$ & 0.934 \\
Recall@0.75 & VC & 0.0859 $\pm$ 0.0739 & 0.0489 $\pm$ 0.0614 & 0.0369 & [0.0311, 0.0428] & $3.08668\times 10^{-23}$ & 0.954 \\
Recall@0.75 & CC & 0.0859 $\pm$ 0.0739 & 0.0415 $\pm$ 0.0564 & 0.0444 & [0.0370, 0.0524] & $2.55314\times 10^{-24}$ & 0.974 \\
Recall@0.75 & L+VC & 0.0859 $\pm$ 0.0739 & 0.0627 $\pm$ 0.0622 & 0.0232 & [0.0188, 0.0280] & $1.31258\times 10^{-20}$ & 1.000 \\
Recall@0.75 & L+CC & 0.0859 $\pm$ 0.0739 & 0.0517 $\pm$ 0.0630 & 0.0342 & [0.0280, 0.0407] & $8.65895\times 10^{-22}$ & 1.000 \\
Recall@0.75 & VC+CC & 0.0859 $\pm$ 0.0739 & 0.0582 $\pm$ 0.0661 & 0.0277 & [0.0228, 0.0327] & $6.08732\times 10^{-22}$ & 1.000 \\
\addlinespace[2pt]
Boundary precision & L & 0.3965 $\pm$ 0.1213 & 0.4056 $\pm$ 0.1283 & -0.0090 & [-0.0146, -0.0034] & $1.50739\times 10^{-4}$ & -0.309 \\
Boundary precision & VC & 0.3965 $\pm$ 0.1213 & 0.4326 $\pm$ 0.1386 & -0.0361 & [-0.0456, -0.0268] & $9.70891\times 10^{-12}$ & -0.571 \\
Boundary precision & CC & 0.3965 $\pm$ 0.1213 & 0.4433 $\pm$ 0.1506 & -0.0468 & [-0.0569, -0.0371] & $4.51538\times 10^{-19}$ & -0.742 \\
Boundary precision & L+VC & 0.3965 $\pm$ 0.1213 & 0.4148 $\pm$ 0.1306 & -0.0183 & [-0.0241, -0.0128] & $6.57272\times 10^{-10}$ & -0.518 \\
Boundary precision & L+CC & 0.3965 $\pm$ 0.1213 & 0.4095 $\pm$ 0.1330 & -0.0130 & [-0.0183, -0.0080] & $2.47028\times 10^{-7}$ & -0.431 \\
Boundary precision & VC+CC & 0.3965 $\pm$ 0.1213 & 0.4459 $\pm$ 0.1450 & -0.0493 & [-0.0577, -0.0414] & $5.41261\times 10^{-25}$ & -0.856 \\
\addlinespace[2pt]
Boundary recall & L & 0.9772 $\pm$ 0.0357 & 0.9324 $\pm$ 0.0713 & 0.0448 & [0.0365, 0.0535] & $4.26946\times 10^{-26}$ & 0.864 \\
Boundary recall & VC & 0.9772 $\pm$ 0.0357 & 0.8304 $\pm$ 0.1647 & 0.1468 & [0.1259, 0.1688] & $8.61688\times 10^{-34}$ & 0.994 \\
Boundary recall & CC & 0.9772 $\pm$ 0.0357 & 0.8318 $\pm$ 0.1439 & 0.1454 & [0.1271, 0.1641] & $8.61688\times 10^{-34}$ & 1.000 \\
Boundary recall & L+VC & 0.9772 $\pm$ 0.0357 & 0.9325 $\pm$ 0.0785 & 0.0447 & [0.0362, 0.0537] & $8.61688\times 10^{-34}$ & 1.000 \\
Boundary recall & L+CC & 0.9772 $\pm$ 0.0357 & 0.9246 $\pm$ 0.0847 & 0.0526 & [0.0433, 0.0630] & $8.61688\times 10^{-34}$ & 1.000 \\
Boundary recall & VC+CC & 0.9772 $\pm$ 0.0357 & 0.8708 $\pm$ 0.1199 & 0.1064 & [0.0917, 0.1218] & $8.61688\times 10^{-34}$ & 1.000 \\
\addlinespace[2pt]
Boundary F1 & L & 0.5530 $\pm$ 0.1219 & 0.5531 $\pm$ 0.1272 & -0.0001 & [-0.0058, 0.0058] & 0.906022 & -0.049 \\
Boundary F1 & VC & 0.5530 $\pm$ 0.1219 & 0.5562 $\pm$ 0.1392 & -0.0033 & [-0.0133, 0.0069] & 0.470856 & -0.115 \\
Boundary F1 & CC & 0.5530 $\pm$ 0.1219 & 0.5579 $\pm$ 0.1311 & -0.0049 & [-0.0152, 0.0052] & 0.210673 & -0.158 \\
Boundary F1 & L+VC & 0.5530 $\pm$ 0.1219 & 0.5612 $\pm$ 0.1260 & -0.0082 & [-0.0133, -0.0031] & 0.0165729 & -0.240 \\
Boundary F1 & L+CC & 0.5530 $\pm$ 0.1219 & 0.5531 $\pm$ 0.1236 & -0.0001 & [-0.0051, 0.0049] & 0.906022 & -0.061 \\
Boundary F1 & VC+CC & 0.5530 $\pm$ 0.1219 & 0.5725 $\pm$ 0.1271 & -0.0195 & [-0.0264, -0.0126] & $1.99702\times 10^{-9}$ & -0.512 \\
\addlinespace[2pt]
Group count & L & 100.31 $\pm$ 50.35 & 111.14 $\pm$ 45.31 & -10.83 & [-15.66, -6.12] & $3.50305\times 10^{-6}$ & -0.381 \\
Group count & VC & 100.31 $\pm$ 50.35 & 116.35 $\pm$ 58.47 & -16.05 & [-21.98, -10.01] & $1.00003\times 10^{-6}$ & -0.413 \\
Group count & CC & 100.31 $\pm$ 50.35 & 124.02 $\pm$ 70.85 & -23.71 & [-30.44, -17.18] & $8.68143\times 10^{-10}$ & -0.518 \\
Group count & L+VC & 100.31 $\pm$ 50.35 & 42.52 $\pm$ 22.75 & 57.79 & [53.35, 62.66] & $8.56038\times 10^{-34}$ & 1.000 \\
Group count & L+CC & 100.31 $\pm$ 50.35 & 47.89 $\pm$ 26.23 & 52.41 & [48.38, 56.64] & $8.56038\times 10^{-34}$ & 1.000 \\
Group count & VC+CC & 100.31 $\pm$ 50.35 & 41.44 $\pm$ 27.15 & 58.87 & [55.16, 62.81] & $8.56038\times 10^{-34}$ & 1.000 \\
\addlinespace[2pt]
Runtime (s) & L & 0.1064 $\pm$ 0.0121 & 0.0306 $\pm$ 0.0061 & 0.0757 & [0.0744, 0.0771] & $8.61688\times 10^{-34}$ & 1.000 \\
Runtime (s) & VC & 0.1064 $\pm$ 0.0121 & 0.0627 $\pm$ 0.0108 & 0.0436 & [0.0425, 0.0448] & $8.61688\times 10^{-34}$ & 1.000 \\
Runtime (s) & CC & 0.1064 $\pm$ 0.0121 & 0.0689 $\pm$ 0.0118 & 0.0374 & [0.0361, 0.0388] & $8.61688\times 10^{-34}$ & 1.000 \\
Runtime (s) & L+VC & 0.1064 $\pm$ 0.0121 & 0.0566 $\pm$ 0.0063 & 0.0498 & [0.0487, 0.0509] & $8.61688\times 10^{-34}$ & 1.000 \\
Runtime (s) & L+CC & 0.1064 $\pm$ 0.0121 & 0.0617 $\pm$ 0.0057 & 0.0447 & [0.0435, 0.0459] & $8.61688\times 10^{-34}$ & 1.000 \\
Runtime (s) & VC+CC & 0.1064 $\pm$ 0.0121 & 0.0917 $\pm$ 0.0082 & 0.0146 & [0.0139, 0.0155] & $8.61688\times 10^{-34}$ & 1.000 \\
\end{longtable}
\end{landscape}

\clearpage
\begin{landscape}
\scriptsize
\setlength{\tabcolsep}{4pt}
\begin{longtable}{@{}llrrrrrr@{}}
\caption{Paired comparisons between complete DPG and the three conventional
grouping methods. For each metric, $\Delta$ is the mean paired image-level
difference defined as complete DPG minus the comparison method. CI denotes
the 95\% paired-bootstrap confidence interval, $p_{\mathrm{H}}$ the
within-metric Holm-adjusted Wilcoxon $p$-value, and $r_{\mathrm{rb}}$ the
matched-pairs rank-biserial correlation.}
\label{tab:dpg-baseline-paired-statistics}\\
\toprule
Metric & Comparison & Complete DPG mean $\pm$ SD & Comparison mean $\pm$ SD & $\Delta$ & 95\% CI & $p_{\mathrm{H}}$ & $r_{\mathrm{rb}}$ \\
\midrule
\endfirsthead
\multicolumn{8}{c}{\tablename\ \thetable\ -- continued} \\
\toprule
Metric & Comparison & Complete DPG mean $\pm$ SD & Comparison mean $\pm$ SD & $\Delta$ & 95\% CI & $p_{\mathrm{H}}$ & $r_{\mathrm{rb}}$ \\
\midrule
\endhead
\midrule
\multicolumn{8}{r}{Continued on next page} \\
\endfoot
\bottomrule
\endlastfoot
Covering & Felzenszwalb & 0.5251 $\pm$ 0.1095 & 0.5260 $\pm$ 0.1255 & -0.0010 & [-0.0167, 0.0141] & 0.77151 & 0.024 \\
Covering & Watershed--RAG & 0.5251 $\pm$ 0.1095 & 0.4238 $\pm$ 0.1493 & 0.1013 & [0.0818, 0.1211] & $1.06555\times 10^{-17}$ & 0.705 \\
Covering & Quickshift & 0.5251 $\pm$ 0.1095 & 0.2171 $\pm$ 0.0689 & 0.3079 & [0.2896, 0.3265] & $4.30844\times 10^{-34}$ & 1.000 \\
\addlinespace[2pt]
Recall@0.50 & Felzenszwalb & 0.2729 $\pm$ 0.1190 & 0.2674 $\pm$ 0.1293 & 0.0054 & [-0.0107, 0.0213] & 0.336721 & 0.081 \\
Recall@0.50 & Watershed--RAG & 0.2729 $\pm$ 0.1190 & 0.2247 $\pm$ 0.1111 & 0.0481 & [0.0311, 0.0653] & $3.74230\times 10^{-7}$ & 0.432 \\
Recall@0.50 & Quickshift & 0.2729 $\pm$ 0.1190 & 0.1327 $\pm$ 0.0900 & 0.1402 & [0.1228, 0.1581] & $1.62001\times 10^{-29}$ & 0.932 \\
\addlinespace[2pt]
Recall@0.75 & Felzenszwalb & 0.0859 $\pm$ 0.0739 & 0.1173 $\pm$ 0.0883 & -0.0315 & [-0.0429, -0.0201] & $2.14720\times 10^{-7}$ & -0.453 \\
Recall@0.75 & Watershed--RAG & 0.0859 $\pm$ 0.0739 & 0.0932 $\pm$ 0.0775 & -0.0074 & [-0.0180, 0.0031] & 0.0554478 & -0.166 \\
Recall@0.75 & Quickshift & 0.0859 $\pm$ 0.0739 & 0.0461 $\pm$ 0.0465 & 0.0397 & [0.0288, 0.0517] & $1.92024\times 10^{-9}$ & 0.536 \\
\addlinespace[2pt]
Boundary precision & Felzenszwalb & 0.3965 $\pm$ 0.1213 & 0.6548 $\pm$ 0.1486 & -0.2583 & [-0.2711, -0.2455] & $4.30844\times 10^{-34}$ & -1.000 \\
Boundary precision & Watershed--RAG & 0.3965 $\pm$ 0.1213 & 0.5198 $\pm$ 0.1923 & -0.1233 & [-0.1418, -0.1048] & $1.53187\times 10^{-24}$ & -0.834 \\
Boundary precision & Quickshift & 0.3965 $\pm$ 0.1213 & 0.4959 $\pm$ 0.1448 & -0.0993 & [-0.1109, -0.0874] & $3.06129\times 10^{-27}$ & -0.887 \\
\addlinespace[2pt]
Boundary recall & Felzenszwalb & 0.9772 $\pm$ 0.0357 & 0.7247 $\pm$ 0.1251 & 0.2525 & [0.2352, 0.2702] & $4.30844\times 10^{-34}$ & 1.000 \\
Boundary recall & Watershed--RAG & 0.9772 $\pm$ 0.0357 & 0.6765 $\pm$ 0.2000 & 0.3007 & [0.2744, 0.3275] & $4.30844\times 10^{-34}$ & 1.000 \\
Boundary recall & Quickshift & 0.9772 $\pm$ 0.0357 & 0.6854 $\pm$ 0.0928 & 0.2918 & [0.2787, 0.3049] & $4.30844\times 10^{-34}$ & 1.000 \\
\addlinespace[2pt]
Boundary F1 & Felzenszwalb & 0.5530 $\pm$ 0.1219 & 0.6764 $\pm$ 0.1200 & -0.1235 & [-0.1367, -0.1105] & $1.72981\times 10^{-29}$ & -0.927 \\
Boundary F1 & Watershed--RAG & 0.5530 $\pm$ 0.1219 & 0.5539 $\pm$ 0.1613 & -0.0009 & [-0.0166, 0.0148] & 0.645513 & -0.038 \\
Boundary F1 & Quickshift & 0.5530 $\pm$ 0.1219 & 0.5641 $\pm$ 0.1143 & -0.0112 & [-0.0222, 0.0003] & 0.0146396 & -0.219 \\
\addlinespace[2pt]
Group count & Felzenszwalb & 100.31 $\pm$ 50.35 & 64.50 $\pm$ 23.83 & 35.80 & [30.30, 41.56] & $6.97114\times 10^{-27}$ & 0.885 \\
Group count & Watershed--RAG & 100.31 $\pm$ 50.35 & 1097.82 $\pm$ 874.51 & -997.51 & [-1120.15, -882.67] & $5.82008\times 10^{-34}$ & -0.998 \\
Group count & Quickshift & 100.31 $\pm$ 50.35 & 97.89 $\pm$ 266.33 & 2.42 & [-39.65, 35.13] & $1.77251\times 10^{-21}$ & 0.780 \\
\addlinespace[2pt]
Runtime (s) & Felzenszwalb & 0.1064 $\pm$ 0.0121 & 0.1504 $\pm$ 0.0145 & -0.0440 & [-0.0465, -0.0419] & $4.30844\times 10^{-34}$ & -1.000 \\
Runtime (s) & Watershed--RAG & 0.1064 $\pm$ 0.0121 & 0.2749 $\pm$ 0.2868 & -0.1686 & [-0.2133, -0.1357] & $5.50798\times 10^{-32}$ & -0.960 \\
Runtime (s) & Quickshift & 0.1064 $\pm$ 0.0121 & 4.0096 $\pm$ 0.0922 & -3.9033 & [-3.9164, -3.8911] & $4.30844\times 10^{-34}$ & -1.000 \\
\end{longtable}
\end{landscape}